\PassOptionsToPackage{table}{xcolor}
\documentclass[11pt,a4paper]{youtu}
\PassOptionsToPackage{table}{xcolor}
\usepackage[authoryear,round]{natbib}
\usepackage{fontspec}
\newfontfamily\zhfont{FandolSong-Regular.otf}[BoldFont=FandolSong-Bold.otf]
\newcommand{\zh}[1]{{\zhfont\XeTeXlinebreaklocale "zh"\XeTeXlinebreakskip=0pt plus 1pt #1}}
\usepackage{microtype}
\usepackage{amsmath,amssymb,amsthm}
\usepackage{graphicx,booktabs,multirow,tabularx,makecell,bigdelim}
\usepackage{adjustbox}
\usepackage{caption,subcaption}
\usepackage{xcolor,xspace,pifont,enumitem}
\usepackage{placeins}
\usepackage{wrapfig}
\usepackage{needspace}
\usepackage[most]{tcolorbox}
\usepackage{listings}
\tcbset{breakable,boxrule=0.5pt,left=4pt,right=4pt,top=4pt,bottom=4pt,fonttitle=\bfseries\small}
\usepackage{xfp}
\newtheorem{definition}{Definition}
\definecolor{rpolight}{RGB}{245,248,255}
\definecolor{TabBlue}{HTML}{00A2E8}
\definecolor{TabRed}{HTML}{FF99CC}
\definecolor{myorange}{HTML}{F39C12}
\definecolor{mycyan}{HTML}{48C9B0}
\definecolor{mypink}{HTML}{C0392B}
\definecolor{myblue}{HTML}{2980B9}
\definecolor{SymbolGray}{HTML}{333333}
\newcommand{\dotc}[1]{\textcolor{#1}{$\bullet$}\xspace}
\newcommand{\mycmark}{\textcolor{myblue}{\ding{51}}}
\newcommand{\myxmark}{\textcolor{mypink}{\ding{55}}}

\newcommand{\valdiff}[2]{#1\hspace{0.3em}{\scriptsize\edef\diffval{\fpeval{round(#1-(#2),1)}}\ifdim\diffval pt>0pt\textcolor{green!60!black}{+\diffval}\else\textcolor{red}{\diffval}\fi}}

\title{VDE Bench: Evaluating the Capability of Image Editing Models to Modify Visual Documents}
\renewcommand\authorformat[2][]{\par\begingroup\centering\small\setstretch{1.15}\tencentsanswseven{#2$^{#1}$}\par\endgroup}
\renewcommand\affiliationformat[2][]{\par\begingroup\centering\small\tencentsanswseven{$^{#1}$#2}\par\endgroup}
\renewcommand\contributionlist{\par\vspace{3pt}{\centering\footnotesize $^{\ddagger}$Corresponding author.\par}}
\author{
Hongzhu Yi\textsuperscript{$1,8$}, 
Yujia Yang\textsuperscript{$1,8$},  
Yuanxiang Wang\textsuperscript{$1,8$}, 
Tong Li\textsuperscript{$5$}, 
Zhenyu Guan\textsuperscript{$1,8$}, \\
Tianyu Zong\textsuperscript{$1,8$}, 
Jiahuan Chen\textsuperscript{$6$}, 
Chenxi Bao\textsuperscript{$1$}, 
Tiankun Yang\textsuperscript{$1,8$}, 
Haopeng Jin\textsuperscript{$3,8$}, 
Yixuan Yuan\textsuperscript{$7$}, 
Xinming Wang\textsuperscript{$2$}, 
Tao Yu\textsuperscript{$3,8$},
Ruilin Gao\textsuperscript{$4$}, 
Ruiwen Tao\textsuperscript{$3$}, 
Haijin Liang\textsuperscript{$3$}, 
Jin Ma\textsuperscript{$3$}, 
Jinwen Luo\textsuperscript{$3$}, 
Yeshani\textsuperscript{$3$}, 
Xinyu Zuo\textsuperscript{$3$}, 
Bingshi Kang\textsuperscript{$1$}, 
Jungang Xu\textsuperscript{$1,8, \ddagger$}
}

\affiliation{
\textsuperscript{$1$}UCAS
\textsuperscript{$2$}CASIA \ 
\textsuperscript{$3$}Tencent \ 
\textsuperscript{$4$}CMU \
\textsuperscript{$5$}WashU \
\textsuperscript{$6$}SJTU \
\textsuperscript{$7$}XDU \ 
\textsuperscript{$8$}Coserve AI
}

\hypersetup{pdfauthor={Hongzhu Yi, Yujia Yang, Yuanxiang Wang, Tong Li, Zhenyu Guan, Tianyu Zong, Jiahuan Chen, Chenxi Bao, Tiankun Yang, Haopeng Jin, Yixuan Yuan, Xinming Wang, Tao Yu, Ruilin Gao, Ruiwen Tao, Haijin Liang, Jin Ma, Jinwen Luo, Yeshani, Xinyu Zuo, Bingshi Kang, Jungang Xu}}
\abstract{
Image editing models can modify visual content through natural language, but editing dense documents requires precise local changes amid tightly coupled text, tables, and layout. We introduce \textbf{VDE Bench}, an evaluation resource for studying instruction-guided editing of Chinese and English document images. VDE Bench contains 3,506 instruction-guided image editing samples, including 1,578 text-region edits and 1,928 table-region edits. The benchmark pairs editing instructions and reference images with annotated editing regions and structured OCR outputs. Its evaluation protocol combines region-level and full-page measurements to assess textual correctness, structural fidelity, and visual content preservation. By focusing on the interaction between local edits and their surrounding document context, VDE Bench complements existing visual text editing benchmarks and provides a resource for diagnosing document-specific editing failures.
}
\metadata[GitHub]{\url{https://github.com/zion-zion-zion/VDE-Bench}}
\metadata[Hugging Face]{\url{https://huggingface.co/datasets/hongzhuyi/VDEBench}}
\begin{document}
\maketitle
\setlength{\parskip}{5pt plus 1pt}
\raggedbottom

% 中文标题：引言
\section{Introduction}

% 中文：我们提出视觉文档编辑基准（Visual Document Editing Bench，VDE Bench），用于评测指令引导的密集文档图像编辑。
We introduce \textbf{V}isual \textbf{D}ocument \textbf{E}diting Bench (VDE Bench), a benchmark for evaluating instruction-guided editing of dense document images.

\label{sec:intro}
% 中文标题：研究动机
\subsection{Motivation}
% 中文：图像编辑模型日益支持通过自然语言直接操作视觉内容。
% 本文研究仅以图像为输入的场景：模型接收一张渲染、扫描或拍摄得到的复杂文档页面，以及一条编辑指令，然后由图像编辑模型根据编辑指令精确修改该文档页面内容。文档编辑对指令式图像提出了十分严格要求：无论替换一个短语还是修改一个表格单元格，都需要在相邻文本中准确定位目标，同时保持页面的可读性与组织结构，这对当前的图像编辑模型而言是最为复杂的任务。
\noindent\begin{minipage}[t]{0.55\textwidth}
\vspace{0pt}
Image editing models increasingly support direct manipulation of visual content through natural language \citep{wu2025qwenimagetechnicalreport,liu2025step1x-edit,LongCat-Image}. We study an image-only setting in which a model receives a rendered, scanned, or photographed page of a complex document together with an editing instruction, and precisely modifies the page content according to that instruction. Document editing places stringent demands on instruction-guided image editing: whether replacing a phrase or modifying a table cell, the model must accurately locate the target among neighboring text while preserving the page's readability and organization. These demands make document editing one of the most complex tasks for current image editing models.
\end{minipage}\hfill
\begin{minipage}[t]{0.42\textwidth}
\vspace{0pt}
\captionsetup{font=footnotesize,skip=4pt,hypcap=false}
    \centering
    \includegraphics[width=0.82\linewidth]{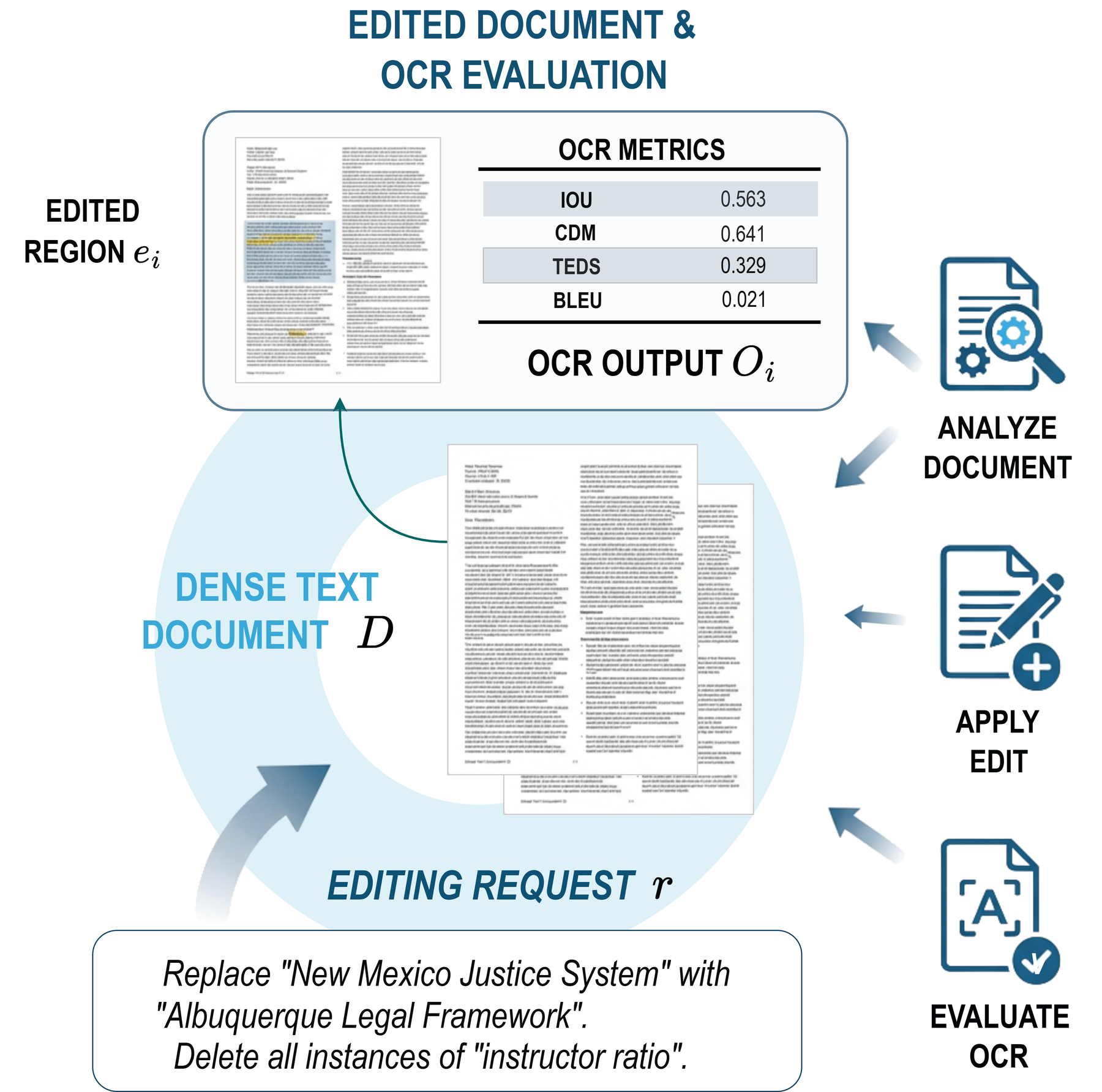}

% 中文：给定文档 D 和编辑请求 r，流程首先分析 D 的版面布局，再执行 r，生成编辑后的文档 D′ 及其编辑区域集合 {e_i}，最后运行 OCR，得到识别输出集合 {O_i}。
    \captionof{figure}{Given a document $D$ and an editing request $r$, the pipeline first analyzes the layout of $D$, then applies $r$ to produce the edited document $D'$ with edited regions $\{e_i\}$, and finally runs OCR to obtain outputs $\{O_i\}$.}
    \label{fig:vde define}
    
\end{minipage}
\par\medskip

% 中文：尽管有一些工作，例如AnyText、GlyphControl 和 TextCtrl 已研究视觉文本生成与编辑。TextEditBench 与 TextEdit 则采用互补的评测协议，进一步评估指令驱动的文本修改。但这些工作的文本修改并不复杂，这进而引出了一个更具体的问题：针对极其密集的文档页面中的文本区域与表格内容时，图像编辑模型究竟能在多大程度上可靠地完成修改？

Prior work, including AnyText, GlyphControl, and TextCtrl, has explored visual text generation and editing \citep{tuo2024anytextmultilingualvisualtext,yang2024glyphcontrol,zeng2024textctrldiffusionbasedscenetext}. TextEditBench and TextEdit further evaluate instruction-driven text modification through complementary assessment protocols \citep{gui2025texteditbenchevaluatingreasoningawaretext,tian2026internvl}. However, the text modifications considered in these works are relatively simple. This raises a more specific question: how reliably can image editing models modify text regions and tabular content within extremely dense document pages?

% 中文：VDE Bench 正是面向这一文档场景而构建。论文、试卷、报告等版面要求模型同时做到三点：准确定位编辑目标、精确完成文本修改，并保持周边内容不变。为在不同语言下考察这些能力，VDE Bench 同时涵盖中文与英文文档。借助编辑区域标注，我们可以将局部错误与整页变化结合起来分析，而不必将模型表现压缩为单一的图像级相似度分数。由此，VDE Bench 在统一的评测资源中同时考察精确文本修改、区域定位与文档上下文保持。
VDE Bench is built specifically for this document setting. Layouts such as academic papers, examination papers, and reports require models to \emph{locate} the editing target accurately, \emph{apply} the precise textual change, and \emph{preserve} the surrounding content. VDE Bench includes both Chinese and English documents to examine these capabilities across languages. Editing-region annotations allow local errors to be analyzed alongside full-page changes, without reducing model performance to a single image-level similarity score. VDE Bench thus assesses precise text editing, region localization, and document-context preservation within a unified evaluation resource.

\begin{table*}[t]

\normalsize
\renewcommand{\arraystretch}{1}
\begin{tabular*}{\textwidth}{@{\extracolsep{\fill}}p{5.2cm} p{1.4cm} >{\centering\arraybackslash}p{1.45cm} >{\centering\arraybackslash}p{1.35cm} >{\centering\arraybackslash}p{1.1cm} p{2.8cm}@{}}
\toprule
\multirow{2}{*}{\textbf{Name and Reference}} & \multirow{2}{*}{\textbf{Turns}} & \textbf{Human} & \textbf{Text} & \multirow{2}{*}{\textbf{Mask}} & \makecell[c]{\textbf{Problem}} \\
& & \textbf{Verified} & \textbf{Edited} & & \makecell[c]{\textbf{Framing}} \\
\midrule
I2EBench \citep{ma2024i2ebench} & \dotc{myorange} single & \mycmark & \myxmark & \myxmark
& \rdelim\}{18}{2.2cm}[\parbox{1.8cm}{\raggedright\textbf{General image editing}}]
\\
    EditBench \citep{wang2023imagen} & \dotc{myorange} single & \mycmark & \mycmark & \mycmark \\
EditVal \citep{basu2023editval} & \dotc{myorange} single & \mycmark & \myxmark & \myxmark \\
EmuEdit \citep{sheynin2024emu} & \dotc{myorange} single & \mycmark & \myxmark & \myxmark \\
AnyEdit \citep{Yu_2025_CVPR} & \dotc{myorange} single & \mycmark & \myxmark & \mycmark \\
CompBench \citep{jia2025compbench} & \dotc{myblue} multi & \mycmark & \myxmark & \myxmark \\
Omni-IIE Bench \citep{yang2026omniiiebench} & \dotc{myblue} multi & \mycmark & \myxmark & \mycmark \\
MagicBrush \citep{zhang2023magicbrush} & \dotc{myblue} multi & \mycmark & \myxmark & \mycmark \\
ImgEdit-Bench \citep{ye2025imgedit} & \dotc{myblue} multi & \mycmark & \myxmark & \myxmark \\
MuCIE \citep{zhou2025multi} & \dotc{myblue} multi & \myxmark & \myxmark & \myxmark \\
GIE-Bench \citep{qian2025giebench} & \dotc{myorange} single & \mycmark & \myxmark & \mycmark \\
Complex-Edit \citep{yang2025complexedit} & \dotc{myorange} single & \mycmark & \myxmark & \myxmark \\
EBench-18K \citep{xu2025lmm4edit} & \dotc{myorange} single & \mycmark & \myxmark & \myxmark \\
HQ-Edit \citep{hui2024hqedit} & \dotc{myorange} single & \myxmark & \myxmark & \myxmark \\
AURORA-Bench \citep{krojer2024aurora} & \dotc{myorange} single & \mycmark & \myxmark & \myxmark \\
PIE-Bench++ \citep{huang2024paralleledits} & \dotc{myorange} single & \mycmark & \myxmark & \myxmark \\
TEdBench++ \citep{brack2024leditslimitlessimageediting} & \dotc{myorange} single & \mycmark & \myxmark & \myxmark \\
ImagenWorld \citep{sani2026imagenworld} & \dotc{myorange} single & \mycmark & \myxmark & \myxmark

\\

AnyText-Bench \citep{tuo2024anytextmultilingualvisualtext} & \dotc{myorange} single & \mycmark & \mycmark & \mycmark
& \rdelim\}{4}{2.2cm}[\parbox{1.8cm}{\raggedright\textbf{Text-in-image editing}}]
\\
TextEditBench \citep{gui2025texteditbenchevaluatingreasoningawaretext} & \dotc{myorange} single & \mycmark & \mycmark & \mycmark \\
TextEdit \citep{tian2026internvl} & \dotc{myorange} single & \mycmark & \mycmark & \myxmark \\
Kontext-Bench \citep{labs2025flux1kontext} & \dotc{myorange} single & \myxmark & \mycmark & \myxmark

\\
\midrule
\addlinespace[2pt]
\makecell[l]{\textbf{VDE Bench (Ours)}} & \dotc{myorange} single & \mycmark & \mycmark & \mycmark
& {\fontsize{9}{10.5}\selectfont\makecell[l]{\textbf{Complex document}\\\textbf{editing}}}
\\
\bottomrule
\end{tabular*}
% 中文：VDE Bench 与现有图像编辑基准的比较。
% VDE Bench 专注于复杂文档编辑，提供经过人工核验的参考结果和编辑区域标注。
% “multi”表示包含多轮评测；文字列包括文字渲染或编辑。
\caption{\textbf{Comparison of VDE Bench with existing image editing benchmarks.} VDE Bench focuses on complex document editing with human-verified references and editing-region annotations. Multi denotes inclusion of multi-turn evaluation; Text Edited includes text rendering or editing; $\times$ denotes an unsupported or unreported feature.}
\label{tab:compare}
\end{table*}

% 中文标题：指导原则
\subsection{Guiding Principles}
\label{sec:guiding principles}

% 中文：评测应当区分所请求修改的正确性与文档上下文的保持情况。
% 因此，VDE Bench 提供成对的源图像与参考图像，以及编辑区域标注，从而支持相互补充的局部测量与整页测量。
Evaluation should distinguish the correctness of the requested change from the preservation of the document context. VDE Bench therefore provides paired source and reference images together with editing-region annotations to support complementary local and full-page measurements.

% 中文：VDE Bench 从两个互补维度评估密集文本文档上的编辑操作质量：OCR 文本层面的评测（见附录中的文本内容指标部分），以及图像视觉层面的评测（见图像指标附录）。
% 下面给出 VDE Bench 任务的形式化定义。
VDE Bench evaluates editing quality in dense text documents along two complementary dimensions: \textbf{OCR-based textual evaluation} (Appendix~\ref{subsec:text content metrics}) and \textbf{image-based visual evaluation} (Appendix~\ref{sec:image metrics}). We formally define the VDE Bench task as follows.

% 中文：给定密集文本文档 D 和编辑请求 r，将 r 应用于 D，得到编辑后的文档 D′ 及其对应的编辑区域集合 {e_i}。
% 同时，对编辑后的文档运行 OCR 系统，获得识别输出集合 {O_i}。
% 最终，通过一组评测指标，衡量模型生成的输出 (O_i, e_i, D′) 与真实参考结果 ({O_i*}, {e_i*}, D′*) 之间的差异；完整形式化表述见“期望性质”附录。
Given a dense text document $D$ and an editing request $r$, we apply $r$ to $D$ to obtain the edited document $D'$ and its corresponding set of edited regions $\{e_i\}$. We also run an OCR system on the edited document to obtain the recognition outputs $\{O_i\}$. A suite of evaluation metrics then measures the discrepancy between the model-generated outputs $(O_i, e_i, D')$ and the ground truth $(\{O_i^{*}\}, \{e_i^{*}\}, D'^{*})$ (see Appendix~\ref{sec:properties} for the full formulation).

% 中文标题：主要贡献
\subsection{Contributions}

% 中文：文档编辑资源。
% 我们提供3,506 条文档编辑评测样本，其中包含 1,578 条文本区域编辑与 1,928 条表格区域编辑。每条样本都将一条编辑指令、一张原图像和一张参考图像和与编辑区域标注、结构化 OCR 标注关联起来。
\noindent\textbf{Document editing resource.}
We provide 3,506 document-editing evaluation samples, comprising 1,578 text-region edits and 1,928 table-region edits. Each sample pairs an editing instruction, a source image, and a reference image with editing-region and structured OCR annotations.

% 中文：区域感知评测。
% 我们在局部与整页两个尺度上组织基于 OCR 和基于图像的测量，以考察密集文档编辑中的文本准确性、结构保真度与视觉内容保持情况，详见“指导原则”一节。
\noindent\textbf{Region-aware evaluation.}
We organize OCR-based and image-based measurements at both local and full-page scales to assess textual accuracy, structural fidelity, and visual content preservation in dense document editing (\S\ref{sec:guiding principles}).

% 中文：诊断性分析。
% 我们在 3,506 条样本版本上，按编辑类型和语言比较七种模型，并给出当前图像编辑模型失败模式的分析。
\noindent\textbf{Diagnostic analysis.}
We compare seven models across edit types and languages on the 3,506-sample release and analyze the failure modes of current image editing models.
% These analyses characterize model behavior under document-specific editing constraints.

% 中文标题：相关工作
\section{Related Work}
\label{sec:related work}

% 中文：图像编辑模型与基准。
% InstructPix2Pix 从合成的训练样本对中学习遵循编辑指令。
% 近期模型包括 Step1X-Edit、Qwen-Image、LongCat-Image 和 Nano Banana Pro。
% 通用编辑基准主要考察指令遵循与内容保持。
\paragraph{Image Editing Models and Benchmarks.}
InstructPix2Pix learns to follow editing instructions from synthesized training pairs \citep{brooks2023instructpix2pix}. Recent models include Step1X-Edit, Qwen-Image, LongCat-Image, and Nano Banana Pro \citep{liu2025step1x-edit,wu2025qwenimagetechnicalreport,LongCat-Image,google2025nanoBanana}. Riemannian Residual Line Search selects among candidates along a residual editing path to balance target-prompt alignment and source preservation \citep{yi2026rrls}. General editing benchmarks examine instruction following and content preservation \citep{sheynin2024emu,basu2023editval,ma2024i2ebench,jia2025compbench,ye2025imgedit}. Omni IIE Bench further diagnoses editing consistency across semantic scales and successive instructions \citep{yang2026omniiiebench}.

% 中文：视觉文本生成与编辑。
% 场景文本编辑研究如何在保持外观的同时替换文本内容。
% AnyText、GlyphControl 和 TextCtrl 构建了具有文本感知能力的生成或编辑机制。
% TextEditBench 和 TextEdit 评估细粒度文本修改，而 CVTG-2K 和 ChineseWord 考察视觉文本生成。
% VDE Bench 通过研究密集文档页面中的局部编辑，对这些资源形成补充。
\paragraph{Visual Text Generation and Editing.}
Scene-text editing studies how to replace textual content while preserving its appearance \citep{wu2019srnet,qu2023mostel}. AnyText, GlyphControl, and TextCtrl develop text-aware generation or editing mechanisms \citep{tuo2024anytextmultilingualvisualtext,yang2024glyphcontrol,zeng2024textctrldiffusionbasedscenetext}. TextEditBench and TextEdit evaluate fine-grained text manipulation \citep{gui2025texteditbenchevaluatingreasoningawaretext,tian2026internvl}, while CVTG-2K and ChineseWord examine visual text generation \citep{du2025textcrafteraccuratelyrenderingmultiple,wu2025qwenimagetechnicalreport}. VDE Bench complements these resources by studying localized editing within dense document pages.

% 中文：文档理解与编辑。
% 布局感知模型与无需 OCR 的模型分别捕捉文档内容的不同方面，二者相互补充。
% UDOP 曾将文档理解、生成和编辑统一起来，但仅局限于英文页面，VDE Bench则引入了更复杂的中文。
% InfoEdBench 利用区域级指令评测密集信息图中的多区域同步文本替换，但仅聚焦文本翻译。而 APEX-Bench 研究基于可编辑 PPTX 表示的指令式学术海报编辑，文档版式较为单一。
% VDE Bench 聚焦文档图像中的精确文本与表格编辑，并涵盖极为广泛的文档格式。
% 扩展研究基础附录进一步讨论文档、表格、渲染及评测领域的相关研究。
\paragraph{Document Understanding and Editing.}
Layout-aware and OCR-free models capture complementary aspects of document content \citep{xu2020layoutlm,kim2022donut}. UDOP unifies document understanding, generation, and editing in an English-focused setting \citep{tang2023udop}, while VDE Bench extends evaluation to more challenging Chinese documents. InfoEdBench evaluates simultaneous text replacement in dense infographics using region-level instructions, but focuses exclusively on text translation \citep{zaccagnino2026infoedbench}. APEX-Bench studies instruction-guided academic poster editing with editable PPTX representations, covering a relatively narrow range of document layouts \citep{shi2026apex}. PaperX uses a structured intermediate representation to generate multiple academic presentation formats from scientific papers \citep{yu2026paperx}. VDE Bench focuses on precise text and table edits in document images across a wide range of document types and layouts. Appendix~\ref{sec:expanded foundations} further discusses related research on documents, tables, rendering, and evaluation.

% 中文标题：VDE Bench 的构建
\section{VDE Bench Construction}
\label{sec:visual doc bench}

\begin{figure*}[t]
    \centering
    \includegraphics[width=\linewidth]{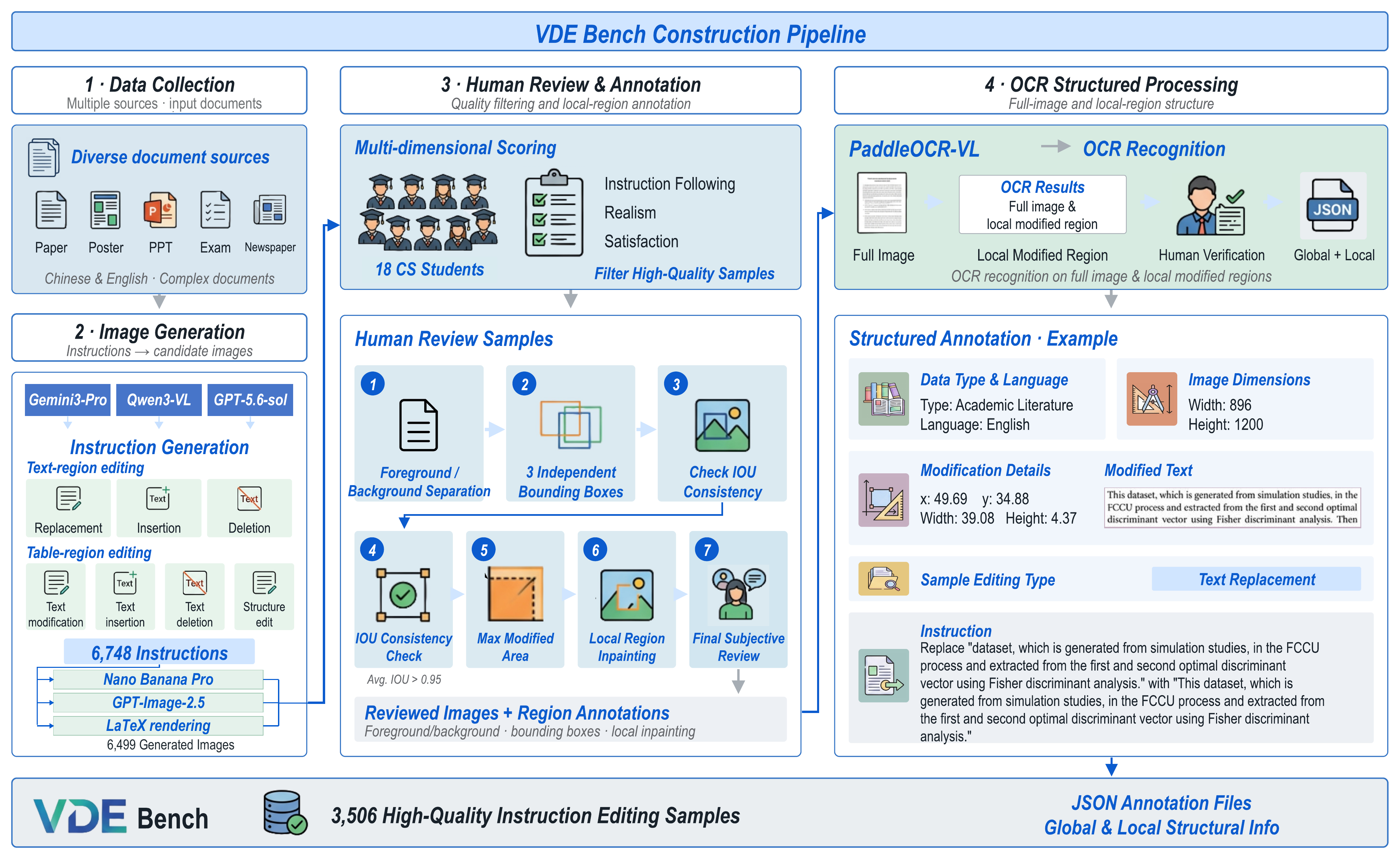}
% 中文：VDE Bench 构建流程与样本标注概览。
% 流程包括四个阶段：（1）收集涵盖多个类别与多种语言的文档；（2）通过 Gemini3-Pro、Qwen3-VL 和 GPT-5.6-sol 生成编辑指令，并使用 Nano Banana Pro/ GPT-Image-2.5-Sunburt或者直接使用latex渲染生成编辑后的图像；（3）由 18 名标注人员开展人工审核与标注，包括多维度评分、基于 IoU 的区域核验和主观质量筛选；（4）使用 PaddleOCR-VL 进行结构化 OCR 处理，并通过人工核验。
% VDE Bench 包含 3,506 条高质量指令编辑样本，并提供丰富的结构标注，右侧展示了一个示例。
    \caption{\textbf{Overview of the VDE Bench construction pipeline and a sample annotation.}
The pipeline consists of four stages:
(1)~collecting documents across multiple categories and languages;
(2)~generating editing instructions with Gemini3-Pro, Qwen3-VL, and GPT-5.6-sol and producing edited images with Nano Banana Pro, GPT-Image-2.5-Sunburt, or direct LaTeX rendering;
(3)~human review and annotation by 18 annotators, including multidimensional scoring, IoU-based region verification, and subjective quality filtering;
(4)~structured OCR processing with PaddleOCR-VL followed by human verification.
VDE Bench contains 3{,}506 high-quality instruction-based editing samples with rich structural annotations; an example is shown on the right.}
    \label{fig:overview of annotion}
\end{figure*}

% 中文：构建 VDE Bench 需要大量人工标注与处理。
% 图中的总体框架展示了四个阶段：数据收集、候选编辑图像生成、人工审核与区域标注，以及结构化 OCR 处理与人工核验。
% VDE Bench 与社区现有开源图像编辑基准的主要比较汇总于基准对比表；已标注数据的代表性示例见案例研究附录。
Constructing VDE Bench requires extensive manual annotation and processing. Figure~\ref{fig:overview of annotion} illustrates the overall pipeline, which comprises four stages: data collection, candidate image generation, human review and region annotation, and structured OCR processing with human verification. Table~\ref{tab:compare} summarizes the main comparisons between VDE Bench and existing open-source image editing benchmarks. Representative annotated examples are provided in Appendix~\ref{sec: append case study}.

% 中文标题：数据收集阶段
\subsection{Data Collection Stage}
% 中文：我们从 OmniDocBench、DocLayNet、PubMed Central（PMC）以及通过 DocGenome 获取的 arXiv 论文中收集中英文文档页面，作为种子数据。
% OmniDocBench 是一个高质量 OCR 基准，包含大量经过人工标注的复杂文本文档，更多细节见数据集说明附录。
\begin{wrapfigure}{R}{0.48\textwidth}
\vspace{-6pt}
\captionsetup{font=footnotesize,skip=4pt}
\centering
    \includegraphics[width=\linewidth]{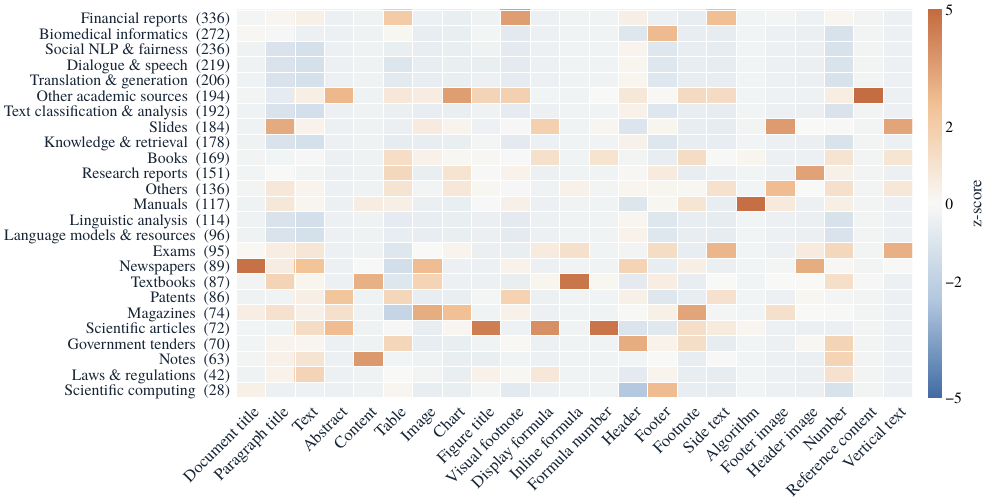}
% 中文：VDE Bench 的 GT 页面布局密度。
% 每个单元格显示 z_{s,c} = (d_{s,c} − μ_c) / σ_c，其中 d_{s,c} 表示来源组 s 中，先将每页类别 c 的全部元素边界框面积相加并除以页面面积，再对页面取平均所得的密度。
% 对每类元素，均值和标准差均在 与数据分布图一致的 25 个文档类别之间计算。
    \caption{\textbf{GT page layout density in VDE Bench.} Each cell shows $z_{s,c}=(d_{s,c}-\mu_c)/\sigma_c$. For each page in source group $s$, we sum the bounding-box areas of all elements in category $c$ and divide by the page area; $d_{s,c}$ is the mean of this ratio across pages. For each element category, the mean and standard deviation are computed across the same 25 document categories as in Figure~\ref{fig:dataset_statistics}.}
    \label{fig:Layout element density across document types}
\end{wrapfigure}

We collected Chinese and English document pages from \textbf{OmniDocBench}~\cite{ouyang2024omnidocbenchbenchmarkingdiversepdf}, DocLayNet~\cite{pfitzmann2022doclaynet}, PubMed Central (PMC), and arXiv papers accessed through DocGenome as seed data. OmniDocBench is a high-quality OCR benchmark containing extensive manually annotated complex text documents. Further dataset details are provided in Appendix~\ref{sec:dataset card}.

% 中文：布局与文档类型的多样性。
% 我们利用 PaddleOCR-VL 的布局解析输出，刻画全部 3,506 张编辑后真实参考（GT）页面的布局。
% 布局密度图展示了 25 个文档类别中 23 类元素的面积占比密度，并按列进行 z-score 标准化。
% 财务报告的平均表格密度最高，报纸的文本密度最高，科学文章的独立公式密度最高。
% 这些分布为评估异构文档布局中的编辑提供了互补场景，详见布局元素分布分析附录。
\paragraph{Layout and Document Types Diversity.} We characterize the layouts of all 3,506 edited ground-truth (GT) pages using their PaddleOCR-VL layout outputs. Figure~\ref{fig:Layout element density across document types} shows the column-wise z-score of area-ratio density for 23 element categories across 25 document categories. Financial reports have the highest mean table density, newspapers the highest text density, and scientific articles the highest display-formula density. These profiles provide complementary settings for evaluating edits in heterogeneous document layouts (Appendix~\ref{sec:layout element distribution analysis}).

\begin{figure*}[t]
\captionsetup{font=footnotesize,skip=4pt}
\centering
    \includegraphics[width=\textwidth]{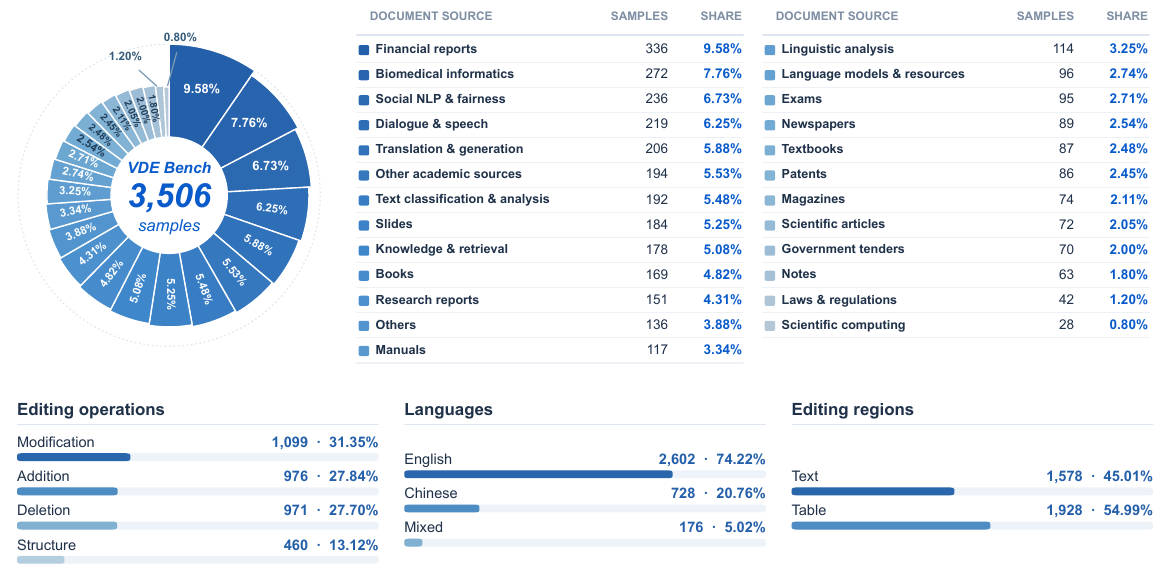}
% 中文：VDE Bench 数据集概况。
% 上方展示文档来源分布，其中学术文献按研究主题进一步划分；扇形角度表示各来源所占比例。
% 下方依次展示编辑操作、语言和编辑区域分布。
% 标签标明数量与百分比。
% Mixed 表示中英文混合文档。
    \caption{\textbf{Dataset profile of VDE Bench.} Top: document-source distribution, with academic literature subdivided by research topic. Sector angles indicate source proportions. Bottom: editing operations, languages, and editing regions. Labels report counts and percentages. Mixed denotes Chinese--English documents.}
    \label{fig:dataset_statistics}
\end{figure*}

% 中文标题：图像生成阶段
\subsection{Image Generation Stage}
% 中文：在图像生成阶段，编辑指令分为七种“区域—操作”类别：文本替换、文本插入、文本删除，以及表格内文本替换、表格内文本插入、表格内文本删除和表格结构修改。我们使用一系列SOTA多模态模型进行数据合成并由人工筛选和修正。具体而言，首先针对文档页面图像，我们使用 Gemini3-Pro、GPT-5.6 sol等生成编辑指令，并从类型感知提示词池中对上述各类指令进行采样，这么做的目的是避免图像编辑风格陷入一个模型的风格特征里，提示词的详情见提示词池附录。随后我们使用Nano Banana Pro、GPT-Image-2.5-Sunburt根据编辑指令直接修改图像，而对于那些有源latex文件的文档，我们则直接让GPT-5.6-sol和Claude-opus4.8直接修改源码。
In the image generation stage, editing instructions fall into seven region--operation categories: text replacement, text insertion, text deletion, in-table text replacement, in-table text insertion, in-table text deletion, and table structure modification. We use a range of state-of-the-art multimodal models to synthesize data, followed by manual selection and correction. Specifically, we first use models including Gemini3-Pro and GPT-5.6 sol to generate editing instructions for document page images, sampling instructions across these categories from a type-aware prompt pool. This approach avoids concentrating the editing style around the characteristics of a single model. Details of the prompt pool are provided in Appendix~\ref{sec:prompt pool}. We then use Nano Banana Pro and GPT-Image-2.5-Sunburt to edit images directly according to the instructions. For documents with available LaTeX source files, we instead use GPT-5.6-sol and Claude-opus4.8 to modify the source code directly.

% 中文：通过直接图像编辑和 LaTeX 源码编辑两种路线，在完成生成及适用的安全检查后，我们共获得 6,499 张候选编辑图像，供后续人工审核。
Together, direct image editing and LaTeX-source editing yield a pool of \textbf{6{,}499 candidate edited images} after generation and applicable safety checks. These candidates undergo the subsequent manual review.

% 中文：图像生成后，为获取编辑图像的页面布局，我们使用 PaddleOCR-VL 对其进行布局检测和文本识别，得到 6499 个包含图像 OCR 信息的 JSON 文件。随后对这些数据执行下述操作。
After image generation, we use PaddleOCR-VL for layout detection and text recognition to obtain the page layouts of the edited images, yielding \textbf{6{,}499 JSON files} containing their OCR information. We then apply the following procedures to these data.
% 中文标题：人工审核阶段
\subsection{Manual Review Stage}
% 中文：图像生成完成后，我们开展了严格的人工审核。
% 人工审核阶段的工作流程如下。
After image generation, we conducted a rigorous manual review stage. The workflow of the manual review stage is as follows.

% 中文：人工标注规则对齐。
% 标注团队由 18 名计算机科学专业的硕士生和博士生组成。
% 正式标注前，我们组织培训，确保所有标注人员充分理解标注指南，并建立统一的标注标准。
% 标注涉及三个指标：指令遵循程度、修改真实性和主观满意度。
% 每项指标采用 1 至 3 分的评分尺度，其中 1 分表示最差，3 分表示最好，详见人工标注协议附录。
\paragraph{Human Annotation Rules Alignment.} 
\label{sec:human annotation rules alignment}
Our annotation team consists of 18 master's and doctoral students majoring in computer science. Prior to the formal annotation, we conducted training sessions to ensure that all annotators fully understood the annotation guidelines and established a unified annotation standard. The annotation process involves three metrics: Instruction Compliance, Modification Authenticity, and Subjective Satisfaction. Each metric is scored on a scale of 1 to 3, where 1 indicates the worst and 3 indicates the best (see Appendix \ref{sec: append human annotation protocol} for details). 
% 中文：全部 18 名标注人员独立评定相同的 50 条试标样本，因此每个维度共有 900 次评分。
% 指令遵循程度、修改真实性和主观满意度的 Krippendorff α 系数分别为 0.941、0.950 和 0.876，见人工一致性图。
% 在正式标注中，样本被接受的条件为：质量总分达到 9 分、三个标注组达成一致，以及边界框的平均两两 IoU 大于 0.95。
All 18 annotators independently scored the same 50 pilot samples, giving 900 ratings per dimension. Krippendorff's $\alpha$ was 0.941 for instruction compliance, 0.950 for modification authenticity, and 0.876 for subjective satisfaction (Figure~\ref{fig:human_agreement}). During production, acceptance required a total quality score of 9, agreement among three annotation groups, and mean pairwise box IoU above 0.95.

% 中文：修改区域标注。
\paragraph{Modification Annotation.}

% 中文：在修改区域标注阶段，标注人员首先按照预先定义的规则对所有生成图像评分，收集达到质量阈值的一批图像，阈值见人工标注规则对齐部分。
% 之后，进一步标注这些合格图像中修改区域的边界框。
% 最后，使用 PaddleOCR-VL 识别生成的 GT 图像，并额外开展一轮人工核验，详情见标注细节附录。
% 构建流程概览图展示了整个标注流程，更多信息见标注与人工评测附录。
During the Modification Annotation stage, annotators first score all generated images according to the predefined rules and collect a batch of images that meet the quality threshold (\S \ref{sec:human annotation rules alignment}). After collecting these qualified images, annotators further annotate the bounding boxes of the modified regions. Finally, PaddleOCR-VL is applied to recognize the generated ground truth images, followed by an additional round of manual verification (see Appendix \ref{sec:annotation detail} for detail). Our annotation pipeline is illustrated in Figure ~\ref{fig:overview of annotion}, more detailed information can be found in Appendix~\ref{sec: append annotations}.

% 中文标题：统计分析
\subsection{Statistical Analysis}

% 中文：最终经过筛选的数据包含 3,506 条样本，其中 1,578 条为文本区域编辑，1,928 条为表格区域编辑。
% 按操作划分，修改、添加、删除和结构编辑分别为 1,099、976、971 和 460 条。
% 语言标签包含 2,602 条英文、728 条中文和 176 条混合语言样本。
The final filtered dataset contains \textbf{3,506} samples: 1,578 text-region and 1,928 table-region edits. Its operation counts are 1,099 modifications, 976 additions, 971 deletions, and 460 structural edits. Language labels comprise 2,602 English, 728 Chinese, and 176 mixed-language samples.

% 中文：数据集统计图汇总了文档来源、编辑操作、语言和编辑区域的分布。
% 学术文献共 1,735 条，并按研究主题进一步分组；财务报告共 336 条。
% 英文、中文和混合语言样本分别占 74.22%、20.76% 和 5.02%。
% 修改、添加和删除各占该版本约 28%—31%，结构编辑占 13.12%。
% 联合样本组成附录给出了语言、区域、操作与文档来源的联合计数。
Figure~\ref{fig:dataset_statistics} summarizes document sources, editing operations, languages, and editing regions. Academic literature accounts for 1,735 samples and is further grouped by research topic; financial reports contribute 336. English, Chinese, and mixed-language samples account for 74.22\%, 20.76\%, and 5.02\%, respectively. Modification, addition, and deletion each account for roughly 28--31\% of the release, while structural edits account for 13.12\%. Appendix~\ref{sec:joint_counts} gives the joint language, region, operation, and document-source counts.

% 中文标题：评测协议
\section{Evaluation Protocol}
\label{sec:eval}

% 中文：VDE Bench 在局部与全局两个尺度上评估编辑后的页面：局部评测作用于依据标注编辑区域裁剪出的图像，全局评测则作用于完整的生成页面。两个尺度评估的是同一份模型输出。其中，基于 OCR 的指标刻画编辑定位、字符准确性、文本重合度与结构化内容；图像指标则独立刻画视觉表现。
VDE Bench evaluates each edited page at two scales. Local evaluation operates on the crop defined by the annotated editing region, whereas global evaluation operates on the full generated page. Both scales assess the same model output. OCR-based metrics characterize localization, character accuracy, text overlap, and structured content, while image metrics independently characterize visual quality.

% 中文：空间定位。我们按候选框对的 IoU 从高到低进行一对一贪心匹配，匹配阈值为 0.1。最终 IoU 对全部参考框求平均，未匹配参考框计 0；匹配与汇总细节见附录。
\paragraph{Spatial localization.}
\label{sec:spatial metrics}
We greedily match predicted and reference OCR blocks in descending order of \textbf{IoU}, enforcing one-to-one correspondence and a threshold of 0.1. The reported IoU averages over all reference blocks, with unmatched reference blocks assigned zero (Appendix~\ref{sec: append match strategy}--\ref{sec: append iou}).

% 中文：文本与结构。对于成功匹配的文本块，CDM 衡量归一化的字符编辑相似度，BLEU-4 衡量 n 元组重合度，TEDS-like 则比较结构化的 OCR 表示（定义见指标附录）。编辑区域内的参考文本均经过人工核验。对于删除操作，编辑后的裁剪区域既可能仍包含周边文本，也可能变为空白；评测将这两种情况均视为有效结果并予以保留。
\paragraph{Text and structure.}
\label{sec:text content metrics}
For matched blocks, \textbf{CDM} measures normalized character-edit similarity, \textbf{BLEU-4} measures $n$-gram overlap, and \textbf{TEDS-like} compares structured OCR representations (Appendix~\ref{subsubsec:CDM}--\ref{subsubsec:TEDS-like Similarity}). All reference text within editing regions is manually verified. For deletion, the edited crop may either retain surrounding text or become blank; both outcomes are treated as valid and retained in the evaluation.

% 中文：视觉外观。PSNR、SSIM 与 LPIPS 将生成图像与参考图像进行比较，CLIP 则衡量图像与文本的对齐程度（详见图像指标附录）。这些结果与 OCR 分数分开报告。只要识别出的文本、文本结构与边界框保持不变，即使栅格化方式或颜色存在差异，基于 OCR 的分数也不会随之变化；边界框位置或尺寸的变化则由 IoU 单独刻画。这种解耦使我们能够将"改了什么、改在哪里"与"渲染得如何"区分开来。
\paragraph{Visual appearance.}
\textbf{PSNR}, \textbf{SSIM}, and \textbf{LPIPS} compare the generated image against the reference, while \textbf{CLIP} measures image--text alignment (Appendix~\ref{sec:image metrics}). These measurements are reported separately from the OCR scores. As long as the recognized text, its structure, and its bounding boxes remain unchanged, OCR-based scores are invariant to differences in rasterization or color; changes in box position or size are captured separately by IoU. This decoupling separates \emph{what} is edited and \emph{where} from \emph{how} it is rendered.

% 中文：报告配置。主表还额外报告了 TextSim 分数（定义见文本内容指标附录的 Text Similarity 小节）以及分布层面的 FID。所有模型共用同一套参考 OCR 结果，值得注意的一个细节是，由于部分被修改区域是会出现空白的情况，例如部分文字的大面积删除，因此OCR 输出为空的页面仍然算是有效评测结果。
\paragraph{Reported configuration.}
The main table additionally reports the TextSim score (Appendix~\ref{subsubsec:textsim}) and distribution-level FID. All models share the same reference OCR results. Edited regions may become blank, for example after deleting large portions of text, so pages with empty OCR output remain valid evaluation outcomes.
% 中文标题：基准评测与分析
\section{Benchmarks and Analysis}
\label{sec:experiment}

% 中文：全局与局部设置上下排列，共用 OCR 和图像指标列；指标统一保留三位小数。
\begin{table*}[!t]
\centering
\fontsize{8}{9.5}\selectfont
\setlength{\tabcolsep}{2.2pt}
\renewcommand{\arraystretch}{1.12}
\begin{tabularx}{\textwidth}{@{}l*{10}{>{\centering\arraybackslash}X}@{}}
\toprule
& \multicolumn{5}{c}{\textbf{OCR metrics} $\uparrow$} & \multicolumn{5}{c}{\textbf{Image metrics}} \\
\cmidrule(lr){2-6}\cmidrule(lr){7-11}
Model & IoU & TextSim & CDM & BLEU & TEDS & PSNR$\uparrow$ & SSIM$\uparrow$ & LPIPS$\downarrow$ & CLIP$\uparrow$ & FID$\downarrow$ \\
\midrule
\multicolumn{11}{l}{\textbf{Global} ($n=3{,}506$)} \\
\midrule
Step1X & \cellcolor[HTML]{99D0EE} 0.830 & \cellcolor[HTML]{B5DDF3} 0.615 & \cellcolor[HTML]{B2DBF3} 0.634 & \cellcolor[HTML]{C6E5F6} 0.304 & \cellcolor[HTML]{B9DFF4} 0.552 & \cellcolor[HTML]{B0DAF2} 20.096 & \cellcolor[HTML]{ADD9F2} 0.859 & \cellcolor[HTML]{F3CCDD} 0.230 & \cellcolor[HTML]{9AD0EF} 0.956 & \cellcolor[HTML]{E9A6C3} 5.763 \\
LongCat & \cellcolor[HTML]{8BC9EC} 0.876 & \cellcolor[HTML]{98CFEE} 0.779 & \cellcolor[HTML]{96CEEE} 0.794 & \cellcolor[HTML]{A9D7F1} 0.410 & \cellcolor[HTML]{91CCED} 0.780 & \cellcolor[HTML]{BDE0F4} 18.792 & \cellcolor[HTML]{ACD8F1} 0.861 & \cellcolor[HTML]{EBADC8} 0.127 & \cellcolor[HTML]{93CDED} 0.966 & \cellcolor[HTML]{E9A5C2} 5.018 \\
FireRed & \cellcolor[HTML]{91CCED} 0.856 & \cellcolor[HTML]{8CCAEC} 0.847 & \cellcolor[HTML]{8AC9EC} 0.862 & \cellcolor[HTML]{94CDEE} 0.485 & \cellcolor[HTML]{90CBED} 0.789 & \cellcolor[HTML]{A8D7F1} 20.803 & \cellcolor[HTML]{A0D3F0} 0.883 & \cellcolor[HTML]{EAA9C5} 0.112 & \cellcolor[HTML]{90CCED} 0.970 & \cellcolor[HTML]{E9A5C3} 5.273 \\
Qwen & \cellcolor[HTML]{84C6EB} \textbf{0.901} & \cellcolor[HTML]{84C6EB} \textbf{0.891} & \cellcolor[HTML]{84C6EB} \textbf{0.897} & \cellcolor[HTML]{84C6EB} \textbf{0.541} & \cellcolor[HTML]{84C6EB} \textbf{0.857} & \cellcolor[HTML]{84C6EB} \textbf{24.279} & \cellcolor[HTML]{84C6EB} \textbf{0.936} & \cellcolor[HTML]{E8A0BF} \textbf{0.082} & \cellcolor[HTML]{84C6EB} \textbf{0.986} & \cellcolor[HTML]{E8A0BF} \textbf{1.387} \\
Instruct & \cellcolor[HTML]{FFFFFF} 0.483 & \cellcolor[HTML]{FFFFFF} 0.200 & \cellcolor[HTML]{FFFFFF} 0.196 & \cellcolor[HTML]{FFFFFF} 0.102 & \cellcolor[HTML]{FFFFFF} 0.151 & \cellcolor[HTML]{D9EDF9} 16.136 & \cellcolor[HTML]{E2F2FA} 0.757 & \cellcolor[HTML]{FAEAF1} 0.331 & \cellcolor[HTML]{BDE1F4} 0.908 & \cellcolor[HTML]{ECB2CB} 14.432 \\
HiDream & \cellcolor[HTML]{E8F4FB} 0.560 & \cellcolor[HTML]{A0D3F0} 0.734 & \cellcolor[HTML]{9DD2EF} 0.752 & \cellcolor[HTML]{BEE1F4} 0.333 & \cellcolor[HTML]{B0DAF2} 0.606 & \cellcolor[HTML]{FFFFFF} 12.472 & \cellcolor[HTML]{FFFFFF} 0.703 & \cellcolor[HTML]{FFFFFF} 0.400 & \cellcolor[HTML]{FFFFFF} 0.818 & \cellcolor[HTML]{FFFFFF} 71.783 \\
JoyAI & \cellcolor[HTML]{C2E3F5} 0.689 & \cellcolor[HTML]{96CEEE} 0.789 & \cellcolor[HTML]{94CEEE} 0.803 & \cellcolor[HTML]{A3D4F0} 0.430 & \cellcolor[HTML]{93CDEE} 0.769 & \cellcolor[HTML]{CEE8F7} 17.198 & \cellcolor[HTML]{D3EAF8} 0.787 & \cellcolor[HTML]{F0C2D6} 0.197 & \cellcolor[HTML]{8FCBED} 0.972 & \cellcolor[HTML]{E9A4C2} 4.147 \\
\midrule
\multicolumn{11}{l}{\textbf{Local} ($n=3{,}506$)} \\
\midrule
Step1X & \cellcolor[HTML]{A1D3F0} 0.702 & \cellcolor[HTML]{CAE6F6} 0.455 & \cellcolor[HTML]{C0E2F5} 0.515 & \cellcolor[HTML]{DCEFF9} 0.196 & \cellcolor[HTML]{BFE1F5} 0.515 & \cellcolor[HTML]{B4DCF3} 14.432 & \cellcolor[HTML]{D6ECF8} 0.579 & \cellcolor[HTML]{FBEDF3} 0.507 & \cellcolor[HTML]{D6ECF8} 0.858 & \cellcolor[HTML]{F9E5EE} 64.343 \\
LongCat & \cellcolor[HTML]{89C8EC} 0.756 & \cellcolor[HTML]{94CDEE} 0.744 & \cellcolor[HTML]{8BC9EC} 0.801 & \cellcolor[HTML]{9FD3EF} 0.457 & \cellcolor[HTML]{8ECBED} 0.788 & \cellcolor[HTML]{B5DDF3} 14.412 & \cellcolor[HTML]{BBE0F4} 0.627 & \cellcolor[HTML]{ECB2CB} 0.342 & \cellcolor[HTML]{9AD0EF} 0.903 & \cellcolor[HTML]{EEB9D0} 30.766 \\
FireRed & \cellcolor[HTML]{94CDEE} 0.731 & \cellcolor[HTML]{91CCED} 0.762 & \cellcolor[HTML]{8AC9EC} 0.810 & \cellcolor[HTML]{9BD0EF} 0.476 & \cellcolor[HTML]{96CEEE} 0.744 & \cellcolor[HTML]{95CEEE} 15.913 & \cellcolor[HTML]{A1D4F0} 0.673 & \cellcolor[HTML]{E9A2C1} 0.299 & \cellcolor[HTML]{8CCAEC} 0.914 & \cellcolor[HTML]{EAA8C5} 18.007 \\
Qwen & \cellcolor[HTML]{84C6EB} \textbf{0.767} & \cellcolor[HTML]{84C6EB} \textbf{0.829} & \cellcolor[HTML]{84C6EB} \textbf{0.841} & \cellcolor[HTML]{84C6EB} \textbf{0.574} & \cellcolor[HTML]{84C6EB} \textbf{0.846} & \cellcolor[HTML]{84C6EB} \textbf{16.705} & \cellcolor[HTML]{84C6EB} \textbf{0.725} & \cellcolor[HTML]{E8A0BF} \textbf{0.293} & \cellcolor[HTML]{84C6EB} \textbf{0.921} & \cellcolor[HTML]{E8A0BF} \textbf{11.581} \\
Instruct & \cellcolor[HTML]{D5EBF8} 0.586 & \cellcolor[HTML]{FFFFFF} 0.172 & \cellcolor[HTML]{FFFFFF} 0.175 & \cellcolor[HTML]{FFFFFF} 0.044 & \cellcolor[HTML]{FFFFFF} 0.156 & \cellcolor[HTML]{D7ECF8} 12.828 & \cellcolor[HTML]{FFFFFF} 0.507 & \cellcolor[HTML]{FFFFFF} 0.556 & \cellcolor[HTML]{FFFFFF} 0.826 & \cellcolor[HTML]{FAEBF2} 68.935 \\
HiDream & \cellcolor[HTML]{FFFFFF} 0.491 & \cellcolor[HTML]{B2DBF2} 0.586 & \cellcolor[HTML]{A9D7F1} 0.640 & \cellcolor[HTML]{BFE1F5} 0.321 & \cellcolor[HTML]{B2DBF2} 0.589 & \cellcolor[HTML]{FFFFFF} 10.927 & \cellcolor[HTML]{F4FAFD} 0.526 & \cellcolor[HTML]{FAEAF1} 0.497 & \cellcolor[HTML]{FAFDFE} 0.830 & \cellcolor[HTML]{FFFFFF} 83.785 \\
JoyAI & \cellcolor[HTML]{A2D4F0} 0.701 & \cellcolor[HTML]{94CDEE} 0.743 & \cellcolor[HTML]{8DCAEC} 0.794 & \cellcolor[HTML]{9CD1EF} 0.472 & \cellcolor[HTML]{91CCED} 0.774 & \cellcolor[HTML]{C6E4F6} 13.617 & \cellcolor[HTML]{EAF5FC} 0.544 & \cellcolor[HTML]{F0C3D6} 0.389 & \cellcolor[HTML]{9AD0EF} 0.904 & \cellcolor[HTML]{EEB9D0} 30.296 \\
\bottomrule
\end{tabularx}
% 中文：VDE Bench上的主要评测结果。
% 每个模型在两种设置下均覆盖同一批样本。
% 除 LPIPS 和 FID 越低越好（↓）外，其余指标均越高越好。
% 蓝色与粉色背景表示编辑模型中较好的分数；加粗标记每列最佳的编辑模型。
\caption{\textbf{Main evaluation results on VDE Bench.} Each model covers the same samples in both settings. Higher is better except LPIPS and FID ($\downarrow$). Blue and pink shading indicate better editing-model scores; bold marks the best editing model per column.}
\label{tab:overall_eval}

\end{table*}

\begin{table*}[!t]
\centering
\fontsize{8}{9.5}\selectfont
\renewcommand{\arraystretch}{1.12}
\setlength{\tabcolsep}{3pt}
\begin{tabularx}{\textwidth}{@{}l*{5}{>{\centering\arraybackslash}X}*{5}{>{\centering\arraybackslash}X}@{}}
\toprule
& \multicolumn{5}{c}{\textbf{English} ($n=2{,}602$)} & \multicolumn{5}{c}{\textbf{Chinese} ($n=728$)} \\
\cmidrule(lr){2-6}\cmidrule(lr){7-11}
Model & IoU & CDM & BLEU & TEDS & Avg. & IoU & CDM & BLEU & TEDS & Avg. \\
\midrule
Step1X & \cellcolor[HTML]{6EC0B8} 0.767 & \cellcolor[HTML]{94D1CB} 0.609 & \cellcolor[HTML]{BBE2DE} 0.303 & \cellcolor[HTML]{9AD4CE} 0.550 & \cellcolor[HTML]{F2CF9B} 0.557 & \cellcolor[HTML]{76C4BC} 0.762 & \cellcolor[HTML]{AFDDD8} 0.475 & \cellcolor[HTML]{E1F2F0} 0.103 & \cellcolor[HTML]{B4DEDA} 0.460 & \cellcolor[HTML]{F4D5A6} 0.450 \\
LongCat & \cellcolor[HTML]{5CB8AF} 0.801 & \cellcolor[HTML]{64BCB3} 0.798 & \cellcolor[HTML]{7CC7BF} 0.519 & \cellcolor[HTML]{62BBB2} 0.772 & \cellcolor[HTML]{EBB665} 0.723 & \cellcolor[HTML]{55B6AC} 0.851 & \cellcolor[HTML]{5FBAB1} 0.786 & \cellcolor[HTML]{8DCEC7} 0.192 & \cellcolor[HTML]{5CB9AF} 0.838 & \cellcolor[HTML]{EAB15B} 0.667 \\
FireRed & \cellcolor[HTML]{6BBFB7} 0.771 & \cellcolor[HTML]{5AB8AE} 0.838 & \cellcolor[HTML]{6ABFB6} 0.581 & \cellcolor[HTML]{66BDB4} 0.756 & \cellcolor[HTML]{EBB461} 0.737 & \cellcolor[HTML]{54B5AB} 0.855 & \cellcolor[HTML]{53B5AB} 0.830 & \cellcolor[HTML]{7DC7BF} 0.209 & \cellcolor[HTML]{5FBAB1} 0.827 & \cellcolor[HTML]{EAAF56} 0.680 \\
Qwen & \cellcolor[HTML]{4FB3A9} \textbf{0.824} & \cellcolor[HTML]{4FB3A9} \textbf{0.880} & \cellcolor[HTML]{4FB3A9} \textbf{0.676} & \cellcolor[HTML]{4FB3A9} \textbf{0.848} & \cellcolor[HTML]{E8A94A} \textbf{0.807} & \cellcolor[HTML]{4FB3A9} \textbf{0.867} & \cellcolor[HTML]{4FB3A9} \textbf{0.846} & \cellcolor[HTML]{4FB3A9} \textbf{0.258} & \cellcolor[HTML]{4FB3A9} \textbf{0.895} & \cellcolor[HTML]{E8A94A} \textbf{0.716} \\
Instruct & \cellcolor[HTML]{CFEAE8} 0.584 & \cellcolor[HTML]{FFFFFF} 0.187 & \cellcolor[HTML]{FFFFFF} 0.069 & \cellcolor[HTML]{FFFFFF} 0.152 & \cellcolor[HTML]{FFFFFF} 0.248 & \cellcolor[HTML]{FFFFFF} 0.399 & \cellcolor[HTML]{FFFFFF} 0.167 & \cellcolor[HTML]{FFFFFF} 0.072 & \cellcolor[HTML]{FFFFFF} 0.135 & \cellcolor[HTML]{FFFFFF} 0.193 \\
HiDream & \cellcolor[HTML]{FFFFFF} 0.494 & \cellcolor[HTML]{7FC8C1} 0.690 & \cellcolor[HTML]{A1D6D1} 0.394 & \cellcolor[HTML]{92D0CA} 0.584 & \cellcolor[HTML]{F3D2A0} 0.541 & \cellcolor[HTML]{AFDCD8} 0.611 & \cellcolor[HTML]{71C2BA} 0.713 & \cellcolor[HTML]{A4D8D2} 0.168 & \cellcolor[HTML]{84CAC3} 0.667 & \cellcolor[HTML]{F0C687} 0.540 \\
JoyAI & \cellcolor[HTML]{97D2CC} 0.690 & \cellcolor[HTML]{63BCB3} 0.802 & \cellcolor[HTML]{75C3BC} 0.544 & \cellcolor[HTML]{66BDB4} 0.759 & \cellcolor[HTML]{ECBA6D} 0.699 & \cellcolor[HTML]{90CFC9} 0.695 & \cellcolor[HTML]{5EBAB0} 0.787 & \cellcolor[HTML]{8BCDC6} 0.194 & \cellcolor[HTML]{5DB9B0} 0.834 & \cellcolor[HTML]{ECB869} 0.628 \\
\bottomrule\end{tabularx}
% 中文：VDE Bench的分语言结果。每项分数为局部与全局结果的平均值，保留三位小数；Avg. 为四项 OCR 指标的均值。176 条混合语言样本未计入本表，但计入总体结果与按操作分组的结果。青绿色越深表示分数越高，琥珀色突出显示平均值，加粗表示各列最佳结果。
\caption{\textbf{Results by language on VDE Bench.} Each score averages the local and global settings and is reported to three decimal places; Avg.\ is the mean of the four OCR metrics. The 176 mixed-language samples are excluded here but included in the overall and per-operation results. Deeper teal indicates higher scores, amber highlights averages, and bold marks the best result in each column.}
\label{tab:language_results}
\end{table*}

\begin{figure}[t]
\centering
\begin{minipage}[t]{0.485\linewidth}
\vspace{0pt}
\captionsetup{font=footnotesize,skip=4pt}
\centering
\includegraphics[width=\linewidth]{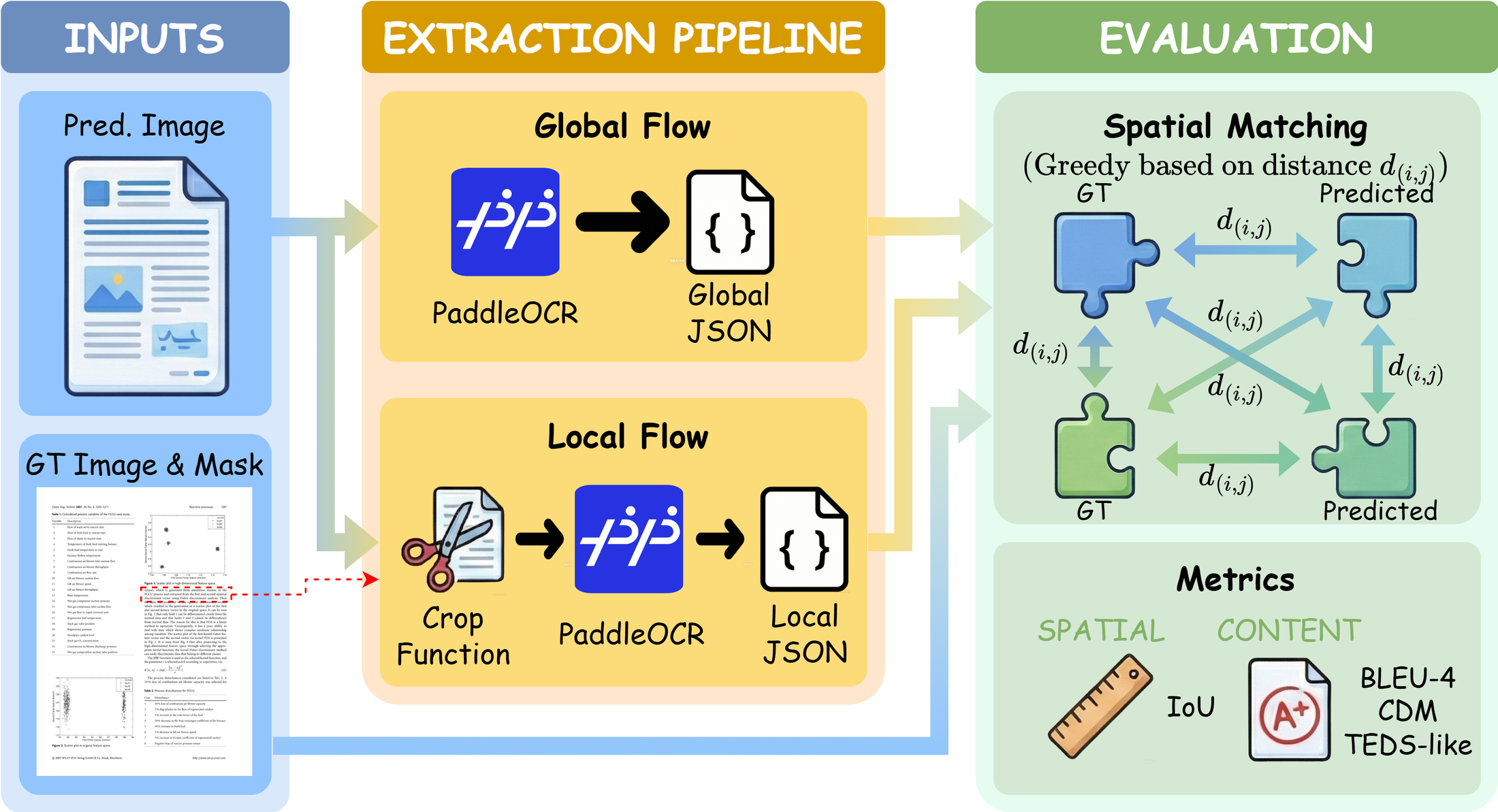}
% 中文：OCR 评测流程。每张生成页面按标注的编辑区域裁剪；PaddleOCR-VL 分别识别裁剪区域与整页，并将识别结果与对应的 GT 进行比较。
\caption{\textbf{OCR evaluation pipeline.} Each generated page is cropped to its annotated editing region. PaddleOCR-VL recognizes the crop and the full page separately, and each output is compared with its corresponding ground truth.}
\label{fig:eval_pipeline}
\end{minipage}\hfill
\begin{minipage}[t]{0.485\linewidth}
\vspace{0pt}
\captionsetup{font=footnotesize,skip=4pt}
\centering
\includegraphics[width=\linewidth]{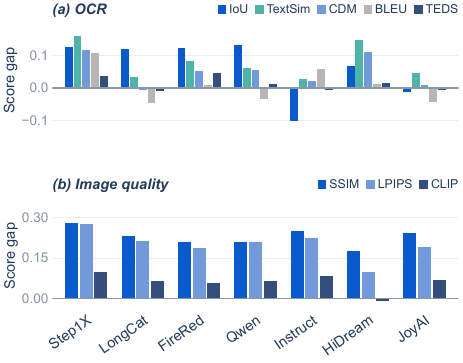}
% 中文：八项指标的局部与全局评测差异。(a) OCR：IoU、TextSim、CDM、BLEU 和 TEDS；(b) 图像质量：SSIM、LPIPS 和 CLIP。除 LPIPS 使用局部减全局外，Δ 均为全局减局部。
\caption{\textbf{Global--local gaps across eight metrics.} (a) OCR; (b) image quality. $\Delta=\text{global}-\text{local}$, except LPIPS uses local minus global.}
\label{fig:scale_gap}
\end{minipage}
\end{figure}

% 中文：图 X 概括了评测流程。我们评测七个基于指令的图像编辑模型：Step1X、Qwen-Image-Edit、LongCat-Image-Edit、InstructPix2Pix、FireRed-Image-Edit、HiDream-O1-Image 与 JoyAI-Image-Edit-Plus，下文沿用表中的简称。局部评测将生成页面裁剪至标注的编辑区域，全局评测则对整页打分。所有模型均在同一组固定的 3,506 条样本上评测，模型版本与生成配置保持固定。
Figure~\ref{fig:eval_pipeline} summarizes the evaluation pipeline. We evaluate seven instruction-based editing models: Step1X~\cite{liu2025step1x-edit}, Qwen-Image-Edit~\cite{wu2025qwenimagetechnicalreport}, LongCat-Image-Edit~\cite{LongCat-Image}, InstructPix2Pix~\cite{brooks2023instructpix2pix}, FireRed-Image-Edit~\cite{superintelligenceteam2026fireredimageedit10technicalreport}, HiDream-O1-Image~\cite{cai2026hidream}, and JoyAI-Image-Edit-Plus~\cite{song2026joy}, referred to below by the short names used in the tables. Local evaluation crops each generated page to its annotated editing region, whereas global evaluation scores the full page. All models are evaluated on the same frozen set of 3,506 samples with fixed model revisions and generation configurations.

% 中文：总体表现。Qwen 在两种设置下的全部 OCR 与图像指标上均取得最佳成绩，但其领先幅度因指标类别而差异悬殊：在局部设置下，其 CDM、BLEU 与 TEDS 分别领先次优模型 0.031、0.098 与 0.058，而 LPIPS 仅比 FireRed 低 0.006。感知相似度几乎无法拉开两者差距，文本与结构指标却揭示了明显差异。在排名靠后的模型之间，不同指标类别的分歧更为明显：LongCat 的全局 IoU 高于 FireRed（0.876 对 0.856），CDM 却更低（0.794 对 0.862）；HiDream 的局部 CDM 高于 Step1X（0.640 对 0.515），IoU 却更低（0.491 对 0.702）。可见，"把文字放对位置"与"把文字写对"是两种不同的能力，任何单一指标都不足以对模型排序。

\paragraph{Overall performance.}
Qwen achieves the best score on every reported OCR and image metric in both settings (Table~\ref{tab:overall_eval}). Its margins, however, vary sharply across metric families. Locally, it surpasses the runner-up by 0.031 in CDM, 0.098 in BLEU, and 0.058 in TEDS, yet its LPIPS is only 0.006 lower than FireRed's. Perceptual similarity thus barely distinguishes the two models, whereas textual and structural metrics reveal a clear gap. Below the top rank, the metric families diverge further. LongCat attains higher global IoU than FireRed (0.876 vs.\ 0.856) but lower CDM (0.794 vs.\ 0.862), and HiDream attains higher local CDM than Step1X (0.640 vs.\ 0.515) but lower IoU (0.491 vs.\ 0.702). Placing text in the right region and writing the right text are therefore distinct capabilities, and no single metric suffices to rank models.

% 中文：局部与全局。从整页切换到编辑区域后，Qwen 的 SSIM 从 0.936 降至 0.725，IoU 从 0.901 降至 0.767。整页图像以未编辑内容为主，因此较高的页面级相似度可能与集中在目标区域的错误并存。七个模型中，有五个的局部 IoU 更低、六个的局部 CDM 更低（见局部与全局差异图）。Step1X 和 HiDream 的 CDM 差距最大，分别为 0.119 和 0.112。不过，这一趋势并非在所有指标上一致：Qwen 的局部 BLEU（0.574）略高于全局（0.541）。不过值得注意的是，由于在当前 BLEU-4 实现下，GT 自比较在两个尺度上的上限本身有所不同（局部 0.810，全局 0.648），因此局部BLEU指标其实反而要更差一些。
% 中文：3506条样本的操作分组结果，使用单栏矢量柱状图。
\Needspace{0.50\textheight}
\begin{wrapfigure}{r}{0.45\textwidth}
\vspace{-8pt}
\captionsetup{font=footnotesize,skip=4pt}
\centering
\includegraphics[width=\linewidth]{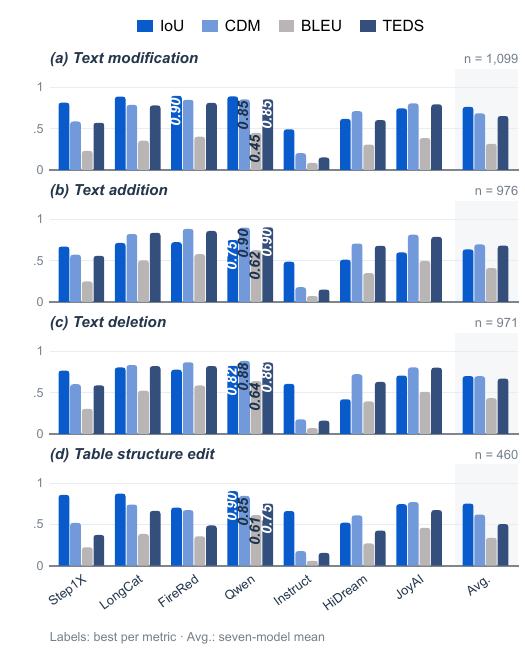}
% 中文：3,506 条样本上按编辑类型划分的表现。
% 分数由分别汇总的全局与局部结果取平均得到。
% Modification 合并替换与修改；添加和删除均包含文本区域与表格区域。
% 颜色区分不同指标；标签标出每种指标、每项操作下表现最佳的模型。
% Avg. 为七种模型的平均值。
% 标签保留两位小数。
\caption{\textbf{Performance by edit type on 3,506 samples.} Scores average the separately aggregated global and local results. Modification combines replacement and modification; addition and deletion include text and table regions. Colors distinguish metrics; labels mark the best model for each metric and operation. Avg. is the seven-model mean. Labels are rounded to two decimal places.}
\label{fig:edit_type_all}
\end{wrapfigure}

\paragraph{Local versus global.}
Moving from the full page to the editing region, Qwen's SSIM drops from 0.936 to 0.725 and its IoU from 0.901 to 0.767. Because a full page is dominated by unedited content, high page-level similarity can coexist with errors concentrated in the requested region. Across the seven models, local scores are lower for five on IoU and six on CDM (Figure~\ref{fig:scale_gap}). Step1X and HiDream exhibit the largest CDM gaps, at 0.119 and 0.112. The trend is not uniform across metrics, however: Qwen's BLEU is slightly higher locally (0.574) than globally (0.541). Ground-truth self-comparison under our BLEU-4 implementation yields different ceilings at the two scales (0.810 locally vs.\ 0.648 globally). Relative to these respective ceilings, local BLEU performance is lower.

% 中文：图中显示OCR的全局评估并没有显著优于局部评估，相当一部分 OCR 指标在局部更差，而七个模型的局部图像指标除了LPIPS外均更低，这表明细粒度编辑区域在保持视觉保真度方面仍然更具挑战性。
Figure~\ref{fig:scale_gap} shows that global OCR evaluation does not substantially outperform
local evaluation, although a considerable fraction of OCR-based metrics are worse locally.
In contrast, all seven models score lower on every local image-quality metric except LPIPS,
suggesting that preserving visual fidelity remains more challenging in fine-grained edited regions.

% 中文：分操作分析。图 X 给出各操作在局部与全局设置下的平均分数，两种设置的单独结果见附录。添加操作最难定位：其局部模型平均 IoU 仅为 0.535，明显低于修改（0.778）与删除（0.660），但其 CDM（0.671）与修改（0.697）相近。这表明模型大多能生成所需文字，却难以将其放到正确位置，可能是因为添加操作缺少可供锚定的原有内容。结构编辑则呈现相反的失败模式：其局部平均 IoU 达到 0.772，CDM 与 TEDS 却仅为 0.524 与 0.499。也就是说，表格可以保持原有的占位范围，而单元格内容与组织结构仍是错误的。即便是 Qwen，其结构编辑 CDM 也从全局 0.925 降至局部 0.768。此外，总体领先并不意味着在每种操作上都领先：在修改操作上，FireRed 的局部 IoU（0.899）与 CDM（0.886）最高，Qwen 则在 BLEU 与 TEDS 上领先。
\paragraph{Operation-specific challenges.}
Figure~\ref{fig:edit_type_all} reports scores averaged over the local and global settings for each operation; Appendix~\ref{sec:appendix instruction types analysis} reports the two settings separately. Addition is the hardest operation to localize. Its local model-average IoU is 0.535, well below modification (0.778) and deletion (0.660), even though its CDM (0.671) approaches that of modification (0.697). Models thus largely generate the requested text but struggle to place it correctly, plausibly because addition lacks existing content to anchor the edit. Structure editing exhibits the opposite failure mode. Its local mean IoU reaches 0.772, yet CDM and TEDS are only 0.524 and 0.499: a table can preserve its footprint while its cell contents and organization are wrong. Even Qwen's structure-editing CDM falls from 0.925 globally to 0.768 locally. Finally, overall leadership does not hold uniformly across operations. On modification, FireRed achieves the highest local IoU (0.899) and CDM (0.886), whereas Qwen leads in BLEU and TEDS.

% 中文：分语言表现。Qwen 在两个语言子集上均领先，但语言差距的方向与大小随指标而变。英文 BLEU 远高于中文（0.676 对 0.258），CDM 的差距却很小（0.880 对 0.846）；在 IoU（0.867 对 0.824）与 TEDS（0.895 对 0.848）上，中文甚至高于英文。
\paragraph{Language-specific performance.}
Qwen leads both language subsets (Table~\ref{tab:language_results}), but the direction and magnitude of the language gap depend on the metric. English BLEU far exceeds Chinese BLEU (0.676 vs.\ 0.258), whereas the CDM gap is small (0.880 vs.\ 0.846), and Chinese even surpasses English in IoU (0.867 vs.\ 0.824) and TEDS (0.895 vs.\ 0.848).

% 中文：评测可靠性。基于 1,000 张抽样图像的人工评审，七个模型在框对齐与文本正确性上的人工和自动排名均高度一致（两项 Spearman ρ 均为 0.964）；对 200 张生成图像的 OCR 结果进行人工检查，字符准确率约为 99.7%。同时，我们还通过Hunyuan-OCR进行测评，结果表明，正文主表的十组指标排名与其高度相关（平均 Spearman ρ = 0.971），Qwen 仍在九组指标中领先。这说明了主要排名结论对 OCR 模型选择具有稳健性，详细覆盖范围与名次变化见 OCR 对照附录。
\paragraph{Reliability of evaluation.}
Human review of 1,000 sampled images yields seven-model rankings that closely agree with automated box-alignment and text-correctness rankings (both Spearman's $\rho=0.964$; Appendix~\ref{sec:human eval}), and manual inspection of OCR outputs on 200 generated images finds approximately 99.7\% character accuracy (Appendix~\ref{sec:ocr_engine_check}). We also evaluate with Hunyuan-OCR. Its model rankings correlate closely with those in the main table across ten metric-setting combinations (mean Spearman's $\rho=0.971$), with Qwen remaining first in nine. These results indicate that the main ranking conclusions are robust to the choice of OCR model; coverage and individual rank changes are detailed in Appendix~\ref{sec:ocr_engine_check}.
% 中文：复杂文档编辑要求模型根据指令精确改写目标文本，同时保持页面其余内容不变。为此，我们提出 VDE Bench，一个包含 3,506 条高质量、人工精心标注筛选的中英文文档编辑样本的评测基准。每条样本均配有参考结果、编辑区域标注与结构化 OCR 输出，使"改了什么、改在哪里、渲染得如何"能够在局部与整页两个尺度上分别得到评估。
\section{Conclusion}
\label{sec: conclusion}
Editing complex documents requires models to precisely modify the target text according to instructions while leaving the rest of the page unchanged. To this end, we introduce VDE Bench, a benchmark comprising 3,506 high-quality Chinese and English document-editing samples that have been carefully annotated and curated by humans. Each sample includes a reference result, editing-region annotations, and structured OCR outputs, enabling separate assessment of what is edited, where, and how it is rendered at both local and full-page scales.

% 中文：正文结束后另起一页排版局限性。
% 中文标题：局限性
\section*{Limitations}
\label{sec:limitation}

% 中文：VDE Bench 主要覆盖中文和英文；其结果尚不能说明模型在韩文、日文等其他语言中的编辑表现。
VDE Bench covers primarily Chinese and English; its results do not establish editing performance in other languages, such as Korean or Japanese.

\bibliography{acl_references}
\clearpage
\appendix
% 中文标题：基准说明
\section{Benchmark Specification}

% 中文标题：数据集说明
\subsection{Dataset Card}
\label{sec:dataset card}

% 中文：数据集概述。
% 视觉文档编辑基准（Visual Document Editing Bench，VDE Bench）是一个综合性数据集，旨在大规模评估图像编辑模型直接编辑视觉内容丰富的文档的能力。
\textbf{DatasetSummary.} The Visual Document Editing (VDE) Bench is a comprehensive dataset designed to evaluate the direct editing capabilities of image editing models on visually rich documents at scale.

% 中文：构建动机。
% 现有图像编辑基准往往缺乏评估复杂文本场景表现所需的复杂性。
% 我们构建这一基准，以填补复杂文本文档编辑任务上的评测空白。
\textbf{Curation Rationale.} Existing image editing benchmarks often lack the complexity required to evaluate performance in intricate textual scenarios. This benchmark was established to bridge the gap in complex text document editing tasks.

% 中文：源数据。
% 种子文档来自 OmniDocBench、DocLayNet、PubMed Central（PMC）以及通过 DocGenome 获取的 arXiv 论文。
% 语料覆盖金融、法律、商业和教育等多个领域。
\textbf{Source Data.} Seed documents were collected from \textbf{OmniDocBench}~\cite{ouyang2024omnidocbenchbenchmarkingdiversepdf}, DocLayNet~\cite{pfitzmann2022doclaynet}, PubMed Central (PMC), and arXiv papers accessed through DocGenome. The corpus spans multiple domains, including finance, law, commerce, and education.

% 中文：许可与预期用途。
% VDE Bench 面向视觉文档编辑的非商业研究，其源页面来自多个数据集和论文集合。
% 其中，OmniDocBench 的版权声明限定其数据仅用于研究和非商业用途，脚注链接指向该版权声明。
% 对于 DocLayNet、DocGenome、PMC 和 arXiv 材料，应分别遵守适用的数据集条款和单篇文档许可。
% 源页面及包含这些材料的编辑后参考图像，仍受相应源数据条款和原始内容权利人的权利约束。
\textbf{Licensing and intended use.} VDE Bench is intended for non-commercial research on visual document editing. Its source pages are drawn from multiple datasets and article collections. OmniDocBench's copyright statement specifies research-only, non-commercial use for its material.\footnote{\href{https://github.com/opendatalab/OmniDocBench\#copyright-statement}{OmniDocBench copyright statement}.} DocLayNet, DocGenome, PMC, and arXiv materials are subject to their respective dataset terms and individual-document licenses. Source pages and edited reference images remain subject to the applicable source-data terms and the rights of the original content holders.

% 中文：不同资源的许可。
% OmniDocBench 软件仓库所显示的 Apache-2.0 许可证，不被视为对文档图像的统一授权。
% 评测代码、新创建的指令及区域标注的许可，与使用或再分发源材料所需的权限是分开的。
% 为这些原创贡献指定的任何许可证，都不会覆盖或取消底层第三方内容的限制。
% 再分发源图像或衍生图像时，必须遵守适用的授权、署名要求及源数据条款。
\textbf{Asset-specific licensing.} The Apache-2.0 license displayed for the OmniDocBench software repository is not treated as a blanket license for the document images. Licensing for evaluation code and newly created instructions and region annotations is separate from permission to use or redistribute the source material. Any license assigned to these original contributions does not override restrictions on underlying third-party content. Redistribution of source or derived images must respect the applicable permissions, attribution requirements, and source-data terms.

% 中文标题：期望性质
\subsection{Desired Properties}
\label{sec:properties}

% 中文：我们形式化定义视觉文档编辑（VDE）任务：系统需在密集文本文档上执行精确、受指令引导的修改，同时保持文本保真度与视觉完整性。
We formally define \textit{Visual Document Editing (VDE)}, a task requiring systems to perform precise, instruction-guided modifications on dense text documents while preserving both textual fidelity and visual integrity.\\

\begin{definition}[Dense Text Document]
% 中文：设 D = (p_1, p_2, …, p_M) 为包含 M 个页面的密集文本文档。
% 每个页面 p_i 包含视觉内容（布局、字体排印、间距和图形元素），以及文本内容 T(p_i)；后者表示由 OCR 系统识别得到的词元序列。
Let $D = (p_1, p_2, \dots, p_M)$ be a dense text document consisting of $M$ pages, where each page $p_i$ comprises visual content (layout, typography, spacing, graphical elements) and textual content $\mathcal{T}(p_i)$ representing its token sequence as recognized by an OCR system.
\end{definition}

\begin{definition}[Visual Document Editing]
% 中文：给定密集文本文档 D = (p_1, p_2, …, p_M) 和自然语言编辑请求序列 R = (r_1, r_2, …, r_M)，其中每个 r_i 指定页面 p_i 的编辑要求，该任务需要输出以下结果：
Given a dense text document $D = (p_1, p_2, \dots, p_M)$ and a sequence of natural language editing requests $\mathcal{R} = (r_1, r_2, \dots, r_M)$, where each $r_i$ specifies the editing instruction for page $p_i$, the task is to produce:
\begin{enumerate}
% 中文：编辑后的文档 D′ = (p′_1, p′_2, …, p′_M)，其中每个页面 p′_i = f(p_i, r_i) 都忠实反映 r_i 所指定的修改。
    \item An edited document $D' = (p'_1, p'_2, \dots, p'_M)$, where each page $p'_i = f(p_i, r_i)$ faithfully reflects the modification specified by $r_i$,
% 中文：对每个页面 p′_i，给出编辑区域 e_i，它属于该页面的区域集合，用于指明实际实施修改的精确位置。
    \item For each page $p'_i$, an edited region $e_i \subseteq \mathcal{R}(p'_i)$ identifying the precise location where the edit has been applied, and
% 中文：对每个页面 p′_i，在其编辑区域 e_i 上运行 OCR 系统，得到识别输出 O_i，以表示 e_i 内识别出的文本。
    \item For each page $p'_i$, an OCR recognition output $O_i$ obtained by executing an OCR system on the edited region $e_i$, representing the recognized text within $e_i$.
\end{enumerate}
\end{definition}

% 中文：编辑质量通过一组评测指标衡量，这些指标共同考察 OCR 层面的文本准确性与图像层面的视觉保真度，并体现为以下性质：
The quality of the editing is assessed through a suite of evaluation metrics that jointly measure both \textit{OCR-level textual accuracy} and \textit{image-level visual fidelity}, characterized by the following properties:

% 中文：性质 1：文本正确性。
% 每个编辑区域的 OCR 输出必须与相应编辑请求所指定的真实参考文本相匹配，即文本相似度应达到阈值 τ_text。
\paragraph{Property 1: Textual Correctness.} The OCR output of each edited region must match the ground-truth text specified by the corresponding editing request:
\begin{equation}
\forall\, i \in [M]:\; \textsc{TextSim}(O_i, O_i^{*}) \geq \tau_{\text{text}}
\end{equation}
% 中文：其中，O_i* 表示区域 e_i 的真实参考 OCR 输出，TextSim(·,·) 为文本相似度指标，例如归一化编辑距离或字符级 F1 分数。
where $O_i^{*}$ denotes the ground-truth OCR output for region $e_i$, and $\textsc{TextSim}(\cdot, \cdot)$ is a text similarity metric (e.g., normalized edit distance or character-level F1).

% 中文：性质 2：视觉保真度。
% 编辑区域应与真实参考渲染在视觉上保持一致，保留字体风格、字号、颜色、对齐方式和背景，即视觉相似度应达到阈值 τ_vis。
\paragraph{Property 2: Visual Fidelity.} The edited region must be visually consistent with the ground-truth rendering, preserving font style, size, color, alignment, and background:
\begin{equation}
\forall\, i \in [M]:\; \textsc{VisualSim}(e_i, e_i^{*}) \geq \tau_{\text{vis}}
\end{equation}
% 中文：其中，e_i* 是真实参考编辑区域，VisualSim(·,·) 刻画像素级或感知层面的相似性，例如 SSIM、LPIPS。
where $e_i^{*}$ is the ground-truth edited region and $\textsc{VisualSim}(\cdot, \cdot)$ captures pixel-level or perceptual similarity (e.g., SSIM, LPIPS).

% 中文：性质 3：全局文档保持。
% 每个页面中编辑区域以外的部分必须保持不变，确保编辑操作不会引入非预期伪影。
\paragraph{Property 3: Global Document Preservation.} Regions outside the edited area on each page must remain unaltered, ensuring that the editing operation does not introduce unintended artifacts:
\begin{equation}
\forall\, i \in [M],\; \forall\, p \in p'_i \setminus e_i:\; p = p^{*}
\end{equation}
% 中文：其中，p* 表示真实参考文档中的对应区域；这一条件要求未编辑内容得到完整保持。
where $p^{*}$ denotes the corresponding region in the ground-truth document, enforcing that non-edited content is perfectly preserved.

% 中文：性质 4：空间精确性。
% 预测编辑区域必须准确定位每个页面上的修改位置，即预测区域与真实参考区域之间的 IoU 应达到阈值 τ_loc。
\paragraph{Property 4: Spatial Precision.} The predicted edited region must accurately localize the area of modification on each page:
\begin{equation}
\forall\, i \in [M]:\; \textsc{IoU}(e_i, e_i^{*}) \geq \tau_{\text{loc}}
\end{equation}
% 中文：其中，e_i* 为页面 p_i 的真实参考编辑区域，IoU 衡量预测边界区域与真实参考边界区域之间的交并比。
where $e_i^{*}$ is the ground-truth edited region for page $p_i$ and $\textsc{IoU}$ measures the intersection-over-union between the predicted and ground-truth bounding region.

% 中文：性质 5：编辑完整性。
% 每个页面的编辑请求 r_i 都必须完整执行；遗漏任何页面级编辑都将受到惩罚。
\paragraph{Property 5: Edit Completeness.} The editing request $r_i$ for each page must be fully executed; omitting any page-level edit is penalized:
\begin{equation}
\forall\, i \in [M]:\; e_i \neq \emptyset \iff e_i^{*} \neq \emptyset
\end{equation}
% 中文：这可确保模型不会跳过任何页面上所要求的修改。
This ensures that the model does not skip any required modification on any page.

% 中文：最终，我们通过上述指标衡量模型生成输出 ({O_i}, {e_i}, D′) 与真实参考结果 ({O_i*}, {e_i*}, D′*) 之间的差异，从细粒度 OCR 准确性到整体视觉质量进行综合评估。
Ultimately, we assess the discrepancy between the model-generated outputs $(\{O_i\}, \{e_i\}, D')$ and the ground truth $(\{O_i^{*}\}, \{e_i^{*}\}, D'^{*})$ through the above metrics, providing a comprehensive evaluation that spans from fine-grained OCR accuracy to holistic visual quality.

% 中文标题：文档语料
\section{Document Corpus}
% 中文标题：文档类别
\subsection{Categories}

% 中文：数据集涵盖从正式出版物到日常文档的广泛类型。
% 既包括学术论文、书籍和报纸等结构清晰、格式严谨的内容，也包括演示文稿（PPT）、课堂笔记和会议记录等形式更灵活多样的材料，同时还包含专利、财务报告、法律法规、科学文章、政府招标文件与手册。
% 这种多来源、异构的构成，使模型能够在风格与复杂程度各异的真实场景中接受充分评估。
The dataset covers a wide range of document types, spanning from formal publications to everyday documents. It includes well-structured and rigorously formatted content such as academic papers, books, and newspapers, as well as more flexible and diverse formats like presentation slides (PPTs), class notes, and meeting notes, together with patents, financial reports, laws and regulations, scientific articles, government tenders, and manuals. This multi-source and heterogeneous composition enables models to be thoroughly evaluated across real-world scenarios with varying styles and levels of complexity.

% 中文：3,506 条样本版本包含 728 条中文、2,602 条英文和 176 条混合语言样本。
The 3,506-sample release contains 728 Chinese, 2,602 English, and 176 mixed-language samples.

% 中文标题：联合样本组成
\subsection{Joint Sample Composition}
\label{sec:joint_counts}
% 中文：联合计数表按语言、区域、操作和文档来源列出全部 3,506 条样本。
% 1,578 条文本区域编辑由 575 条修改、504 条添加和 499 条删除组成。
% 1,928 条表格区域编辑由 524 条修改、472 条添加、472 条删除和 460 条结构编辑组成。
Table~\ref{tab:joint_counts} enumerates all 3,506 samples by language, region, operation, and document source. The 1,578 text-region edits comprise 575 modifications, 504 additions, and 499 deletions. The 1,928 table-region edits comprise 524 modifications, 472 additions, 472 deletions, and 460 structural edits.

\begin{table*}[t]
\centering\scriptsize
\setlength{\tabcolsep}{1.1pt}
\renewcommand{\arraystretch}{1.12}
\begin{tabular*}{\textwidth}{@{\extracolsep{\fill}}l*{22}{r}@{}}
\toprule
Document category & \multicolumn{7}{c}{Chinese} & \multicolumn{7}{c}{English} & \multicolumn{7}{c}{Mixed} & Total \\
\cmidrule(lr){2-8}\cmidrule(lr){9-15}\cmidrule(lr){16-22}
 & \multicolumn{3}{c}{Text} & \multicolumn{4}{c}{Table} & \multicolumn{3}{c}{Text} & \multicolumn{4}{c}{Table} & \multicolumn{3}{c}{Text} & \multicolumn{4}{c}{Table} & \\
\cmidrule(lr){2-4}\cmidrule(lr){5-8}\cmidrule(lr){9-11}\cmidrule(lr){12-15}\cmidrule(lr){16-18}\cmidrule(lr){19-22}
 & M & A & D & M & A & D & S & M & A & D & M & A & D & S & M & A & D & M & A & D & S & \\
\midrule
Financial reports & 0 & 0 & 0 & 0 & 0 & 0 & 0 & 0 & 59 & 57 & 61 & 58 & 58 & 43 & 0 & 0 & 0 & 0 & 0 & 0 & 0 & 336 \\
Biomedical informatics & 20 & 20 & 20 & 19 & 19 & 19 & 19 & 20 & 20 & 20 & 19 & 19 & 19 & 19 & 0 & 0 & 0 & 0 & 0 & 0 & 0 & 272 \\
Social NLP \& fairness & 0 & 0 & 0 & 0 & 0 & 0 & 0 & 0 & 33 & 36 & 26 & 46 & 46 & 49 & 0 & 0 & 0 & 0 & 0 & 0 & 0 & 236 \\
Dialogue \& speech & 0 & 0 & 0 & 0 & 0 & 0 & 0 & 0 & 30 & 38 & 30 & 42 & 38 & 41 & 0 & 0 & 0 & 0 & 0 & 0 & 0 & 219 \\
Translation \& generation & 0 & 0 & 0 & 0 & 0 & 0 & 0 & 0 & 30 & 33 & 30 & 38 & 37 & 38 & 0 & 0 & 0 & 0 & 0 & 0 & 0 & 206 \\
Other academic sources & 0 & 0 & 0 & 0 & 0 & 0 & 0 & 60 & 27 & 26 & 31 & 17 & 18 & 15 & 0 & 0 & 0 & 0 & 0 & 0 & 0 & 194 \\
Text classification \& analysis & 0 & 0 & 0 & 0 & 0 & 0 & 0 & 0 & 31 & 33 & 31 & 32 & 33 & 32 & 0 & 0 & 0 & 0 & 0 & 0 & 0 & 192 \\
Slides & 61 & 16 & 12 & 7 & 1 & 2 & 2 & 58 & 12 & 9 & 4 & 0 & 0 & 0 & 0 & 0 & 0 & 0 & 0 & 0 & 0 & 184 \\
Knowledge \& retrieval & 0 & 0 & 0 & 0 & 0 & 0 & 0 & 0 & 29 & 29 & 29 & 31 & 29 & 31 & 0 & 0 & 0 & 0 & 0 & 0 & 0 & 178 \\
Books & 25 & 15 & 15 & 25 & 12 & 11 & 11 & 41 & 3 & 7 & 3 & 0 & 0 & 0 & 1 & 0 & 0 & 0 & 0 & 0 & 0 & 169 \\
Research reports & 45 & 19 & 13 & 49 & 8 & 9 & 8 & 0 & 0 & 0 & 0 & 0 & 0 & 0 & 0 & 0 & 0 & 0 & 0 & 0 & 0 & 151 \\
Others & 0 & 0 & 0 & 0 & 0 & 0 & 0 & 0 & 0 & 0 & 0 & 0 & 0 & 0 & 0 & 0 & 0 & 3 & 36 & 42 & 55 & 136 \\
Manuals & 0 & 0 & 0 & 0 & 0 & 0 & 0 & 0 & 15 & 22 & 21 & 20 & 21 & 18 & 0 & 0 & 0 & 0 & 0 & 0 & 0 & 117 \\
Linguistic analysis & 0 & 0 & 0 & 0 & 0 & 0 & 0 & 0 & 19 & 19 & 19 & 19 & 19 & 19 & 0 & 0 & 0 & 0 & 0 & 0 & 0 & 114 \\
Language models \& resources & 0 & 0 & 0 & 0 & 0 & 0 & 0 & 0 & 16 & 16 & 16 & 16 & 16 & 16 & 0 & 0 & 0 & 0 & 0 & 0 & 0 & 96 \\
Exams & 27 & 8 & 5 & 12 & 1 & 0 & 0 & 7 & 4 & 5 & 5 & 1 & 1 & 1 & 14 & 0 & 0 & 4 & 0 & 0 & 0 & 95 \\
Newspapers & 35 & 0 & 1 & 0 & 0 & 0 & 0 & 25 & 7 & 6 & 6 & 4 & 2 & 3 & 0 & 0 & 0 & 0 & 0 & 0 & 0 & 89 \\
Textbooks & 27 & 4 & 5 & 4 & 0 & 0 & 1 & 23 & 7 & 5 & 5 & 0 & 1 & 1 & 2 & 0 & 1 & 1 & 0 & 0 & 0 & 87 \\
Patents & 0 & 0 & 0 & 0 & 0 & 0 & 0 & 0 & 15 & 16 & 14 & 15 & 16 & 10 & 0 & 0 & 0 & 0 & 0 & 0 & 0 & 86 \\
Magazines & 25 & 6 & 2 & 2 & 0 & 0 & 1 & 22 & 13 & 3 & 0 & 0 & 0 & 0 & 0 & 0 & 0 & 0 & 0 & 0 & 0 & 74 \\
Scientific articles & 0 & 0 & 0 & 0 & 0 & 0 & 0 & 0 & 13 & 13 & 12 & 12 & 12 & 10 & 0 & 0 & 0 & 0 & 0 & 0 & 0 & 72 \\
Government tenders & 0 & 0 & 0 & 0 & 0 & 0 & 0 & 0 & 9 & 14 & 14 & 14 & 12 & 7 & 0 & 0 & 0 & 0 & 0 & 0 & 0 & 70 \\
Notes & 23 & 9 & 4 & 10 & 0 & 0 & 0 & 0 & 0 & 0 & 0 & 0 & 0 & 0 & 10 & 4 & 3 & 0 & 0 & 0 & 0 & 63 \\
Laws \& regulations & 0 & 0 & 0 & 0 & 0 & 0 & 0 & 0 & 7 & 7 & 8 & 7 & 7 & 6 & 0 & 0 & 0 & 0 & 0 & 0 & 0 & 42 \\
Scientific computing & 2 & 2 & 2 & 2 & 2 & 2 & 2 & 2 & 2 & 2 & 2 & 2 & 2 & 2 & 0 & 0 & 0 & 0 & 0 & 0 & 0 & 28 \\
\midrule
\textbf{Total} & 290 & 99 & 79 & 130 & 43 & 43 & 44 & 258 & 401 & 416 & 386 & 393 & 387 & 361 & 27 & 4 & 4 & 8 & 36 & 42 & 55 & \textbf{3506} \\
\bottomrule
\end{tabular*}
% 中文：全部 3,506 条样本按文档类别、语言、区域与操作划分的联合计数。
% 文档类别与图中的数据分布保持一致，学术文献拆分为 10 个子类，共计 25 类；行顺序按样本数递减排列。
% M、A、D 和 S 分别表示修改、添加、删除与结构编辑；修改合并文本替换和表格文本修改。
% Mixed 表示中英文混合文档；Others 仅表示未指定文档来源。
\caption{Joint counts for all 3,506 samples by document category, language, region, and operation. Categories follow Figure~\ref{fig:dataset_statistics}, with academic literature split into ten subcategories, and are ordered by sample count. M/A/D/S denote modification/addition/deletion/structure; modification combines text replacement and table text modification. Mixed denotes Chinese--English documents; Others refers only to the document-source label.}
\label{tab:joint_counts}
\end{table*}

% 中文标题：布局元素分布可视化
\subsection{Layout Element Distribution Visualization}

% 中文标题：布局元素分布分析
\subsubsection{Layout Element Distribution Analysis}
\label{sec:layout element distribution analysis}

% 中文：为刻画不同异构来源文档的结构组成，我们采用归一化热力图表示，对布局元素在空间中的分布程度进行可视化。
To characterize the structural composition of documents across heterogeneous sources, we visualize the spatial prevalence of layout elements through a normalized heatmap representation.

% 中文：对于 GT 页面布局密度图，我们使用全部 3,506 张编辑后 GT 页面的现有 PaddleOCR-VL 布局结果，覆盖 23 种检测元素类别和 25 个文档类别，分类与数据分布图一致。
% 每条样本对应一张 GT 页面，边界框和页面尺寸使用同一参考坐标系。
% 下文的元素组成与综合统计复用相同的 GT 布局框，覆盖 45,499 个元素。
For Figure~\ref{fig:Layout element density across document types}, we use the existing PaddleOCR-VL layouts of all 3,506 edited GT pages, covering 23 detected element categories and the same 25 document categories as Figure~\ref{fig:dataset_statistics}. Each sample contributes one GT page; bounding boxes and page dimensions use the same reference coordinate system. The element-composition and comprehensive statistics below use the same GT layouts, containing 45,499 detected elements.

% 中文：面积占比密度。
% 基于计数的指标无论元素占据多大空间，都对每个实例赋予相同权重；我们采用面积占比密度来量化元素的分布程度。
% 具体而言，对于文档类别 c 与元素类型 e，定义如下：
\paragraph{Area-Ratio Density.}
Unlike count-based metrics that assign equal weight to all element instances regardless of their spatial extent, we quantify element prevalence through an \emph{area-ratio density} formulation. Specifically, for a document category~$c$ and element type~$e$, we define:
\begin{equation}
    d_{c,e} \;=\; \frac{1}{\lvert P_c \rvert} \sum_{p \,\in\, P_c} \frac{\displaystyle\sum_{k \,\in\, E_{p,e}} \operatorname{area}(k)}{\operatorname{area}(p)},
\end{equation}
% 中文：其中，P_c 为属于类别 c 的页面集合，E_{p,e} 收集页面 p 上类型 e 的全部元素实例。
% area(k) 表示 GT 页面上元素 k 的边界框面积；area(p)=w_p×h_p 为页面总面积。
% 该密度刻画同类元素总面积相对于页面面积的平均占比。
% 不含类型 e 元素的页面以零计入。
where $P_c$ is the set of pages belonging to category~$c$, $E_{p,e}$ collects all instances of element type~$e$ on page~$p$, $\operatorname{area}(k)$ denotes the bounding-box area of element~$k$ on the GT page, and $\operatorname{area}(p) = w_p \times h_p$ is the total page area. The resulting density captures the mean summed element area relative to page area. Pages without element type~$e$ contribute zero.

% 中文：按元素类型进行 z-score 标准化。
% 不同元素类型的基础面积占比差异很大，例如文本块通常占据页面大部分空间，而脚注只占边缘的一小部分。
% 为便于跨元素比较，我们对密度矩阵的每一列分别标准化。
% 对于元素类型 e，计算如下：
\paragraph{Per-Element Z-Score Normalization.}
Different element types exhibit vastly different baseline area occupancies---e.g., text blocks typically dominate page space while footnotes occupy marginal regions. To facilitate cross-element comparison, we standardize each column of the density matrix independently. For element type~$e$:
\begin{equation}
    \begin{aligned}
 z_{c,e} &= \frac{d_{c,e} - \mu_e}{\sigma_e},\\
 \mu_e &= \frac{1}{\lvert C \rvert} \sum_{c \,\in\, C} d_{c,e}.
\end{aligned}
\end{equation}
% 中文：其中，σ_e 为 d_{·,e} 在各文档来源组之间的总体标准差，各组权重相同。
% 标准化后，每种元素类型均以零为中心、方差为一，从而能够直接识别哪些文档类别对特定结构组成分配了异常偏大或偏小的页面面积。
where $\sigma_e$ is the population standard deviation of $d_{\cdot,e}$ across document-source groups, with each group weighted equally. After normalization, each element type is centered at zero with unit variance, enabling direct identification of categories that devote anomalously large or small page area to specific structural components.

% 中文：综合布局统计。
% 我们使用六项互补指标刻画每个文档类别的布局复杂程度。
% 设 P_s 为属于数据来源 s 的页面集合，a(e,p) 和 A(p) 分别表示页面 p 上元素 e 的面积与该页面的总面积。
\paragraph{Comprehensive Layout Statistics.}
We characterize the layout complexity of each document category using six complementary metrics. Let $\mathcal{P}_s$ denote the set of pages belonging to data source $s$, and let $a(e, p)$ and $A(p)$ represent the area of element $e$ on page $p$ and the total page area, respectively.

% 中文：平均面积密度衡量布局元素占据的页面面积比例的平均值：
\textbf{Average Area Density} measures the mean fraction of page area occupied by layout elements:
\begin{equation}
    \text{Density}(s) = \frac{1}{|\mathcal{P}_s|} \sum_{p \in \mathcal{P}_s} \sum_{e \in p} \frac{a(e, p)}{A(p)}
\end{equation}

% 中文：复杂度衡量每页不同元素类型数量的平均值：
\textbf{Complexity} captures the average number of distinct element types per page:
\begin{equation}
    \text{Complexity}(s) = \frac{1}{|\mathcal{P}_s|} \sum_{p \in \mathcal{P}_s} |\{t : n_t(p) > 0\}|
\end{equation}
% 中文：其中，n_t(p) 为页面 p 上类型 t 的元素数量。
where $n_t(p)$ is the count of element type $t$ on page $p$.

% 中文：丰富度衡量每页布局元素总数量的平均值：
\textbf{Richness} measures the average total number of layout elements per page:
\begin{equation}
    \text{Richness}(s) = \frac{1}{|\mathcal{P}_s|} \sum_{p \in \mathcal{P}_s} \sum_{t} n_t(p)
\end{equation}

% 中文：熵通过香农熵量化元素类型分布的均匀程度：
\textbf{Entropy} quantifies the uniformity of element type distribution via Shannon entropy:
\begin{equation}
    \begin{aligned}
H(s) &= -\sum_{t} p(t|s) \log_2 p(t|s),\\
p(t|s) &= \frac{\sum_{p \in \mathcal{P}_s} n_t(p)}{\sum_{p \in \mathcal{P}_s} \sum_{t'} n_{t'}(p)}.
\end{aligned}
\end{equation}

% 中文：文本占比与视觉占比将页面中的元素面积划分为文本组成和视觉组成：
\textbf{Text Ratio} and \textbf{Visual Ratio} decompose the page area into text and visual components:
\begin{equation}
    \begin{aligned}
R_{\text{text}}(s) &= \frac{\bar{A}_{\text{text}}(s)}{\bar{A}_{\text{text}}(s) + \bar{A}_{\text{vis}}(s)},\\
    R_{\text{vis}}(s) &= \frac{\bar{A}_{\text{vis}}(s)}{\bar{A}_{\text{text}}(s) + \bar{A}_{\text{vis}}(s)}.
\end{aligned}
\end{equation}
% 中文：其中，Ā_text(s) 与 Ā_vis(s) 分别为文本元素和视觉元素平均每页占据的面积。
where $\bar{A}_{\text{text}}(s)$ and $\bar{A}_{\text{vis}}(s)$ are the mean page-normalized areas of text and visual elements. Visual elements include tables, images, charts, and header/footer images; the remaining categories form the text group.

% 中文：为了进行可视化，所有指标均按列归一化至 [0,1] 区间。
All metrics are column-wise normalized to $[0, 1]$ for visualization.

\begin{figure*}[t]
\centering
\includegraphics[width=\linewidth]{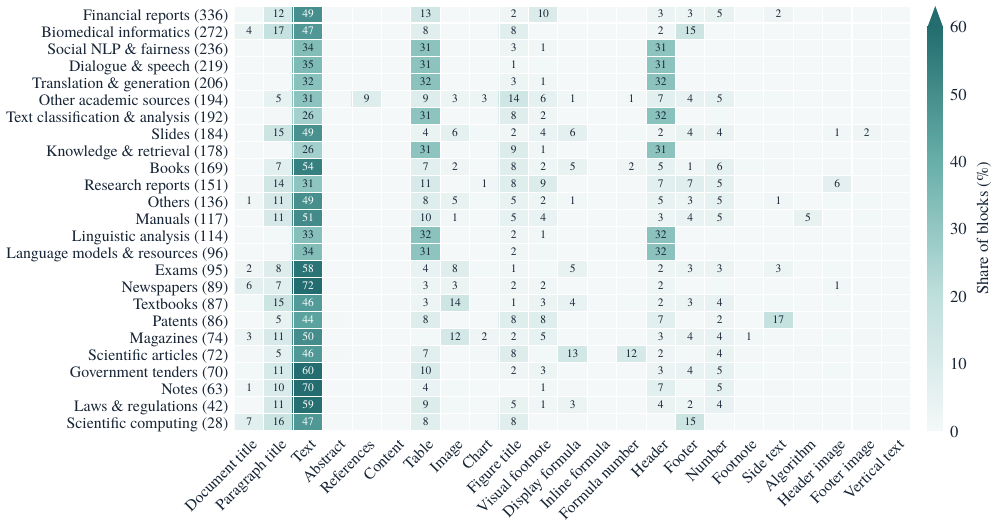}
% 中文：全部 3,506 张编辑后 GT 页面的布局元素组成，覆盖与数据分布图一致的 25 个文档类别和 23 类元素。
% 每行按该来源的元素总数归一化；括号标明页面数，单元格标出不低于 1% 的占比并取整。
% 报纸的文本块占比最高；各学术子类分别汇总，保留其不同的元素组成。
\caption{\textbf{Layout-element composition on all 3,506 edited GT pages.} Rows follow the 25 document categories in Figure~\ref{fig:dataset_statistics}, pool detected blocks within each category, and sum to 100\%; parentheses give page counts. Cell labels show rounded shares of at least 1\%. Newspapers have the highest text-block share; academic subcategories are aggregated separately to show their distinct compositions.}
\label{fig:element_composition}
\end{figure*}

\begin{figure*}[t]
\centering
\includegraphics[width=\linewidth]{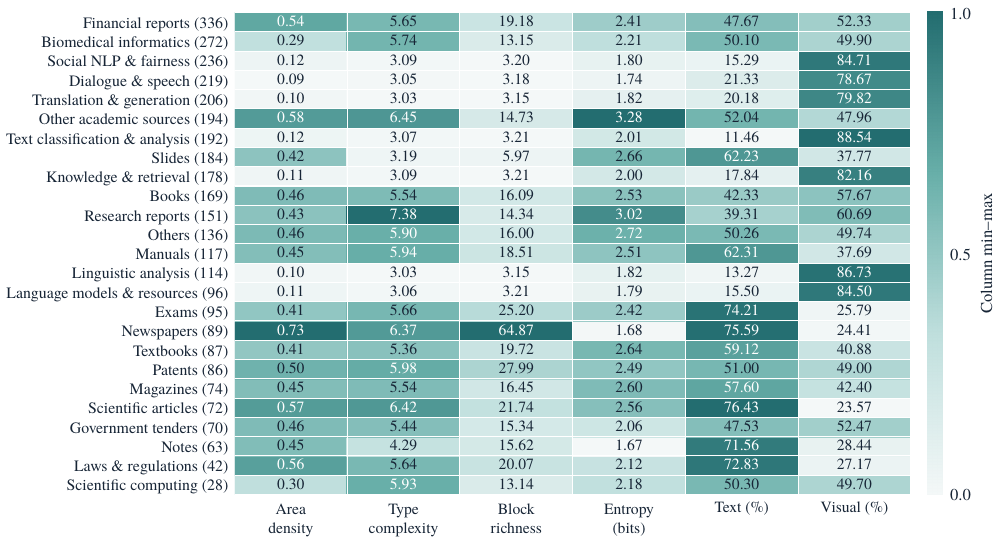}
% 中文：同一批 3,506 张 GT 页面的六项布局统计。
% 数字为实际统计值，颜色按列进行最小—最大归一化；文本和视觉占比以百分数表示。
% 报纸的面积密度和元素数量最高，研究报告的类型复杂度最高，其他学术来源的熵最高。
% 视觉类别包括表格、图像、图表以及页眉与页脚图像。
\caption{\textbf{Comprehensive layout statistics on the same 3,506 GT pages.} Cells report actual values; colors use column-wise min--max normalization. Text and visual ratios are percentages. Newspapers have the highest area density and block richness, research reports the greatest type complexity, and other academic sources the highest entropy. Visual categories comprise tables, images, charts, and header/footer images.}
\label{fig:comprehensive_layout}
\end{figure*}

% 中文标题：数据构建细节
\section{Data Construction Detail}
\label{sec: data construction detail}

% 中文：本节说明构建 VDE Bench 时使用的指令池、候选图像生成方法和参考图像合成方法。
This section describes the instruction pool, candidate image generation, and reference image compositing used to construct VDE Bench.

% 中文标题：修改指令生成
\subsection{Modification Instruction Generation}
\label{sec:modification instruction generation}

% 中文：我们从 Gemini3-Pro、Qwen3-VL-235B-A22B-Instruct 和 GPT-5.6-sol 中随机选择模型，生成详细修改指令。
% 三种指令生成器在相同的纯图像输入设置下提供相互补充的候选指令。
% 生成修改指令时，输入仅包含提示词和原始文档图像，不包含图像经 OCR 解析后的内容。
% 这种做法旨在鼓励模型关注图像本身，避免生成偏向 OCR 解释的指令。
We randomly select among Gemini3-Pro, Qwen3-VL-235B-A22B-Instruct, and GPT-5.6-sol to generate detailed editing instructions. The three instruction generators provide complementary candidate instructions under the same image-only input setting. Each input consists only of a prompt and the original document image, without any OCR-parsed content. This encourages the models to focus on the image itself and avoids instructions biased by OCR interpretations.

% 中文标题：提示词池
\subsubsection{Prompt Pool}
\label{sec:prompt pool}

% 中文：为减轻单一提示词引入的偏差，我们在图像描述生成阶段，从含有 21 个提示词的提示词池中随机采样。
% 具体来说，文本维度中的文本添加、修改和删除各有 3 个提示词；表格维度中的文本添加、修改、删除及表格结构修改也各有 3 个提示词。
% 下面列出若干具有代表性的提示词示例。
To reduce bias from any single prompt, we randomly sample from a pool of 21 prompts during image description generation. The pool contains three prompts for each of text addition, modification, and deletion in text regions, and three prompts for each of text addition, modification, deletion, and table structure modification in table regions. Representative prompts are shown below.

\begin{tcolorbox}[breakable,title=text delete prompt,
                  colback=yellow!10,
                  colframe=yellow!50!black]
                  
% 中文：请针对输入图像生成一条精确的文本删除指令。
% 指令必须明确指出要删除的具体文本元素，而且只能针对其中一个实例。
% 不得修改、引用或包含表格内部或嵌入图像中的任何文本；只有独立标题或正文文本可以作为修改对象。
% 输出必须仅包含纯文本形式的编辑指令，不得附加解释、评论或格式标记。
% 请确保指令使用与图像中文本相同的主要语言。
Generate a precise text deletion instruction for the input image. The instruction must identify the specific text element to delete and target only one instance. Do not modify, refer to, or include any text within tables or embedded images; only standalone titles or body text may be targeted. Output only the editing instruction in plain text, without explanations, comments, or formatting markers. Use the same primary language as the text in the image.

\end{tcolorbox}

\begin{tcolorbox}[breakable,title=text delete prompt,
                  colback=yellow!10,
                  colframe=yellow!50!black]
                  
% 中文：请为输入图像生成一条文本删除指令。
% 不得修改表格内部或图像内部的任何文本，只允许修改标题和正文。
Generate a \textbf{text deletion instruction} for the input image. Do not modify any text inside tables or images; only titles and body text may be modified.

% 中文：1. 回答只能包含编辑指令本身，不得包含其他内容。
1. Your response must contain only the editing instruction itself, with no additional content. 

% 中文：2. 回答必须为纯文本，不得使用任何 Markdown 格式。
2. Your response must be plain text, without any Markdown formatting. 

% 中文：3. 所提供的指令必须清楚说明要删除图像中的哪一处文本。
3. The instruction must clearly identify the text in the image to be deleted.

% 中文：4. 指令语言必须与图像使用的主要语言一致。
% 例如，图像中的主要语言为中文时，请用中文回答；主要语言为英文时，请用英文回答。
4. The instruction must use the primary language of the image. For example, respond in Chinese if the image is primarily in Chinese, and in English if it is primarily in English.

% 中文：5. 只修改一处位置。
5. Modify only one location.

\end{tcolorbox}

\begin{tcolorbox}[breakable,title=text delete prompt,
                  colback=yellow!10,
                  colframe=yellow!50!black]

% 中文：目标：为提供的输入图像生成一条清晰、精确的文本删除指令。
\textbf{Objective}: Generate a clear and precise text deletion instruction for the provided input image.

% 中文：要求：
\textbf{Requirements}:

% 中文：1. 修改范围：
1. \textbf{Scope of Modification}:  
% 中文：每条指令只删除一个文本元素。
   - Only delete one text element per instruction.  
% 中文：目标仅限标题或正文文本。
   - Targets are limited to titles or body text.  
% 中文：不得修改表格内部或嵌入图像中的任何文本。
   - Do not modify any text inside tables or embedded images.

% 中文：2. 指令格式：
2. \textbf{Instruction Format}:  
% 中文：回答只能包含删除指令。
   - The response must consist solely of the deletion instruction.  
% 中文：仅使用纯文本，不得包含 Markdown、解释或额外评论。
   - Use plain text only, without Markdown, explanations, or additional comments.

% 中文：3. 语言一致性：
3. \textbf{Language Consistency}:  
% 中文：指令必须使用图像中的主要语言书写，例如图像文本为英文时使用英文，为中文时使用中文。
   - Write the instruction in the primary language of the image, such as English for English text and Chinese for Chinese text.

% 中文：4. 清晰性与精确性：
4. \textbf{Clarity and Precision}:  
% 中文：准确指出应删除图像中的哪一段文本。
   - Specify exactly which text in the image should be deleted.  
% 中文：避免歧义；编辑人员或自动系统应能够直接执行该指令，而无需进一步澄清。
   - Avoid ambiguity: an editor or automated system should be able to execute the instruction directly, without further clarification.

% 中文：使用示例：
Example Usage:  

% 中文：输入图像包含标题“Monthly Report”（月度报告）和正文“Sales increased by 20%.”（销售额增长了 20%）。
Input image: A title ``Monthly Report'' and body text ``Sales increased by 20\%.''

% 中文：指令输出：删除标题“Monthly Report”。
Instruction output: Delete the title ``Monthly Report''.

\end{tcolorbox}

% 中文标题：生成过程
\subsubsection{Generation Procedure}

% 中文：我们采用块级采样策略，为每张文档图像从适用的编辑类型中抽取提示词。
% 为保证语义有效性，当目标图像不包含任何表格内容时，便从候选池中排除表格相关提示词，例如插入行或列、修改单元格、编辑表格结构，从而避免产生无法忠实执行的不适定指令。
% 这种类型感知采样同时保证了编辑操作的多样性，以及指令与视觉内容的一致性。
% 通过这一流程，我们最终构建了 6,748 条编辑指令，覆盖多种编辑类型与文档布局，共同构成该基准的指令池。
We use a block-level sampling strategy to select prompts from the editing types applicable to each document image. To ensure semantic validity, we exclude table-related prompts, such as row or column insertion, cell modification, and table structure editing, whenever the target image contains no table content. This avoids ill-posed instructions that cannot be faithfully executed. This type-aware sampling ensures both diverse editing operations and consistency between instructions and visual content. The resulting pool contains 6,748 editing instructions covering a wide range of editing types and document layouts.

% 中文标题：图像编辑
\subsection{Image Editing}

% 中文：根据每张源页面及其指令生成候选编辑图像，人工审核负责检查所要求的修改、划定修改区域，并筛选用于基准的参考候选图像。
Candidate edited images are generated from each source page and its instruction. Human review verifies the requested changes, delineates the editing regions, and selects reference candidates for the benchmark.

% 中文：参考图像合成。
\paragraph{Reference compositing.}
% 中文：设 I 为源页面，G(I,r) 为指令 r 对应的已审核候选图像，M∈{0,1}^{H×W} 为标注的编辑区域二值掩码。
% 参考图像按下式构建：在掩码内采用候选编辑图像，在掩码外采用原始页面。
Let $I$ denote the source page, $G(I,r)$ the reviewed candidate image for instruction $r$, and $M\in\{0,1\}^{H\times W}$ the binary mask of the annotated editing region. The reference image uses the candidate image inside the mask and the original page outside it:
\begin{equation}
I^\star=M\odot G(I,r)+(1-M)\odot I.
\end{equation}
% 中文：因此，(1−M)⊙(I*−I)=0：在编码或缩放之前，标注区域以外的像素直接从源页面复制。
% 编辑区域内部的内容则通过人工审核和 OCR 核验评估，详见标注细节附录。
Thus, $(1-M)\odot(I^\star-I)=0$: before encoding or resizing, pixels outside the annotated region are copied directly from the source page. Content inside the edited region is assessed through human review and OCR verification, as detailed in Appendix~\ref{sec:annotation detail}.

% 中文标题：标注与人工评测
\section{Annotation and Human Evaluation}
\label{sec: append annotations}
% 中文标题：标注指南
\subsection{Annotation Guidelines}

% 中文：为确保标注质量与一致性，我们向专业标注人员提供以下指南。
% 标注人员需要严格依据给定的种子文档图像，建立指令驱动的编辑样本对。
% 核心约束如下：
To ensure annotation quality and consistency, professional annotators were provided with the following guidelines. Annotators were tasked with creating instruction-based editing pairs strictly grounded in the provided seed document images. The core constraints were:

\begin{itemize}[leftmargin=*, itemsep=1pt, topsep=2pt]
% 中文：编辑指令必须以文档图像中的视觉内容为依据。
% 例如，禁止引用图像中不存在的文本，也不得要求使用超出文档内容的外部知识。
    \item \textbf{Editing instructions must be grounded in the visual content of the document image} (e.g., forbidding instructions that reference text not present in the image or require external knowledge beyond the document).
% 中文：指令必须没有歧义，而且能够在图像中精确定位。
    \item \textbf{Instructions must be unambiguous and precisely locatable within the image.}
    \begin{itemize}
% 中文：反例：“修改标题。”当图像中存在多个标题时，这一指令有歧义。
        \item \ding{55} Bad: ``Change the title.'' (Ambiguous when multiple titles exist)
% 中文：反例：“修正拼写错误。”这一指令要求隐含地知道正确拼写形式。
        \item \ding{55} Bad: ``Fix the typo.'' (Requires implicit knowledge of the correct form)
% 中文：正例：“将摘要中的 ‘machine leaning’ 替换为 ‘machine learning’。”该指令唯一确定了目标区域，并明确指定具体修改内容。
        \item \ding{51} Good: ``Replace `machine leaning' in the abstract with `machine learning'.'' (Uniquely identifies the target region and specifies the exact modification)
    \end{itemize}
% 中文：编辑后必须忠实保留原有文本样式、字体、颜色与背景上下文。
    \item \textbf{The original text style, font, color, and background context must be faithfully preserved after editing.}
% 中文：指令必须使用与目标文本区域相同的语言：中文文本使用中文指令，英文文本使用英文指令。
    \item \textbf{Instructions must be expressed in the same language as the target text region} (Chinese instructions for Chinese text, English instructions for English text).
\end{itemize}

% 中文：对于每条编辑样本，标注人员同时提供真实参考编辑图像（GT）和精确的编辑区域边界框，以支持整幅图像与局部区域两个层面的细粒度评测。
% 依据修改性质，编辑类型分为以下四类：
For each editing sample, annotators provided the \textbf{Ground-Truth Edited Image} along with the precise \textbf{Edited Region Bounding Box}, enabling fine-grained evaluation at both the full-image and local-region levels. Edit types were categorized into four classes based on the nature of the modification:

\begin{enumerate}[leftmargin=*, itemsep=1pt, topsep=2pt]
% 中文：文本替换／修改：在保持布局的同时，用新内容替换现有文本。
    \item \textbf{Text Replacement / Modify :} Substituting existing text with new content while preserving layout.\\
% 中文：示例：“将页眉中的 ‘2023’ 替换为 ‘2024’。”
    Example: ``Replace `2023' in the header with `2024'.''
% 中文：文本添加：向已有区域插入新的文本内容。
    \item \textbf{Text Addition:} Inserting new textual content into an existing region.\\
% 中文：示例：“在文档标题后添加 ‘(Revised)’（修订版）。”
    Example: ``Add `(Revised)' after the document title.''
% 中文：文本删除：删除指定文本，同时保持视觉连贯性。
    \item \textbf{Text Deletion:} Removing specified text while maintaining visual coherence.\\
% 中文：示例：“删除页面底部的脚注。”
    Example: ``Delete the footnote at the bottom of the page.''
% 中文：表格结构编辑：改变行、列或单元格结构。
    \item \textbf{Table Structure Editing:} Changing row, column, or cell structure.\\
% 中文：示例：“将第一行中的两个单元格合并为一个表头单元格。”
    Example: ``Merge the two cells in the first row into one header cell.''
\end{enumerate}

% 中文标题：人工标注协议
\subsection{Human Annotation Protocol}
\label{sec: append human annotation protocol}

% 中文：标注人员由 18 名计算机科学专业的硕士生与博士生组成，均具备扎实的文档理解和视觉分析背景。
% 标注开始前，我们组织了系统培训，确保标注人员充分理解标注指南，并遵循统一标准。
% 培训包括逐条详细讲解指令、基于示例进行标准校准，以及带反馈的试标轮次，以尽量降低标注人员之间的差异。
Our annotators consist of 18 master's and PhD students in computer science, all of whom possess strong backgrounds in document understanding and visual analysis. Prior to the annotation phase, we conducted structured training sessions to ensure that annotators fully understood the annotation guidelines and followed a unified standard. The training included detailed instruction walkthroughs, example-based calibration, and pilot annotation rounds with feedback to minimize inter-annotator variance.

% 中文：在标注过程中，标注人员需要仔细检查每张修改后的图像，并从多个质量维度进行评价。
% 评测协议旨在同时覆盖客观正确性与主观感知质量，具体使用以下指标：
During the annotation process, annotators were asked to carefully examine each modified image and evaluate it according to multiple quality dimensions. The evaluation protocol is designed to capture both objective correctness and subjective perceptual quality. Specifically, the following metrics were used:

\begin{itemize}[leftmargin=*, itemsep=1pt, topsep=2pt]
% 中文：指令遵循程度：评估修改后的图像是否准确遵循给定修改指令。
% 标注人员检查全部要求的改动是否正确执行，以及是否存在遗漏或非预期修改。
    \item \textbf{Instruction Compliance:} This metric evaluates whether the modified image accurately follows the given modification instructions. Annotators assess if all required changes are correctly applied, without omissions or unintended alterations.
    
% 中文：修改真实性：衡量修改后视觉文档的真实自然程度。
% 标注人员检查输出是否自然、连贯，尤其关注文本完整性，例如是否存在伪影、失真或渲染错误，以及布局和格式是否一致。
    \item \textbf{Modification Authenticity:} This metric measures the realism of the modified visual document. Annotators examine whether the output appears natural and coherent, with particular attention to textual integrity (e.g., absence of artifacts, distortions, or rendering errors) and consistency in layout and formatting.
    
% 中文：主观满意度：反映标注人员对结果的整体主观印象。
% 该指标允许考虑客观标准未能充分覆盖的因素，例如视觉清晰度、可读性，以及图像质量偏低或宽高比错误等问题。
    \item \textbf{Subjective Satisfaction:} This metric captures the annotators' overall subjective impression of the result. It allows them to account for factors that may not be fully covered by objective criteria, such as visual clarity, readability, and issues like low image quality or incorrect aspect ratios.
\end{itemize}

% 中文：每项指标采用 1 至 3 分的离散评分，其中 1 分代表最低质量，3 分代表最高质量。
% 为确保一致性，标注人员遵循共同评分细则；对于模糊案例，必要时通过讨论或二次审核解决。
Each metric is rated on a discrete scale from 1 to 3, where 1 indicates the lowest quality and 3 indicates the highest quality. To ensure consistency, annotators followed a shared scoring rubric, and ambiguous cases were resolved through discussion or secondary review when necessary.

% 中文标题：标注应用系统
\subsection{Application}

% 中文：人工标注为完整基准提供经核验的参考结果和编辑区域边界框。
% 另一项独立研究基于抽取的 1,000 张图像，比较七个模型的人工与自动排名，详见人工评测部分。
Human annotation produces the verified references and editing-region boxes for the full benchmark. A separate study compares human and automated rankings of seven models using 1,000 sampled images (Section~\ref{sec:human eval}).

% 中文：为确保标注流程规范且高效，我们基于 Label Studio 开发了定制网页应用，为标注人员提供一套完整工具：
To ensure standardization and efficiency in the annotation process, we developed a custom web application using label studio, providing annotators with a complete set of tools:

\begin{itemize}[leftmargin=*, itemsep=1pt, topsep=2pt]
% 中文：交互式多图像查看器，支持首页、上一页、下一页、末页及跳转页面等导航操作，使标注人员能够充分浏览全部可标注内容，而非仅限于初始搜索结果。
    \item An interactive multi-image viewer that supports page navigation (first, previous, next, last, jump to page), enabling annotators to thoroughly browse all annotatable content rather than being limited to initial search results.
% 中文：标注人员可以标记图像中的具体位置，作为支持答案的证据。
% 这不仅帮助他们准确定位关键信息，也为解决标注冲突提供依据。
% 当标注不一致时，最终裁决人员可参考不同标注人员给出的证据作出判断，从而确保数据集的准确性与一致性。
    \item Annotators can mark specific locations within images as evidence supporting the answers. This not only helps annotators accurately locate key information, but also provides a basis for resolving annotation conflicts. In cases of inconsistent annotations, the final adjudicator can refer to the evidence provided by different annotators to make informed decisions, ensuring the accuracy and consistency of the dataset.
\end{itemize}

% 中文：该标注系统旨在尽可能提高人工标注的可靠性，同时为后续与 AI 生成图像的比较提供可靠的基准数据。
This annotation system is designed to maximize the reliability of human annotations while providing robust benchmark data for subsequent comparisons with AI-generated images.

% 中文标题：修改区域标注
\subsection{Modification Annotation}
\label{sec:annotation detail}

% 中文：由于密集文本文档的修改通常涉及对整幅图像进行编辑，而背景又包含大量密集文本，因此标注流程也相应更加复杂：
Since modifying dense-text documents typically involves whole-image editing with densely populated background text, the annotation workflow is correspondingly more complex:

\begin{enumerate}[leftmargin=*,itemsep=1pt,topsep=2pt,label=(\arabic*)]
% 中文：为得到文本修改图像的真实参考结果，标注人员首先分离前景与背景。
% 标注人员分为三个组，共享同一个数据池。
% 他们先对文本修改区域评分，并按照人工标注协议中的规则，筛选出总分为 9 分的高质量图像。
\item To obtain the ground truth for text-modified images, annotators first segment the foreground and background. Annotators are divided into three groups and share a common data pool. They initially score the text modification regions and, following the rules described in~\ref{sec: append human annotation protocol}, filter out high-quality images with a total score of 9.
% 中文：对于筛选出的高质量数据，标注人员绘制修改文本区域的边界框。
% 若样本已由三个组全部标注，且边界框的平均两两 IoU 超过 0.95，则判定通过；否则，重置该样本并将其放回数据池。
% 该过程持续进行，直至所有样本通过。
\item For the high-quality filtered data, annotators draw bounding boxes around the modified text regions. If a sample is annotated by all three groups and the average pairwise IOU of the boxes exceeds 0.95, it is considered approved; otherwise, the sample is re-initialized and returned to the pool. This process continues until all samples are approved.
% 中文：完成边界框标注后，选取各次标注中最大的边界框作为真实参考框。
% 提取该框内部的修改后图像区域，并粘贴回原始图像，生成既保持原始背景又包含指定修改的新图像。
\item Once the bounding boxes are completed, the largest box in each annotation is selected as the ground truth. The modified image region within the box is extracted and pasted back onto the original image, producing a new image that preserves the original background while incorporating the modifications.
% 中文：标注人员对生成图像进行最后一轮筛选。
% 此步骤完全基于主观判断：只有所有组均同意时图像才通过，否则予以拒绝。
\item Annotators perform a final screening of the generated images. This step is entirely subjective: an image is approved only if all groups agree; otherwise, it is rejected.
% 中文：图像级标注完成后，进一步使用 PaddleOCR-VL 对生成图像的整页和修改区域进行 OCR 识别。
% 在该步骤中，还须人工核验修改区域内的 OCR 结果；只有这些区域的 OCR 结果正确时，样本才予以通过。
\item After the image-level annotations are completed, we further apply PaddleOCR-VL \cite{cui2025paddleocrvlboostingmultilingualdocument} to perform OCR recognition on both the entire generated image and the modified regions. During this step, manual verification of the OCR results in the modified regions is also required. A sample is approved only if the OCR results for the modified regions are correct.
\end{enumerate}

% 中文：完成上述标注流程后，我们获得了包含 3,506 张按指令修改的图像的数据集，以及相应修改区域的裁剪图像。
% 此外，数据集还包含通过 PaddleOCR-VL 生成的结构化 JSON 标注文件，同时记录整幅图像的全局标注信息与修改区域的局部标注信息。
Upon completion of the aforementioned annotation process, we obtain a dataset consisting of 3,506 instruction-modified images, together with cropped regions corresponding to the modified areas. In addition, the dataset includes structured JSON annotation files generated using PaddleOCR-VL, containing both global annotation information for the entire image and localized annotation information for the modified regions.

% 中文标题：字符识别与结构化处理
\subsection{Character Recognition Structured Processing}
\label{sec:append CR Structure Processing}

% 中文：标注人员依据参考图像和编辑指令，核验每个保留编辑区域内识别出的文本。
% 结构化标注同时包含整页 OCR 输出和这些经过人工检查的局部输出。
Annotators verify the recognized text within every retained editing region against the reference image and instruction. The structured annotations contain both full-page OCR outputs and these manually checked local outputs.

% 中文：删除区域。
% 对于删除指令，标注区域在删除目标文本后既可能保留周围内容，也可能变为空白。
% 两种情况都属于有效编辑结果，均保留在基准与图像评测中；空 OCR 输出本身不构成样本剔除条件。
% OCR 指标按可用文本块汇总：参考与预测均无块时，该区域不贡献块级条目；有参考块但无匹配预测时，参考块的 IoU 计为 0。具体规则见指标定义附录。
\paragraph{Deletion regions.}
For deletion instructions, the annotated region may retain surrounding content or become blank after the target text is removed. Both are valid editing outcomes and remain in the benchmark and image-based evaluation; an empty OCR output alone does not exclude a sample. OCR metrics aggregate the available blocks: a region with no reference or predicted blocks contributes no block-level entries, while a reference block without a matched prediction receives zero IoU. Matching and aggregation follow Appendix~\ref{sec: append match result}.

% 中文：试标一致性。
% 在正式标注前，全部 18 名标注人员分别独立评定 50 条共同样本的指令遵循程度、修改真实性与主观满意度。
% 每个维度由此获得 900 次评分，相应 Krippendorff α 系数分别为 0.941、0.950 和 0.876。
% 人工一致性图报告试标系数及其 95% bootstrap 置信区间。
% 正式标注采用前述三个组审核、边界框重叠阈值、最终视觉检查及局部 OCR 核验流程。
\paragraph{Pilot agreement.}
Before production annotation, all 18 annotators independently rated 50 shared samples on instruction compliance, modification authenticity, and subjective satisfaction. This yields 900 ratings per dimension, with Krippendorff's $\alpha$ of 0.941, 0.950, and 0.876, respectively. Figure~\ref{fig:human_agreement} reports the pilot coefficients and their 95\% bootstrap confidence intervals. Production uses the three-group review, box-overlap threshold, final visual check, and local OCR verification described above.

% 中文标题：人工评测
\subsection{Human Eval}
\label{sec:human eval}
% 中文：我们基于抽取的 1,000 张图像开展人工评测，比较七个图像编辑模型在框对齐与文本正确性两个维度上的人工排名和自动排名。
We conduct human evaluation using 1,000 sampled images, comparing human and automated rankings of seven image editing models on box alignment and text correctness.

\begin{figure*}[t]
    \centering
    \includegraphics[width=\linewidth]{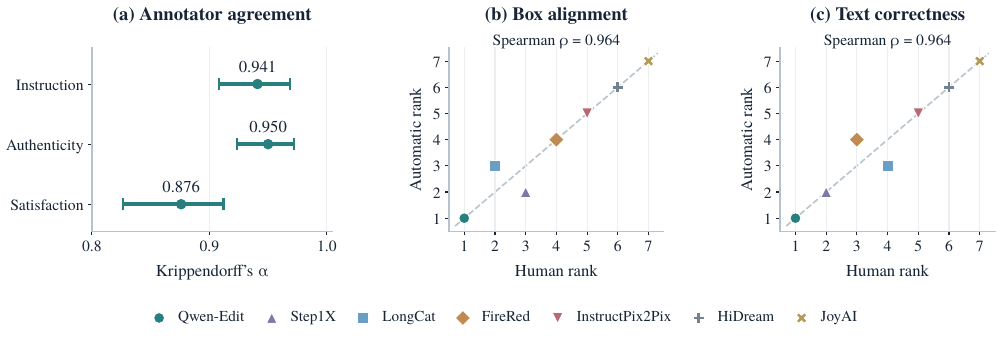}
% 中文：(a) 18 名标注人员在 50 条试标样本上的 Krippendorff α 系数及其 95% bootstrap 置信区间。
% (b,c) 基于 1,000 张抽样图像的人工评审，七个模型在框对齐和文本正确性上的人工与自动排名。每个点对应一个模型，横轴为人工排名，纵轴为自动排名；1 为最佳，7 为最差。两项 Spearman ρ 均为 0.964。
    \caption{(a) Agreement among 18 annotators on 50 pilot samples, measured by Krippendorff's $\alpha$ with 95\% bootstrap confidence intervals. (b,c) Human versus automated rankings of seven models for box alignment and text correctness, based on human evaluation of 1,000 sampled images. Each point represents one model; ranks range from 1 (best) to 7 (worst). Both comparisons yield Spearman's $\rho=0.964$.}
    \label{fig:human_agreement}
\end{figure*}

% 中文：人工评审分别考察框对齐和文本正确性。自动评测中，框对齐对应 IoU，文本正确性由 BLEU-4、CDM 和 TEDS-like 的均值衡量。
% 图中展示本次人工评测研究的模型汇总排名，1 为最佳，7 为最差。
Human reviewers assess box alignment and text correctness separately. For automated evaluation, box alignment corresponds to IoU, while text correctness is measured by the mean of BLEU-4, CDM, and TEDS-like. Figure~\ref{fig:human_agreement} shows the aggregate model rankings for this human-evaluation study, with rank 1 indicating the best performance and rank 7 the worst.

% 中文：两项比较的 Spearman ρ 均为 0.964，分别基于七个模型的汇总排名计算。Qwen-Edit 在两个维度的人工与自动评测中均为第一。
% 框对齐仅 Step1X 与 LongCat 的第 2、3 名互换；文本正确性仅 LongCat 与 FireRed 的第 3、4 名互换。其余模型在对应维度上的排名一致。
% 这些结果表明，自动评测在定位与文本正确性两个维度上均与人工判断高度一致。
Both comparisons yield Spearman's $\rho=0.964$, computed separately over the seven aggregate model rankings (Figure~\ref{fig:human_agreement}). Qwen-Edit ranks first under both human and automated evaluation in both dimensions. For box alignment, the only difference is a swap between Step1X and LongCat at ranks 2 and 3. For text correctness, only LongCat and FireRed swap ranks 3 and 4. All other ranks agree within each dimension. These results show that the automated evaluation closely reflects human judgments of both localization and text correctness.

\FloatBarrier
% 中文标题：案例研究
\section{Case Study}
\label{sec: append case study}

% 中文：案例区采用固定的小间距，避免为撑满栏高而拉大图间空白。
\begingroup
\raggedbottom
\setlength{\intextsep}{6pt}
\captionsetup{skip=4pt}

\par\addvspace{\intextsep}\noindent
\begin{minipage}{\linewidth}\captionsetup{type=figure} 
\centering
\includegraphics[width=1\linewidth,height=0.70\textheight,keepaspectratio]{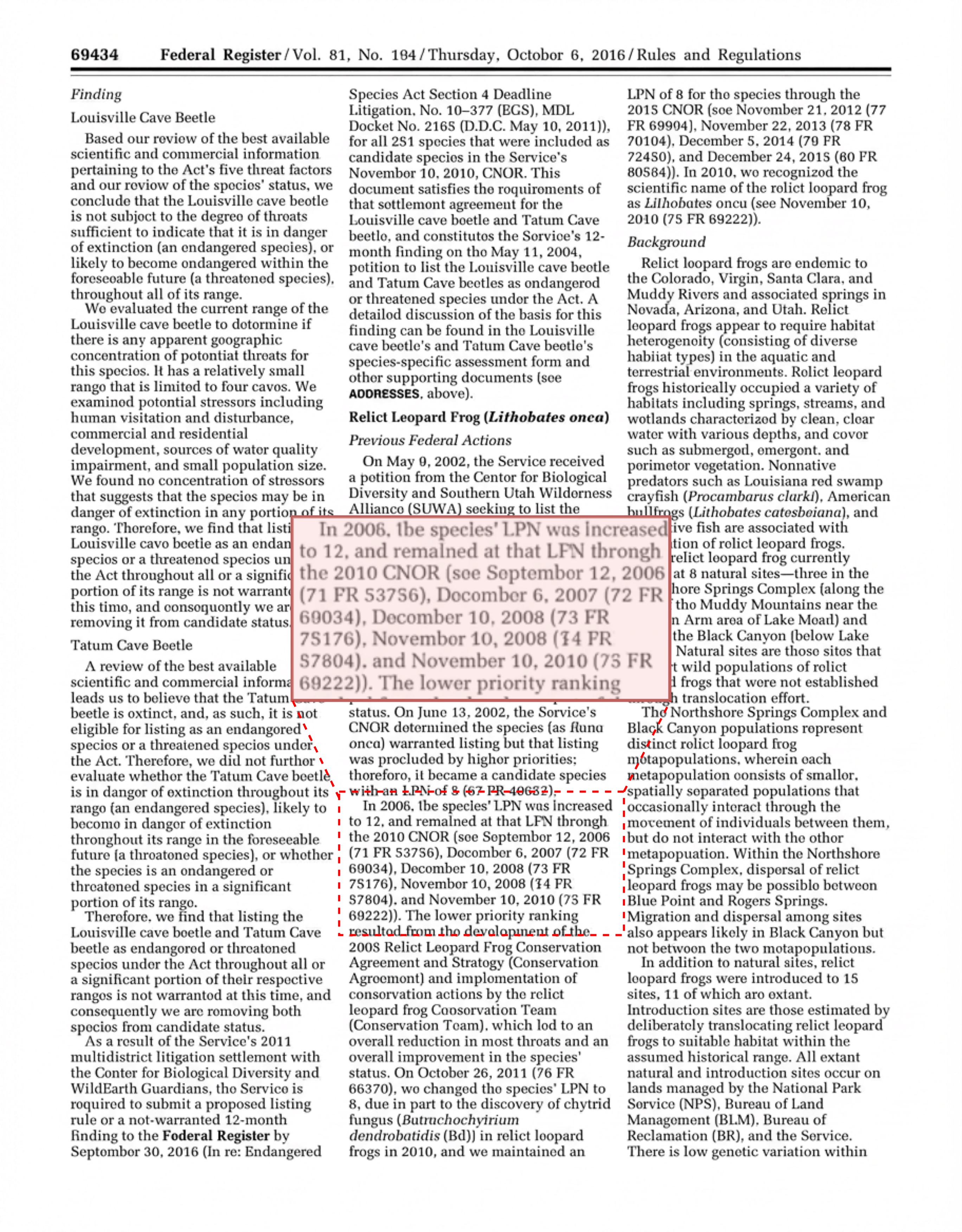} 
% 中文：将“2006 年，该物种的 LPN 下调至 11，并一直保持该 LPN，直至 2010 年的 CNOR（参见 2006 年 9 月 12 日，71 FR 53756；2007 年 12 月 6 日，72 FR 69034；2008 年 12 月 10 日，73 FR 75176；2009 年 11 月 9 日，74 FR 57804；以及 2010 年 11 月 10 日，75 FR 69222）”改为“2008 年，该物种的 LPN 上调至 12，并一直保持该 LPN，直至 2010 年的 CNOR（参见 2006 年 9 月 12 日，71 FR 53756；2007 年 12 月 6 日，72 FR 69034；2008 年 12 月 10 日，73 FR 75176；2009 年 11 月 9 日，74 FR 57804；以及 2010 年 11 月 10 日，75 FR 69222）”。
% 对应英文修改将年份 2006 改为 2008，将“lowered to 11”改为“increased to 12”，其余引用信息保持一致。
\caption{Change ``In 2006, the species' LPN was lowered to 11, and remained at that LPN through the 2010 CNOR (see September 12, 2006 (71 FR 53756), December 6, 2007 (72 FR 69034), December 10, 2008 (73 FR 75176), November 9, 2009 (74 FR 57804), and November 10, 2010 (75 FR 69222))." to ``In 2008, the species' LPN was increased to 12, and remained at that LPN through the 2010 CNOR (see September 12, 2006 (71 FR 53756), December 6, 2007 (72 FR 69034), December 10, 2008 (73 FR 75176), November 9, 2009 (74 FR 57804), and November 10, 2010 (75 FR 69222))."}
\end{minipage}
\par\addvspace{\intextsep}

\par\addvspace{\intextsep}\noindent
\begin{minipage}{\linewidth}\captionsetup{type=figure} 
\centering
\includegraphics[width=1\linewidth,height=0.70\textheight,keepaspectratio]{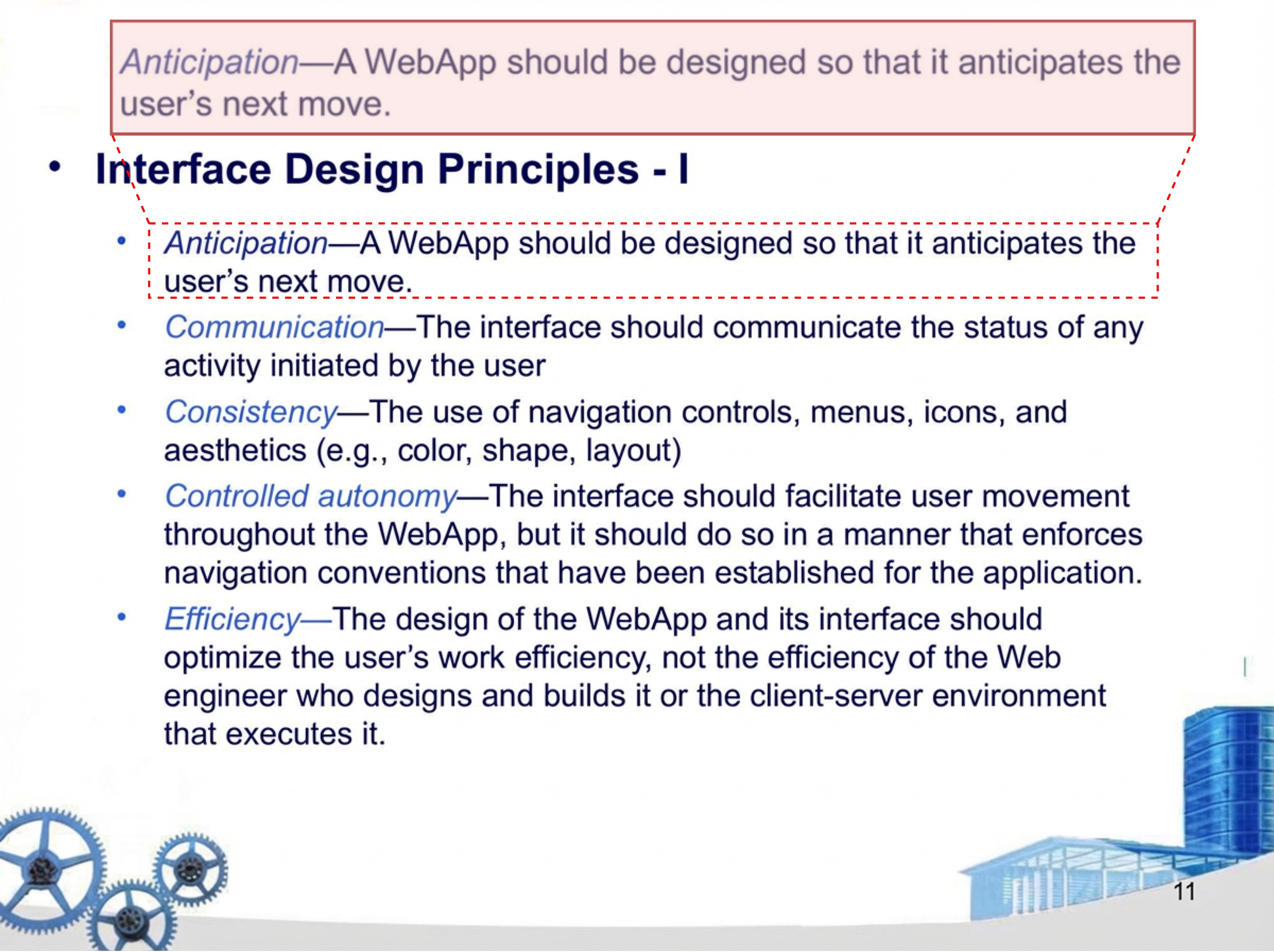} 
% 中文：在句子“Anticipation—A WebApp should be designed so that it anticipates the use's next move.”（预判性——Web 应用的设计应能预判用户的下一步操作）中，将拼写错误的“use's next move”修改为“user's next move”。
\caption{Change ``use's next move" to ``user's next move" in the sentence ``Anticipation—A WebApp should be designed so that it anticipates the use's next move."}
\end{minipage}
\par\addvspace{\intextsep}

\par\addvspace{\intextsep}\noindent
\begin{minipage}{\linewidth}\captionsetup{type=figure} 
\centering
\includegraphics[width=0.9\linewidth,height=0.70\textheight,keepaspectratio]{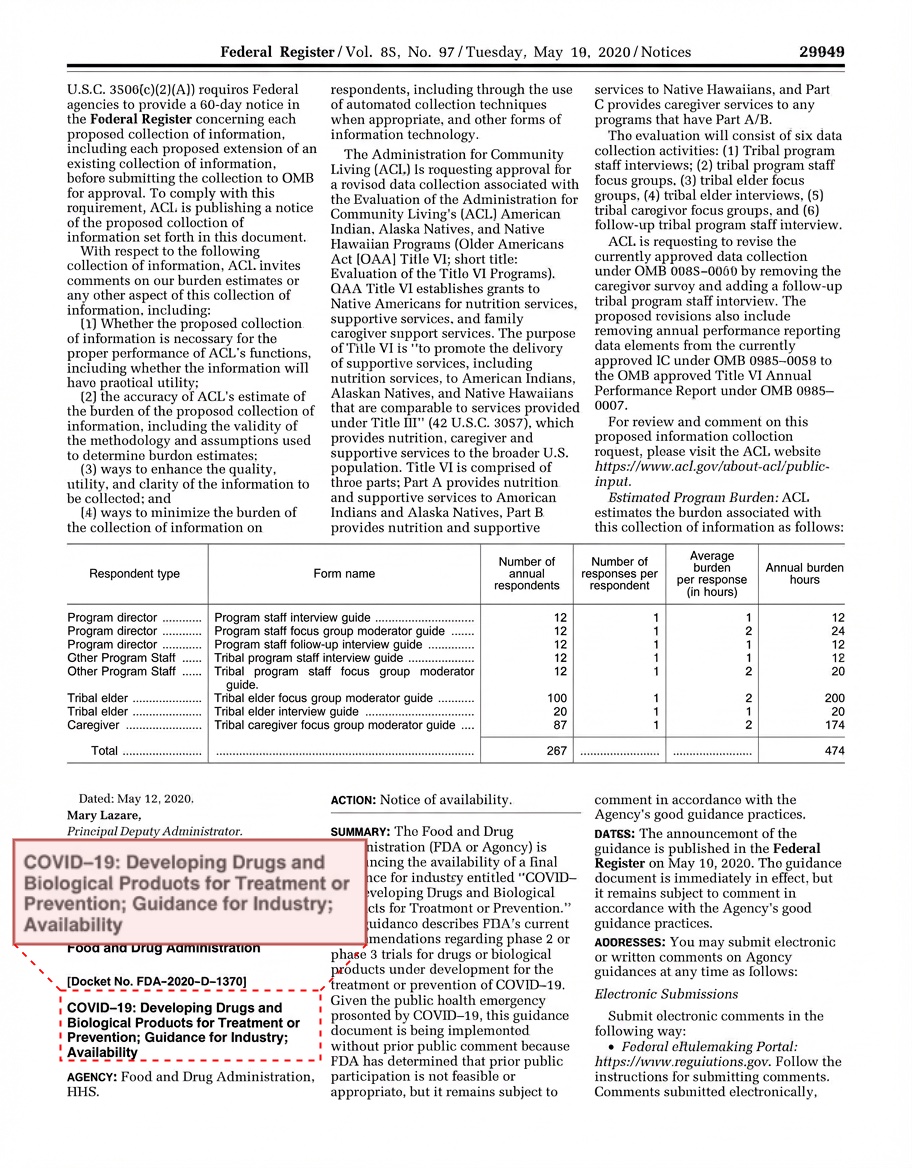} 
% 中文：将“COVID-19: Developing Drugs and Biological Products for Treatment or Prevention; Guidance for Industry; Availability”替换为“Coronavirus Disease 2019: Developing Drugs and Biological Products for Treatment or Prevention; Guidance for Industry; Availability”，即将标题开头的 COVID-19 缩写展开为 Coronavirus Disease 2019；标题意为“2019 冠状病毒病：开发用于治疗或预防的药物与生物制品；行业指南；发布公告”。
\caption{Replace `COVID-19: Developing Drugs and Biological Products for Treatment or Prevention; Guidance for Industry; Availability' with `Coronavirus Disease 2019: Developing Drugs and Biological Products for Treatment or Prevention; Guidance for Industry; Availability'}
\end{minipage}
\par\addvspace{\intextsep}

\par\addvspace{\intextsep}\noindent
\begin{minipage}{\linewidth}\captionsetup{type=figure} 
\centering
\includegraphics[width=0.95\linewidth,height=0.70\textheight,keepaspectratio]{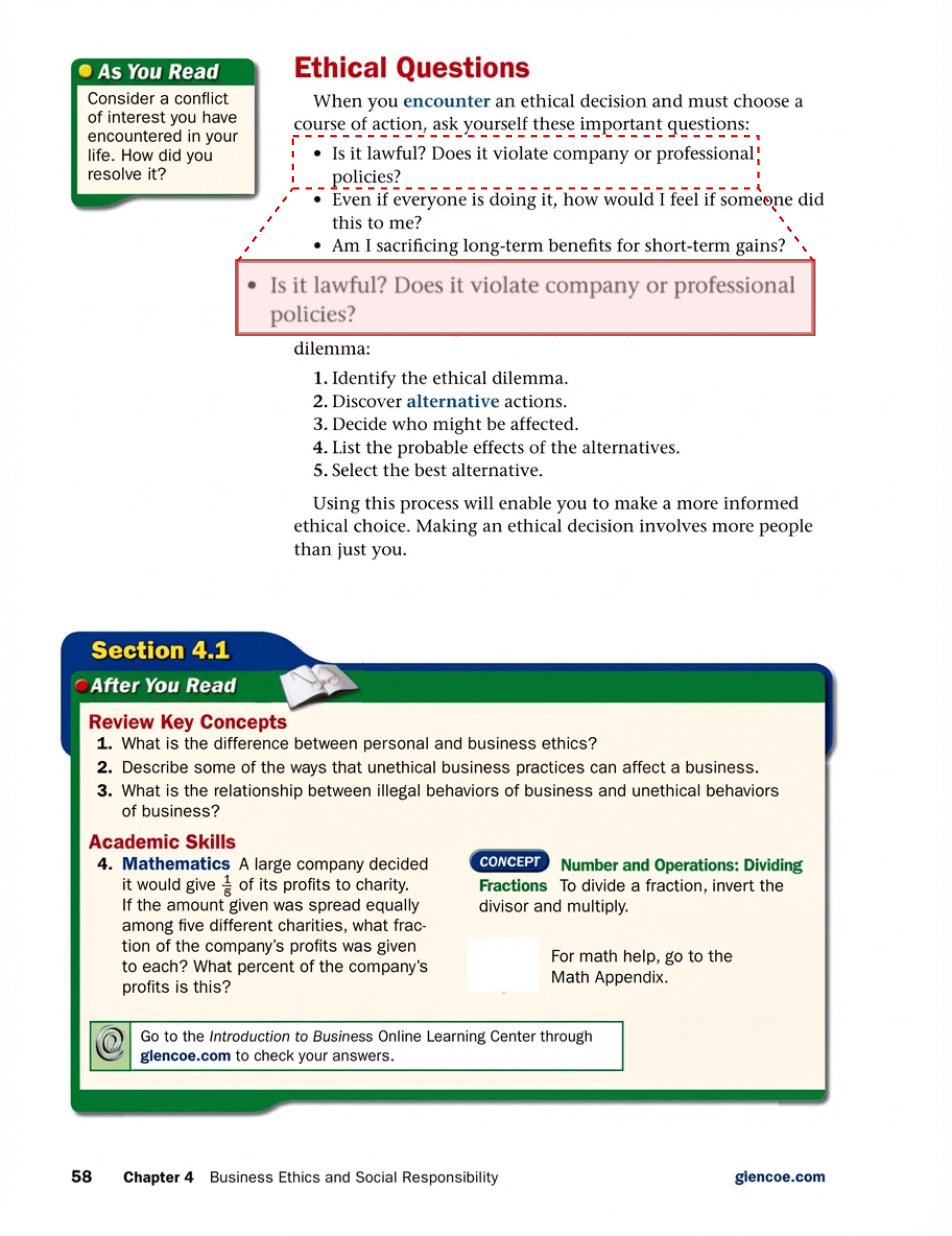} 
% 中文：将“Is it against the law? Does it violate company or professional policies?”（这是否违法？
% 这是否违反公司或行业政策？
% ）修改为“Is it lawful? Does it violate company or professional policies?”（这是否合法？
% 这是否违反公司或行业政策？
% ）。
\caption{Change ``Is it against the law? Does it violate company or professional policies?" to ``Is it lawful? Does it violate company or professional policies?"}
\end{minipage}
\par\addvspace{\intextsep}

\par\addvspace{\intextsep}\noindent
\begin{minipage}{\linewidth}\captionsetup{type=figure} 
\centering
\includegraphics[width=1\linewidth,height=0.70\textheight,keepaspectratio]{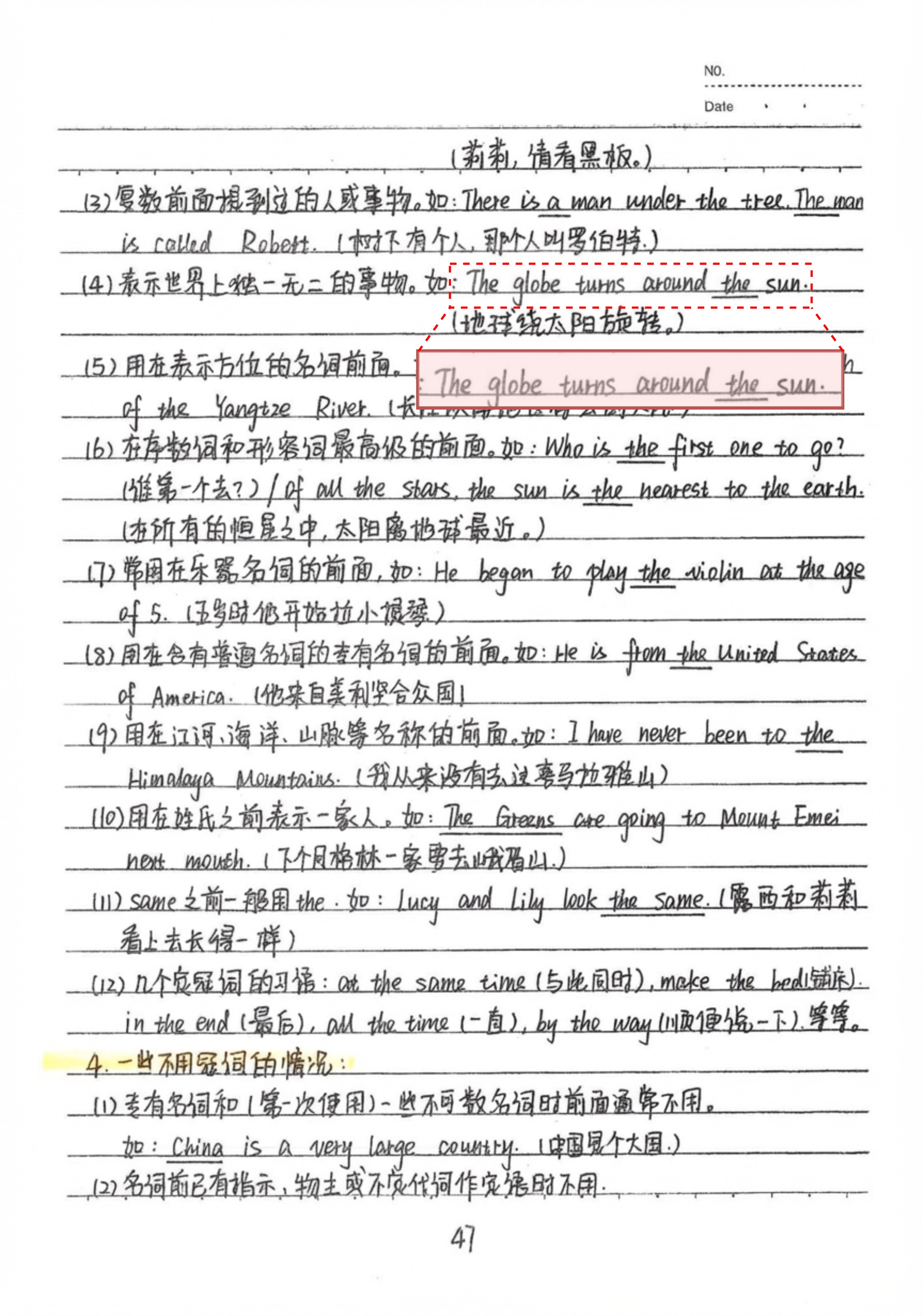} 
% 中文：在第（4）点的句子“The earth turns around the sun.”（地球围绕太阳转动）中，将“earth”替换为“globe”。
\caption{Replace ``earth" with ``globe" in the sentence ``The earth turns around the sun." under point (4).}
\end{minipage}
\par\addvspace{\intextsep}

\par\addvspace{\intextsep}\noindent
\begin{minipage}{\linewidth}\captionsetup{type=figure} 
\centering
\includegraphics[width=1\linewidth,height=0.70\textheight,keepaspectratio]{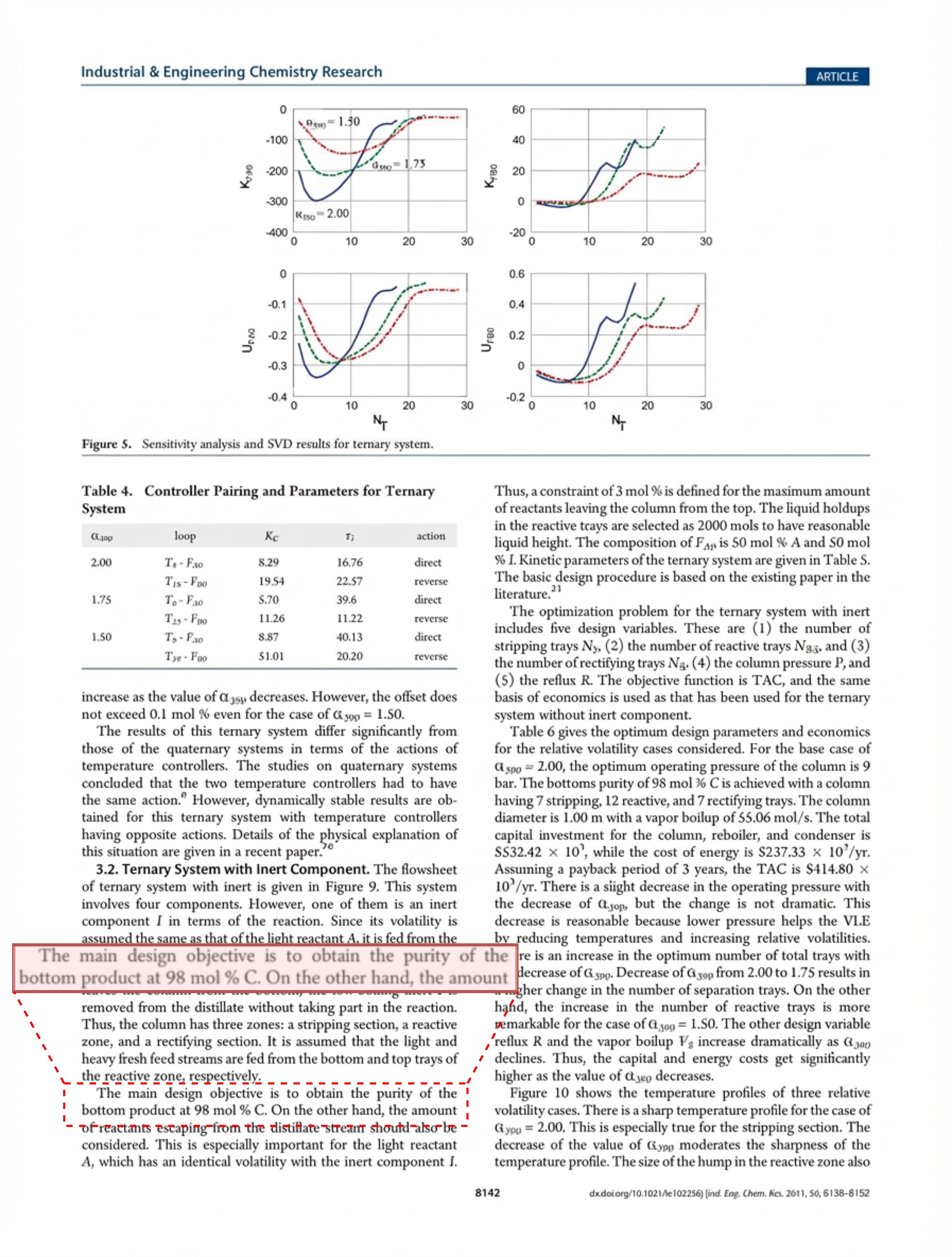} 
% 中文：在句子“The main design objective is to obtain the purity of the bottoms product at 98 mol % C.”（主要设计目标是使塔底产品中 C 的纯度达到 98 mol%）中，将“bottoms product”替换为“bottom product”。
\caption{Replace ``bottoms product" with ``bottom product" in the sentence ``The main design objective is to obtain the purity of the bottoms product at 98 mol \% C."}
\end{minipage}
\par\addvspace{\intextsep}

\par\addvspace{\intextsep}\noindent
\begin{minipage}{\linewidth}\captionsetup{type=figure} 
\centering
\includegraphics[width=1\linewidth,height=0.70\textheight,keepaspectratio]{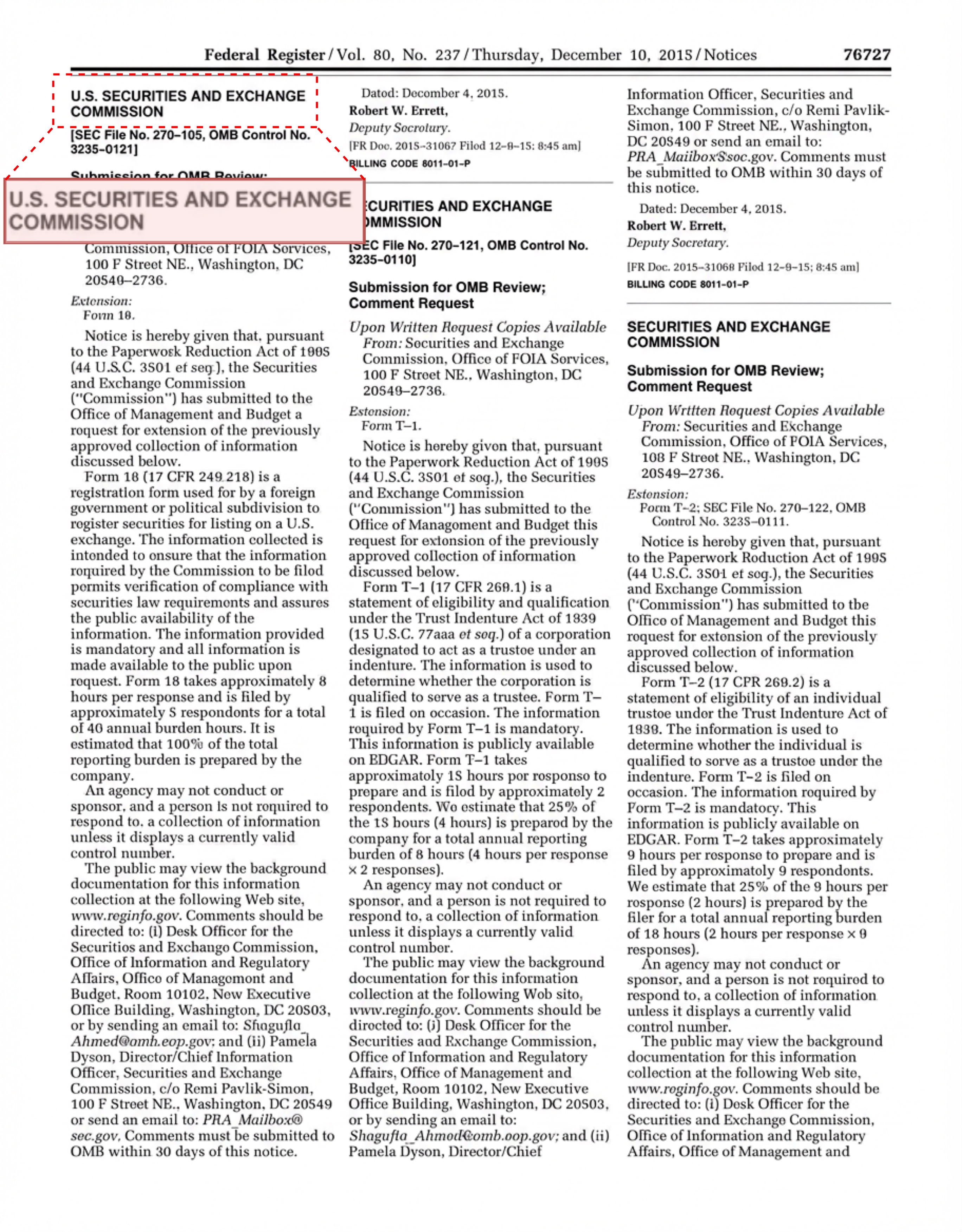} 
% 中文：将第一个主标题中的“SECURITIES AND EXCHANGE COMMISSION”（证券交易委员会）修改为“U.S. SECURITIES AND EXCHANGE COMMISSION”（美国证券交易委员会）。
\caption{Change ``SECURITIES AND EXCHANGE COMMISSION" to ``U.S. SECURITIES AND EXCHANGE COMMISSION" in the first main title.}
\end{minipage}
\par\addvspace{\intextsep}

\par\addvspace{\intextsep}\noindent
\begin{minipage}{\linewidth}\captionsetup{type=figure} 
\centering
\includegraphics[width=0.9\linewidth,height=0.70\textheight,keepaspectratio]{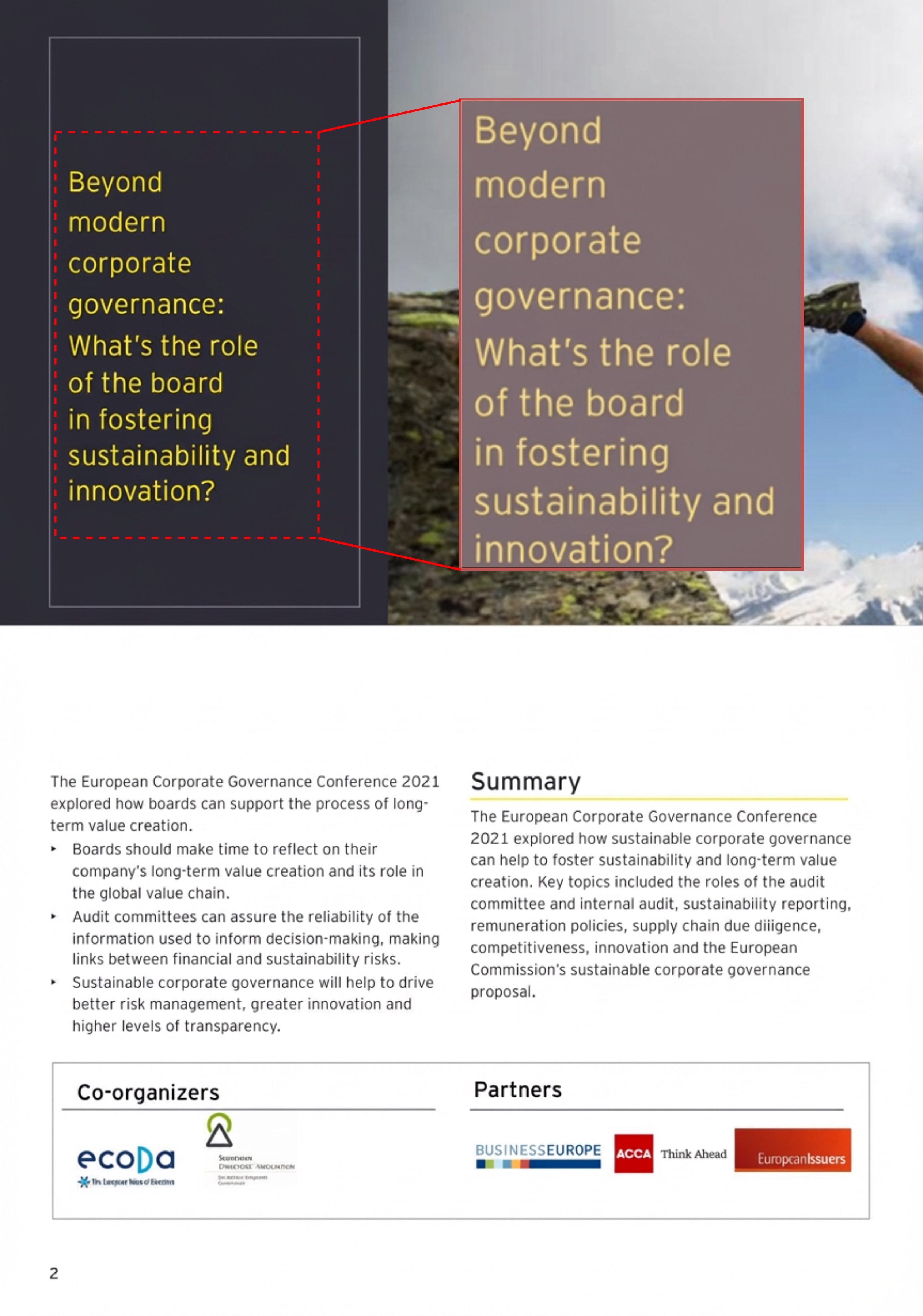} 
% 中文：将“Beyond traditional corporate governance: What's the role of the board in fostering sustainability and innovation?”（超越传统公司治理：董事会在促进可持续发展与创新方面发挥什么作用？
% ）替换为“Beyond modern corporate governance: What's the role of the board in fostering sustainability and innovation?”（超越现代公司治理：董事会在促进可持续发展与创新方面发挥什么作用？
% ）。
\caption{Replace ``Beyond traditional corporate governance: What's the role of the board in fostering sustainability and innovation?" with ``Beyond modern corporate governance: What's the role of the board in fostering sustainability and innovation?"}
\end{minipage}
\par\addvspace{\intextsep}

\par\addvspace{\intextsep}\noindent
\begin{minipage}{\linewidth}\captionsetup{type=figure} 
\centering
\includegraphics[width=0.9\linewidth,height=0.70\textheight,keepaspectratio]{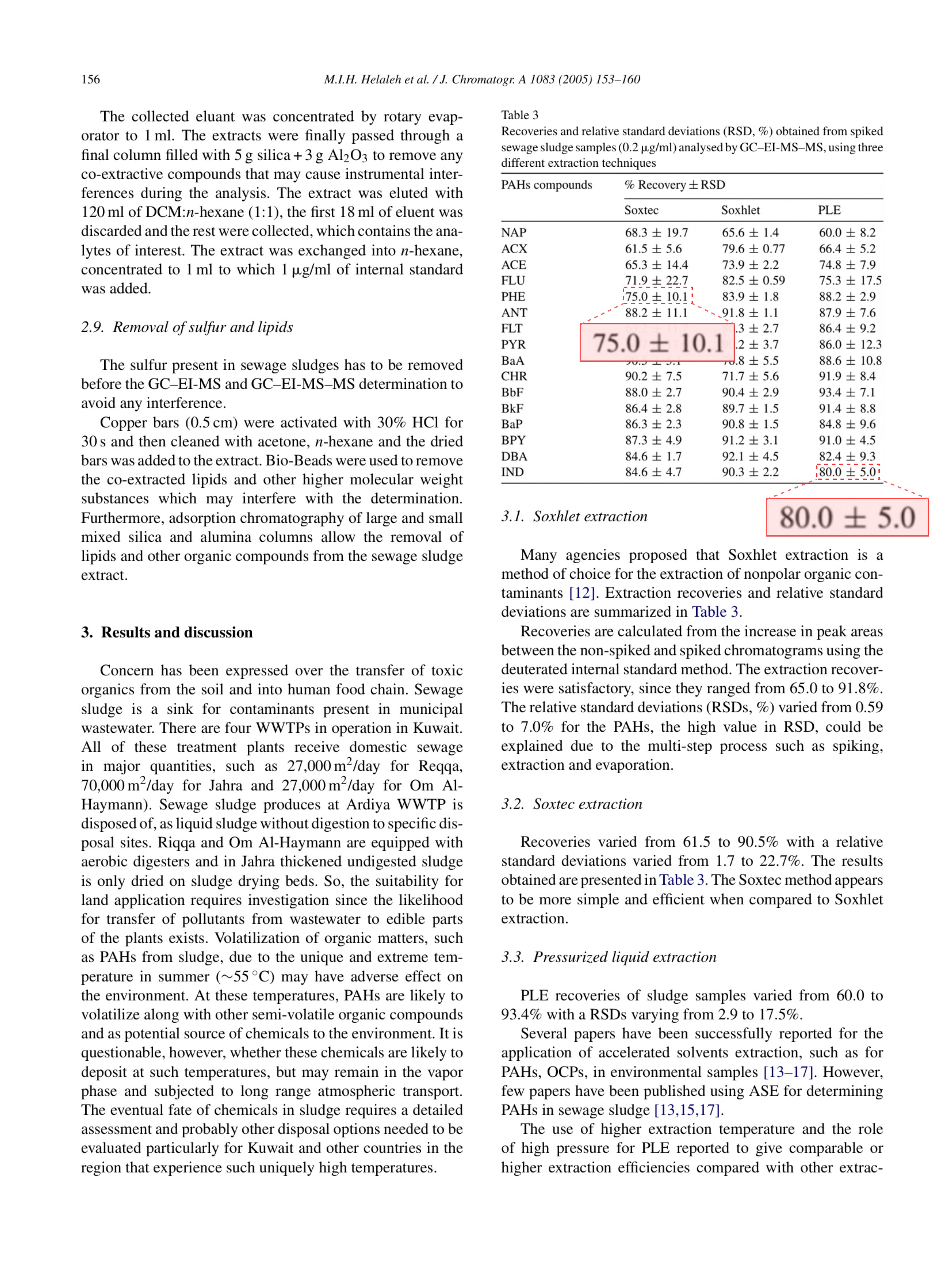} 
% 中文：将表格中“PHE”行、“Soxtec”列的数值从“84.7 ± 16.3”改为“75.0 ± 10.1”；同时将“IND”行、“PLE”列的数值从“85 ± 7.9”改为“80.0 ± 5.0”。
\caption{Change the value in the "PHE" row under the "Soxtec" column from "84.7 ± 16.3" to "75.0 ± 10.1", and change the value in the "IND" row under the "PLE" column from "85 ± 7.9" to "80.0 ± 5.0".}
\end{minipage}
\par\addvspace{\intextsep}

\par\addvspace{\intextsep}\noindent
\begin{minipage}{\linewidth}\captionsetup{type=figure} 
\centering
\includegraphics[width=0.9\linewidth,height=0.70\textheight,keepaspectratio]{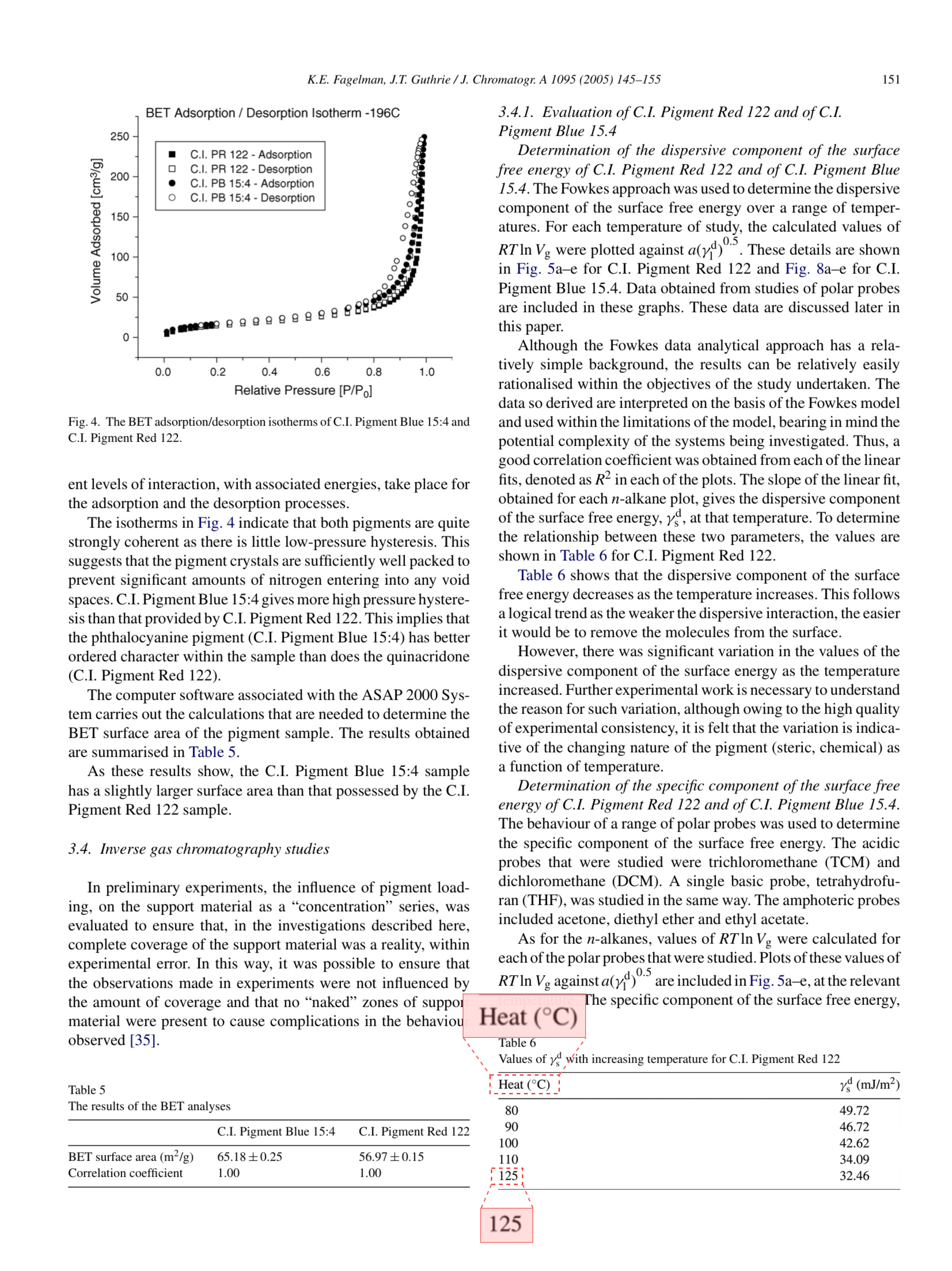} 
% 中文：将列标题“Temperature (°C)”（温度，摄氏度）改为“Heat (°C)”（热量，摄氏度），并将最后一行的数值“120”改为“125”。
\caption{Change the column header "Temperature (°C)" to "Heat (°C)" and change the value "120" in the last row to "125".}
\end{minipage}
\par\addvspace{\intextsep}

\par\addvspace{\intextsep}\noindent
\begin{minipage}{\linewidth}\captionsetup{type=figure} 
\centering
\includegraphics[width=0.9\linewidth,height=0.70\textheight,keepaspectratio]{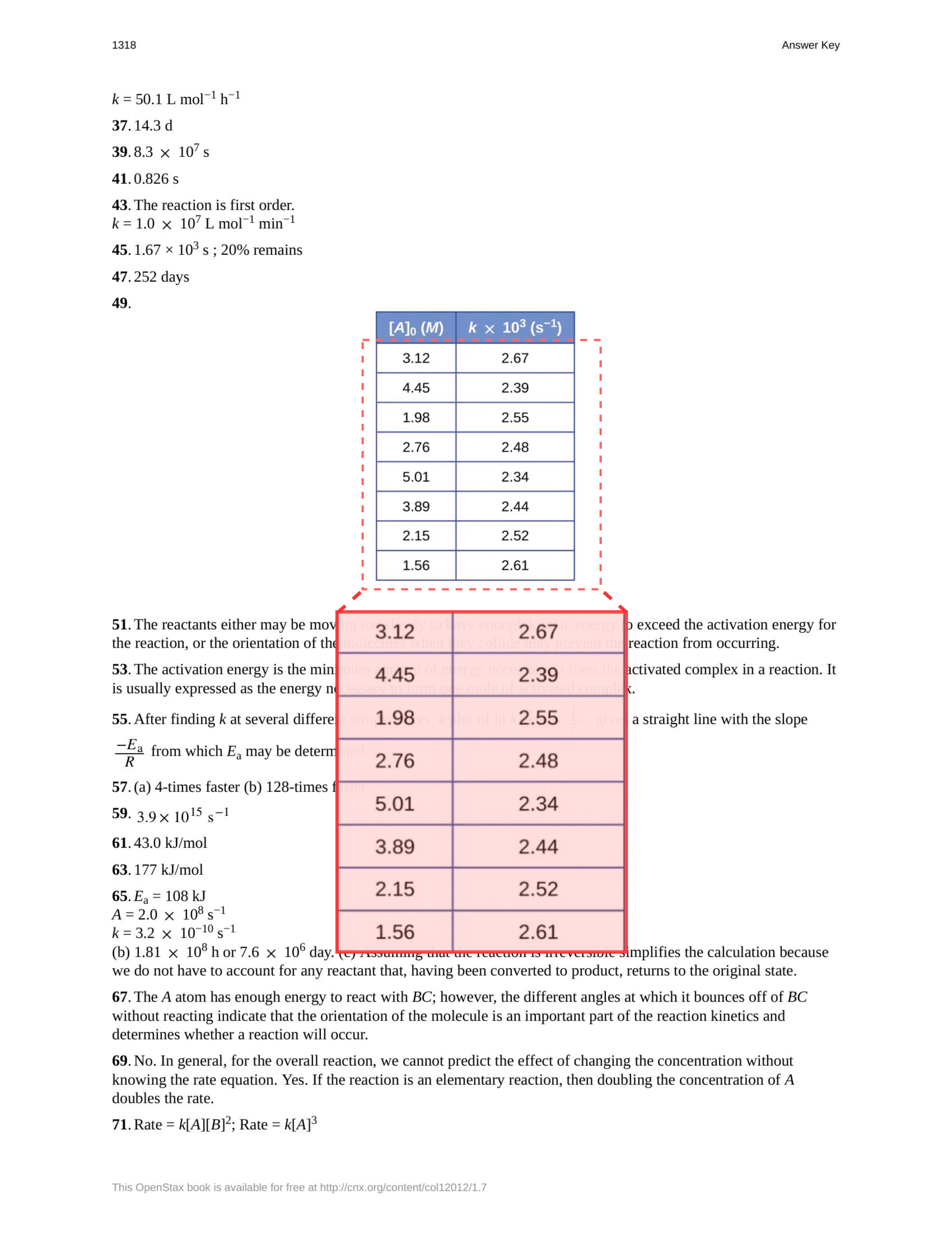} 
% 中文：将第 49 题下方表格中的数值按行替换为以下数据：3.12, 2.67；4.45, 2.39；1.98, 2.55；2.76, 2.48；5.01, 2.34；3.89, 2.44；2.15, 2.52；1.56, 2.61。
\caption{Replace the values in the table under question 49 with the following rows of data: 3.12, 2.67; 4.45, 2.39; 1.98, 2.55; 2.76, 2.48; 5.01, 2.34; 3.89, 2.44; 2.15, 2.52; 1.56, 2.61..}
\end{minipage}
\par\addvspace{\intextsep}

\par\addvspace{\intextsep}\noindent
\begin{minipage}{\linewidth}\captionsetup{type=figure} 
\centering
\includegraphics[width=0.9\linewidth,height=0.70\textheight,keepaspectratio]{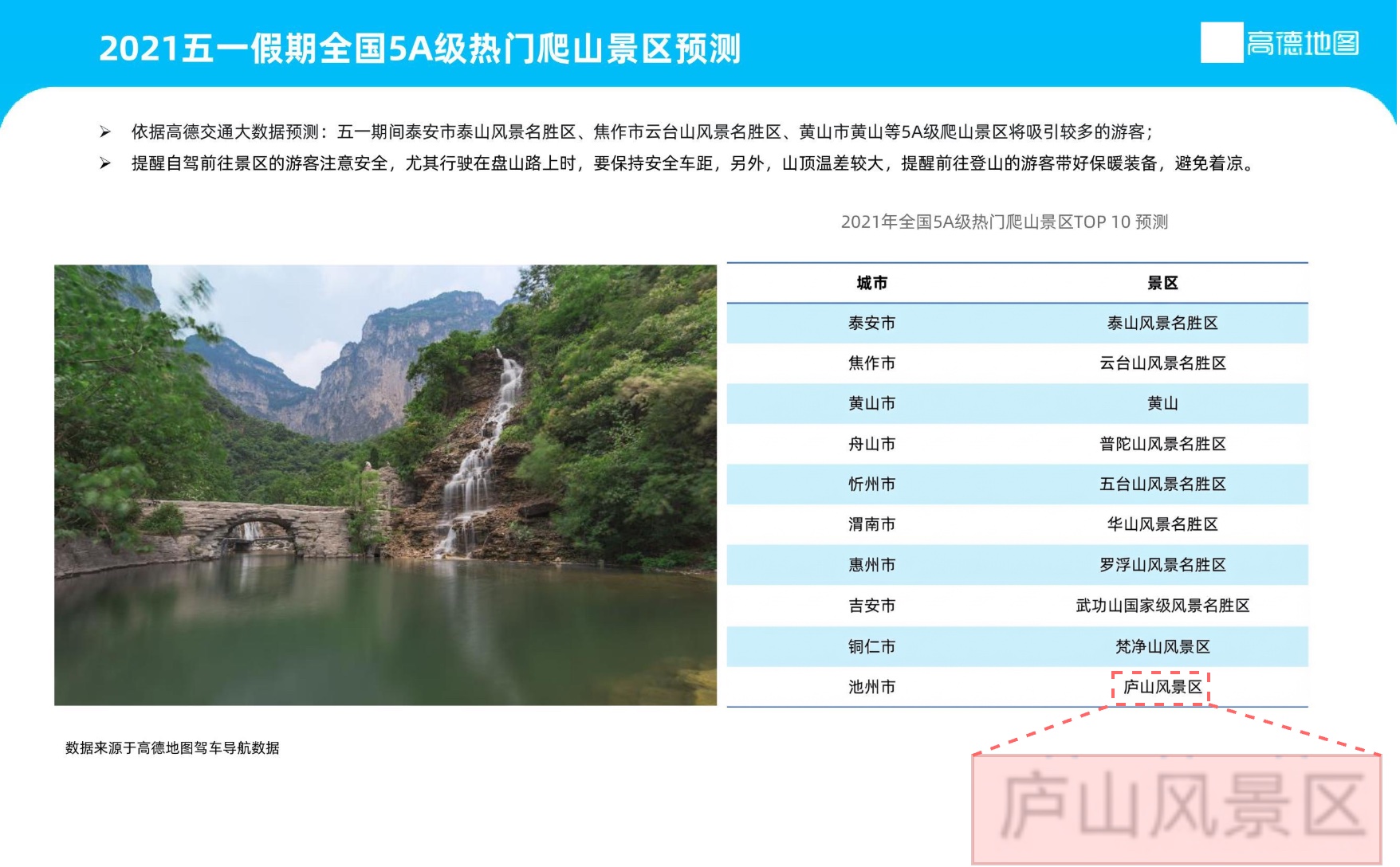} 
% 中文：将表格最后一行最后一列中的“九华山风景区”修改为“庐山风景区”。
\caption{Change the text \zh{“九华山风景区”} in the last column of the last row of the table to \zh{“庐山风景区”}.}
\end{minipage}
\par\addvspace{\intextsep}

\par\addvspace{\intextsep}\noindent
\begin{minipage}{\linewidth}\captionsetup{type=figure}
\centering
\includegraphics[width=0.9\linewidth,height=0.70\textheight,keepaspectratio]{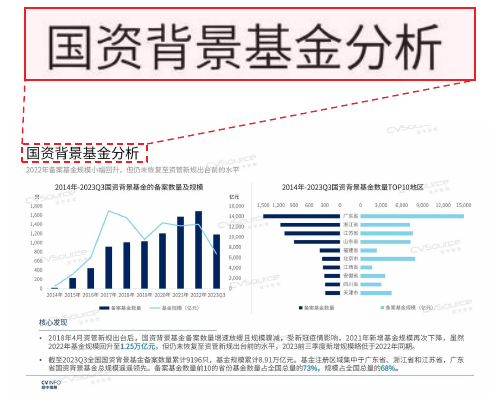}
% 中文：将标题“国资背景基金情况”替换为“国资背景基金分析”。
\caption{Replace the title \zh{“国资背景基金情况”} with \zh{“国资背景基金分析”}.}
\label{fig:case_48234}
\label{fig:extra_cases_1}
\end{minipage}
\par\addvspace{\intextsep}

\par\addvspace{\intextsep}\noindent
\begin{minipage}{\linewidth}\captionsetup{type=figure}
\centering
\includegraphics[width=0.9\linewidth,height=0.70\textheight,keepaspectratio]{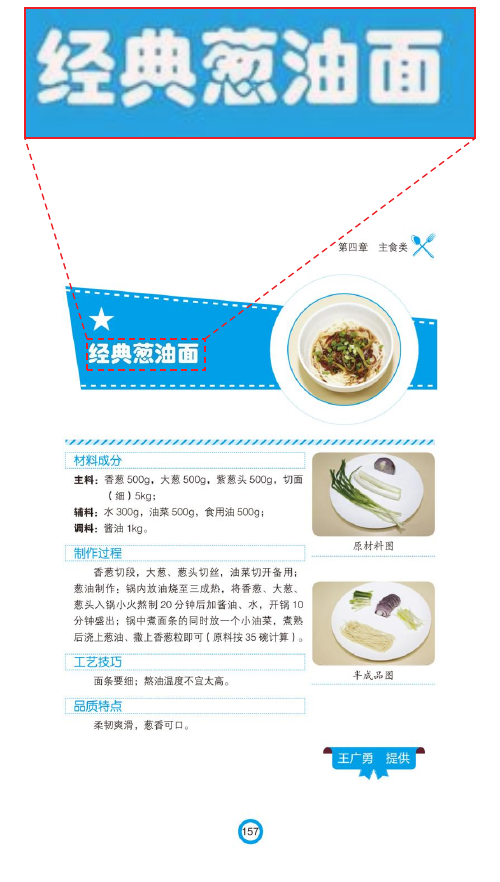}
% 中文：将标题“葱油面”改为“经典葱油面”。
\caption{Change the title \zh{“葱油面”} to \zh{“经典葱油面”}.}
\label{fig:case_48404}
\end{minipage}
\par\addvspace{\intextsep}

\par\addvspace{\intextsep}\noindent
\begin{minipage}{\linewidth}\captionsetup{type=figure}
\centering
\includegraphics[width=0.9\linewidth,height=0.70\textheight,keepaspectratio]{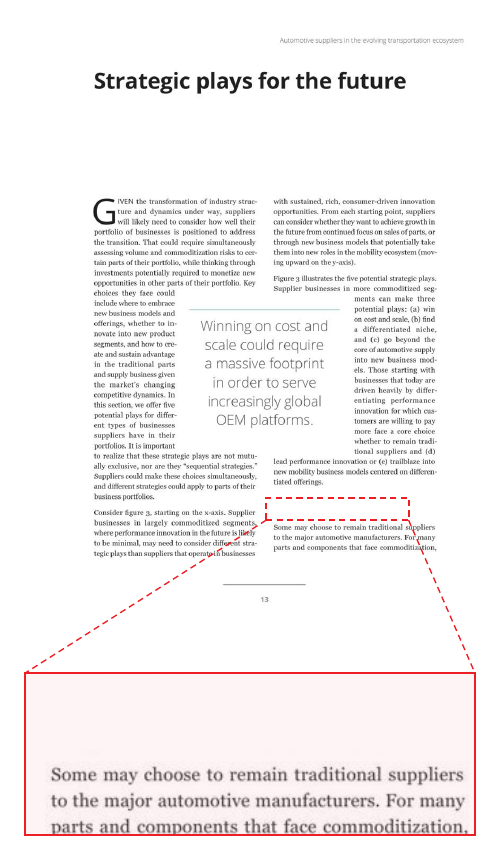}
% 中文：删除副标题“Win on cost and scale”，保留其余内容和版面布局。
\caption{Remove the subtitle ``Win on cost and scale'' while preserving the remaining content and page layout.}
\label{fig:case_49071}
\label{fig:extra_cases_2}
\end{minipage}
\par\addvspace{\intextsep}

\par\addvspace{\intextsep}\noindent
\begin{minipage}{\linewidth}\captionsetup{type=figure}
\centering
\includegraphics[width=0.9\linewidth,trim=0bp 93bp 0bp 17bp,clip,height=0.70\textheight,keepaspectratio]{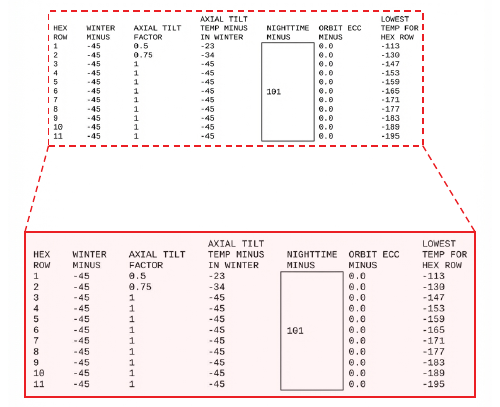}
% 中文：合并 NIGHTTIME MINUS 列第 1 至 11 个数据行的单元格。
\caption{Merge the cells in data rows 1--11 of the ``NIGHTTIME MINUS'' column.}
\label{fig:case_51068}
\end{minipage}
\par\addvspace{\intextsep}

\par\addvspace{\intextsep}\noindent
\begin{minipage}{\linewidth}\captionsetup{type=figure}
\centering
\includegraphics[width=0.9\linewidth,height=0.70\textheight,keepaspectratio]{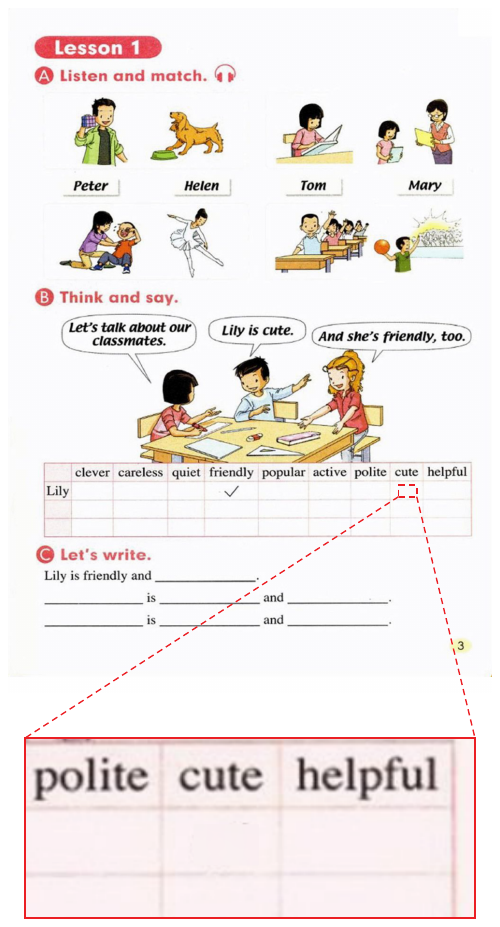}
% 中文：仅删除 Lily 行 cute 列中的勾选标记，保留表格线和其余勾选。
\caption{Remove only the checkmark in the Lily row under ``cute'', preserving the table grid and all other checkmarks.}
\label{fig:case_55454}
\end{minipage}
\par\addvspace{\intextsep}

\par\addvspace{\intextsep}\noindent
\begin{minipage}{\linewidth}\captionsetup{type=figure}
\centering
\includegraphics[width=0.9\linewidth,height=0.70\textheight,keepaspectratio]{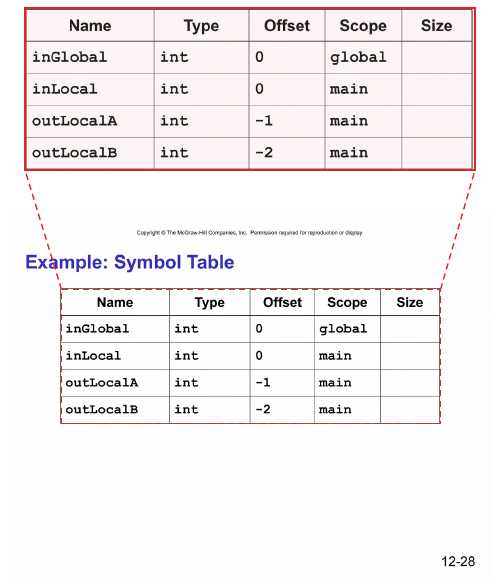}
% 中文：在 Scope 列右侧新增 Size 列。
\caption{Insert a ``Size'' column to the right of ``Scope''.}
\label{fig:case_51013}
\label{fig:extra_cases_3}
\end{minipage}
\par\addvspace{\intextsep}

% 中文：结束案例双栏排版后，将模型比较图置于下一页页首，余下空间接续指标附录。
\clearpage
\endgroup
\noindent
\begin{minipage}{\textwidth}\captionsetup{type=figure,skip=4pt}
\centering
\includegraphics[width=\linewidth,height=0.70\textheight,keepaspectratio]{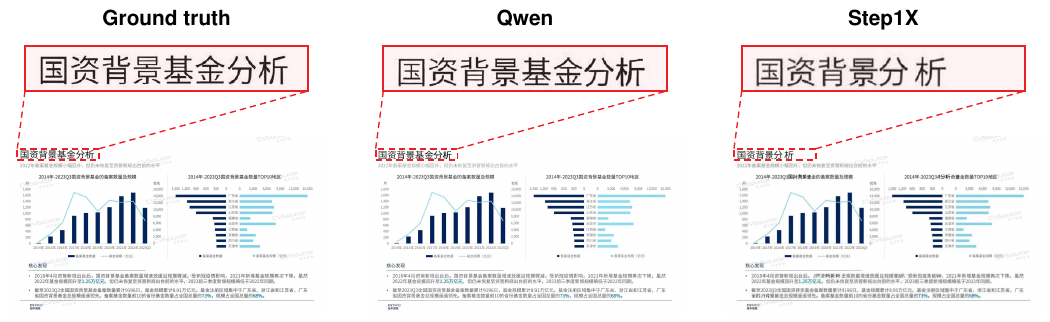}
% 中文：标题替换的实际模型输出。指令将“国资背景基金情况”改为“国资背景基金分析”（ID 48234）。Qwen 正确输出标题，但移除淡色斜向水印；Step1X 输出“国资背景分析”，漏掉“基金”二字，并改变周围文字。此图用于展示具体错误，不估计错误频率。
\caption{\textbf{Model outputs for title replacement.} The instruction replaces \zh{“国资背景基金情况”} with \zh{“国资背景基金分析”} (ID 48234). Qwen renders the requested title but removes faint diagonal watermarks. Step1X outputs \zh{“国资背景分析”}, omitting \zh{“基金”} (fund), and alters surrounding text. This example illustrates specific errors rather than estimating their frequency.}
\label{fig:model_case_title}
\end{minipage}

\vspace{8pt}

\FloatBarrier
% 中文标题：指标的形式化定义
\section{Formal Definition of Metrics}

% 中文标题：空间匹配算法
\subsection{Spatial Matching Algorithm}

% 中文标题：匹配策略
\subsubsection{Matching Strategy}
\label{sec: append match strategy}
% 中文：评测采用基于最大 IoU 的一对一贪心匹配。先计算所有参考框与预测框之间的 IoU，仅保留 IoU 不低于 0.1 的候选框对，再按 IoU 从高到低排序。
% 依次接受双方均未被使用的框对，每个参考框与预测框最多参与一次匹配。匹配依据是框的重叠程度，而非中心点距离。
We use greedy one-to-one matching based on bounding-box IoU. We compute IoU for every reference--prediction pair, retain candidates with IoU at least 0.1, and sort them in descending order. A pair is accepted if neither block has already been matched. Each reference and predicted block therefore participates in at most one match.

% 中文标题：匹配结果类别
\subsubsection{Matching Result Categories}
\label{sec: append match result}
% 中文：匹配结果分为匹配对、未匹配参考块和未匹配预测块。总体 IoU 包含匹配对与未匹配参考块，后者计 0；未匹配预测块单独记录数量，不进入 IoU 的分子或分母。
% TextSim、CDM 和 BLEU 仅对匹配块汇总；TEDS-like 仅对可计算该指标的匹配表格块汇总。
Matching produces matched pairs, unmatched reference blocks, and unmatched predicted blocks. The reported IoU includes matched pairs and unmatched reference blocks, assigning zero to the latter. Unmatched predicted blocks are counted separately and do not enter its numerator or denominator. TextSim, CDM, and BLEU are aggregated over matched blocks; TEDS-like is aggregated over eligible matched table blocks.

% 中文：空白编辑结果仍属于有效样本。若参考与预测均无 OCR 块，该图像对不产生块级条目；若仅参考非空，未匹配参考块计入 IoU 并取 0；若仅预测非空，则记录未匹配预测块，不增加 IoU 的参考框数量。
% 因此，图像对数量不同于各项指标实际汇总的块数量，图像指标仍评估空白区域的外观。
Blank editing outcomes remain valid samples. If neither image has OCR blocks, the pair contributes no block-level entries. If only the reference has blocks, those unmatched reference blocks contribute zero IoU. If only the prediction has blocks, they are recorded as unmatched predictions without increasing the reference-block count. Image-pair counts thus differ from the block counts used by each metric; image-based metrics still assess the appearance of blank regions.

% 中文标题：交并比
\subsubsection{Intersection over Union}
\label{sec: append iou}
% 中文：框的 IoU 定义为交集面积与并集面积之比；无有效并集面积时，实现返回 0。
For reference box $B_i^{\mathrm{GT}}$ and predicted box $B_j^{\mathrm{pred}}$, bounding-box IoU is
\begin{equation}
\mathrm{IoU}(B_i^{\mathrm{GT}},B_j^{\mathrm{pred}})
=\frac{|B_i^{\mathrm{GT}}\cap B_j^{\mathrm{pred}}|}
{|B_i^{\mathrm{GT}}\cup B_j^{\mathrm{pred}}|},
\end{equation}
where $|\cdot|$ denotes area; the implementation returns zero when the union area is nonpositive.
% 中文：令 M 为匹配对集合，N_GT 为参与该设置汇总的参考框总数，包括未匹配参考框。报告的 IoU 为匹配对 IoU 总和除以 N_GT；空汇总时实现返回 0。
Let $\mathcal{M}$ be the matched-pair set and $N_{\mathrm{GT}}$ the total number of reference blocks in the evaluated setting, including unmatched reference blocks. Writing $u_{ij}=\mathrm{IoU}(B_i^{\mathrm{GT}},B_j^{\mathrm{pred}})$, the reported score is
\begin{equation}
\mathrm{IoU}_{\mathrm{all}}=\frac{1}{N_{\mathrm{GT}}}\sum_{(i,j)\in\mathcal{M}}u_{ij},\quad N_{\mathrm{GT}}>0.
\end{equation}
The implementation returns zero for an empty aggregate ($N_{\mathrm{GT}}=0$).
% 中文：该指标刻画参考内容的定位重叠和覆盖程度。未匹配参考块保留在分母中，因此降低得分；未匹配预测块不额外降低得分。
This score measures localization overlap and coverage of reference content. Unmatched reference blocks remain in the denominator and lower the score; unmatched predictions introduce no additional penalty.

% 中文标题：文本内容指标
\subsection{Text Content Metrics}
\label{subsec:text content metrics}

% 中文标题：字符距离指标
\subsubsection{Character Distance Metric}
\label{subsubsec:CDM}

% 中文：字符距离指标（Character Distance Metric，CDM）基于 Levenshtein 距离 d_lev 定义。
% 该距离衡量将一个字符串转换为另一个字符串所需的最少字符级编辑操作数，包括插入、删除和替换。
The Character Distance Metric (CDM) is defined based on the Levenshtein distance $d_{\text{lev}}$ \cite{Levenshtein1965BinaryCC}, which measures the minimum number of character level edit operations (insertions, deletions, and substitutions) required to transform one string into another 
\begin{equation}
    \text{CDM}(c,r) = 1 - d_{\text{lev}}(c,r)/\max(|c|,|r|).
\end{equation}

% 中文：通过用所比较两个字符串的最大长度归一化编辑距离，CDM 得到一个介于 0 与 1 之间的相似度分数。
% 分数越高，表示生成文本 c 与参考文本 r 在字符层面越一致。
% 这一定义支持对长度不同的文本块进行公平比较。
By normalizing the edit distance with respect to the maximum length of the compared strings, CDM yields a similarity score bounded between 0 and 1, where higher values indicate greater character level consistency between the generated text $c$ and the reference text $r$. This formulation enables fair comparison across text blocks of varying lengths.

% 中文：CDM 能在字符层面对文本差异进行细粒度评估，因此对轻微修改错误、拼写错误和符号不匹配高度敏感，而基于词的指标可能无法充分捕捉这些问题。
CDM provides a fine grained evaluation of textual differences at the character level, making it highly sensitive to minor modification errors, typographical mistakes, or symbol mismatches that may not be adequately captured by word based metrics.

% 中文标题：BLEU-4 指标
\subsubsection{BLEU-4}
\label{subsubsec:BLEU-4}

% 中文：BLEU-4 使用 n=1 至 4 的修正 n 元组精确率，并结合加性平滑和长度惩罚，以处理生成文本与参考文本之间的长度差异，计算如下。
BLEU-4 \cite{papineni-etal-2002-bleu} is computed using modified $n$-gram precision for $n=1$ to $4$, combined with additive smoothing and a brevity penalty to account for length discrepancies between the generated and reference texts as
\begin{equation}
    \text{BLEU-4} = \text{BP} \cdot \exp\Big(\frac{1}{4} \sum_{n=1}^{4} \log P_n \Big).
\end{equation}

% 中文：长度惩罚（BP）用于防止过短输出获得较高的 BLEU 分数：若生成文本长度 c 大于参考文本长度 r，则 BP=1；否则，根据 r/c 对分数施加指数形式的衰减。
Brevity Penalty (BP) prevents short outputs from getting high BLEU scores: if the generated length $c > r, BP = 1$; otherwise, it reduces the score exponentially based on $r/c$.

% 中文：BLEU-4 在 n 元组层面评估生成文本与参考文本的一致程度，因而能够捕捉局部词序与短语层面的正确性。
% 结合 BLEU-4 与 CDM，可以同时从整体层面和字符层面评估模型的文本编辑能力。
BLEU-4 evaluates the degree of consistency between the generated text and the reference text at the n-gram level, thereby capturing local word order and phrase level correctness. By combining BLEU-4 and CDM, the model’s text editing capability can be evaluated both at the overall level and at the character level.

% 中文标题：类 TEDS 相似度
\subsubsection{TEDS-like Similarity}
\label{subsubsec:TEDS-like Similarity}

% 中文：遵循 TEDS 基于树编辑距离的定义，设 T_c 和 T_r 分别为候选结构化 OCR 输出与参考结构化 OCR 输出对应的有序树。
% 树节点编码结构元素及其关联文本内容；对于表格，表示中保留行与单元格的层级关系，以及跨行、跨列信息。
% 我们定义如下相似度。
Following the tree-edit-distance formulation of TEDS \cite{zhong2020imagebasedtablerecognitiondata}, let $T_c$ and $T_r$ be the ordered trees representing the candidate and reference structured OCR outputs. Their nodes encode structural elements and associated textual content; for tables, the representation retains the row/cell hierarchy and row/column spans. We define
\begin{equation}
\text{TEDS-like}(T_c,T_r)
=1-\frac{d_{\mathrm{TED}}(T_c,T_r)}{\max\{|T_c|,|T_r|\}},
\end{equation}
% 中文：其中，|T| 表示树 T 中的节点数；d_TED 表示将 T_c 转换为 T_r 所需的节点插入、删除和替换操作的最小总成本。
% 插入与删除的成本均为 1。
% 当结构节点类型或表格跨行、跨列属性不同时，替换成本为 1；对于类型兼容且承载文本的节点，替换成本为内容之间的归一化 Levenshtein 距离。
% 相同节点的替换成本为零。
where $|T|$ is the number of nodes in tree $T$, and $d_{\mathrm{TED}}$ is the minimum total cost of node insertions, deletions, and substitutions needed to transform $T_c$ into $T_r$. Insertions and deletions have unit cost. Substitutions have unit cost when structural node types or table span attributes differ; for compatible text-bearing nodes, the cost is the normalized Levenshtein distance between their contents. Identical nodes have zero substitution cost.

% 中文：这一定义联合比较层级结构与文本。
% 尤其是，d_TED 的操作对象是树，而非展平后的词元序列；仅依靠词元级字符串编辑距离无法衡量表格层级结构。
% 该分数补充了字符级 CDM 与空间 IoU，其含义限于 OCR 输出中实际表示出的结构。
This formulation compares hierarchical structure and text jointly. In particular, $d_{\mathrm{TED}}$ operates on trees rather than on flattened token sequences: a token-level string edit distance alone does not measure table hierarchy. The score complements character-level CDM and spatial IoU; its interpretation is limited to the structure represented by the OCR output.

% 中文标题：文本相似度
\subsubsection{Text Similarity (TextSim)}
\label{subsubsec:textsim}
% 中文：TextSim 使用 Python difflib.SequenceMatcher 计算字符匹配比例，参考文本 r 为第一个输入，预测文本 c 为第二个输入，保留默认 autojunk=True 设置。令 m 为该算法递归识别连续匹配子串后得到的匹配字符总数，则非空文本对的 TextSim 为 2m/(|r|+|c|)。两者均为空时取 1，仅一者为空时取 0。最终得分为成功匹配 OCR 块对的算术平均，汇总规则见匹配结果类别小节。
We compute TextSim with Python's \texttt{difflib.SequenceMatcher}, passing the reference text $r$ first and the predicted text $c$ second, with the default \texttt{autojunk=True} setting. Let $m$ be the total number of matching characters in the blocks identified by recursively matching contiguous subsequences. For $|r|+|c|>0$,
\begin{equation}
\operatorname{TextSim}(r,c)=\frac{2m}{|r|+|c|}.
\end{equation}
The score is 1 when both strings are empty and 0 when only one is empty. We report the arithmetic mean over matched OCR block pairs, following Appendix~\ref{sec: append match result}.

% 中文标题：图像指标
\subsection{Image Metrics}
\label{sec:image metrics}
% 中文标题：峰值信噪比
\subsubsection{PSNR}
% 中文：峰值信噪比（PSNR）量化编辑图像 I_edit 与真实参考图像 I_gt 之间的像素级保真度，定义如下。
The PSNR metric quantifies the pixel-level fidelity between the edited image $I_{\text{edit}}$ and the ground truth image $I_{\text{gt}}$:
\begin{equation}
\text{PSNR}(I_{\text{edit}}, I_{\text{gt}}) = 10 \cdot \log_{10}\left(\frac{\text{MAX}_I^2}{\text{MSE}(I_{\text{edit}}, I_{\text{gt}})}\right),
\end{equation}
% 中文：其中，MAX_I 表示像素可能达到的最大值，MSE 表示均方误差。
where $\text{MAX}_I$ represents the maximum possible pixel value and MSE denotes the mean squared error.

% 中文：PSNR 从信号层面提供重建准确性的基础度量，可作为评估低层视觉信息保持程度的基线。
% 虽然其计算简单且应用广泛，但由于依赖逐像素差异，并不区分错误在视觉上的重要性，因此可能无法充分反映感知质量。
% 不过，PSNR 仍有助于量化偏差幅度，并确保编辑图像在最基础层面保持结构完整性。
PSNR provides a fundamental measure of reconstruction accuracy at the signal level, serving as a baseline for assessing the preservation of low-level visual information. While computationally straightforward and widely adopted, PSNR's reliance on pixel-wise differences means it may not fully capture perceptual quality, as it treats all errors equally regardless of their visual significance. Nevertheless, it remains valuable for quantifying the magnitude of deviations and ensuring that edited images maintain structural integrity at the most basic level.

% 中文标题：SSIM（结构相似性指数）
\subsubsection{SSIM (Structural Similarity Index)}
% 中文：结构相似性指数（SSIM）通过结合局部亮度 l、对比度 c 和结构 s 三个组成部分，比较编辑图像与参考图像，计算如下。
The Structural Similarity Index compares the edited and reference images by combining local luminance $l$, contrast $c$, and structure $s$ components \citep{wang2004image}:
\begin{equation}
\begin{aligned}
\text{SSIM}(I_{\text{edit}}, I_{\text{gt}}) &= l(I_{\text{edit}}, I_{\text{gt}})\\
&\quad\cdot c(I_{\text{edit}}, I_{\text{gt}})\\
&\quad\cdot s(I_{\text{edit}}, I_{\text{gt}}).
\end{aligned}
\end{equation}

% 中文：与 PSNR 的像素级方法不同，SSIM 建立在人类视觉善于从视觉场景中提取结构信息这一原理之上。
% 它对相邻像素的相互依赖进行建模，并考虑图像局部区域内的组织模式，从而提供更符合感知特性的图像质量评估。
% 因此，SSIM 特别适合评估纹理保持、边缘清晰度保持以及整体结构连贯性；这些属性对于强调布局完整性和视觉一致性的文档图像尤为重要。
Unlike PSNR's pixel-level approach, SSIM operates on the principle that human perception is highly adapted to extract structural information from visual scenes. By modeling the interdependencies between neighboring pixels and considering the organizational patterns within local image regions, SSIM provides a more perceptually relevant assessment of image quality. This makes it particularly effective for evaluating edits that involve texture preservation, edge sharpness maintenance, and overall structural coherence—attributes that are crucial for document images where layout integrity and visual consistency are paramount.

% 中文标题：LPIPS（学习感知图像块相似度）
\subsubsection{LPIPS (Learned Perceptual Image Patch Similarity)}
% 中文：学习感知图像块相似度（LPIPS）利用归一化深度特征与学习到的通道权重测量感知差异，定义如下。
LPIPS measures perceptual differences with normalized deep features and learned channel weights \citep{zhang2018unreasonable}:
\begin{equation}
\begin{split}
\text{LPIPS}(I_{\text{edit}}, I_{\text{gt}})\\
= \sum_l \frac{1}{H_lW_l}\sum_{h,w}\\
\left\|w_l\odot\left(\hat\phi_l(I_{\text{edit}})_{hw}-\hat\phi_l(I_{\text{gt}})_{hw}\right)\right\|_2^2.
\end{split}
\end{equation}
% 中文：其中，归一化特征 φ̂_l 来自第 l 层，w_l 为学习到的通道权重，H_l 与 W_l 为该层空间尺寸；⊙ 表示逐元素乘法。
Here $\hat\phi_l$ denotes normalized layer-$l$ features, $w_l$ learned channel weights, and $H_l,W_l$ the spatial dimensions; $\odot$ denotes elementwise multiplication.

% 中文：LPIPS 使用学习到的深度特征距离比较图像外观，作为像素误差与结构相似性的补充。本文分别在整页和编辑区域上计算该指标，以考察视觉变化。
LPIPS compares image appearance through learned deep-feature distances, complementing pixel error and structural similarity. We compute it on both full pages and edited regions to characterize visual changes.

% 中文标题：CLIP（对比式语言—图像预训练）
\subsubsection{CLIP (Contrastive Language-Image Pre-training)}
% 中文：本文使用 CLIP 图像编码器与文本编码器，计算编辑图像与其配对文本的特征余弦相似度。
We use the CLIP image and text encoders \citep{radford2021learning} to measure image--text alignment between the edited image $I_{\text{edit}}$ and its paired text $t$:
\begin{equation}
\begin{split}
\text{CLIP-Sim}(I_{\text{edit}}, t)\\
=\frac{\phi_I(I_{\text{edit}})\cdot\phi_T(t)}{\|\phi_I(I_{\text{edit}})\|\,\|\phi_T(t)\|}.
\end{split}
\end{equation}
% 中文：其中，φ_I 与 φ_T 分别为 CLIP 图像编码器和文本编码器；t 表示评测时与编辑图像配对的文本。
Here $\phi_I$ and $\phi_T$ are the CLIP image and text encoders, respectively, and $t$ denotes the text paired with the edited image during evaluation.

% 中文：该指标衡量 CLIP 共享表征空间中的图文语义对齐，补充像素级和感知距离。文字是否被精确修改，则由 OCR 内容指标单独检查。
This metric measures image--text semantic alignment in CLIP's shared representation space, complementing pixel-level and perceptual distances. Exact textual changes are checked separately by the OCR content metrics.

% 中文标题：更多分析
\section{More Analysis}
\label{sec: append more analysis}

% 中文标题：OCR 引擎对照
\subsection{OCR-Engine Comparison}
\label{sec:ocr_engine_check}
% 中文：我们使用 Hunyuan-OCR 对 3,506 条样本版本中七个编辑模型的参考图像和生成图像进行 OCR 评测。两表分别报告局部和全局结果，并给出各模型全部有效样本与七个模型共同的 3,469 条样本两种口径。各模型有效图像对数量为 3,487 至 3,499；未能完成解析的样本不计入汇总，也不赋零分。
We evaluate the reference and generated images of all seven editors in the 3,506-sample release with Hunyuan-OCR. Tables~\ref{tab:hunyuan_local} and~\ref{tab:hunyuan_global} report local and global results, respectively, using both all available pairs for each model and a common subset of 3,469 samples shared by all seven models. Model-specific coverage ranges from 3,487 to 3,499 image pairs. Unresolved parsing failures are excluded from aggregation rather than assigned zero scores.

% 中文：Hunyuan 的全局指标使用整页 OCR 结果，局部指标保留与标注编辑区域重叠的 OCR 块，并沿用相同的块匹配和指标实现（IoU 匹配阈值为 0.1）。未成功完成整页解析的图像进一步采用版面检测和逐块识别，逐块识别的最大生成长度为 512 token；触及生成上限但已保存的解码结果仍纳入计算。IoU 在参考块上平均，TextSim、CDM 和 BLEU 在匹配块上平均，TEDS 在符合条件的匹配表格块上平均；n 不表示每项指标的块数。
For Hunyuan-OCR, global evaluation uses full-page OCR outputs, while local evaluation retains OCR blocks overlapping the annotated editing region. We reuse the block-matching and metric implementations with an IoU matching threshold of 0.1. Images that fail full-page parsing undergo layout detection followed by block-wise recognition with a maximum of 512 generated tokens per block; saved decoded outputs that reach the token cap are retained. IoU is averaged over reference blocks, TextSim, CDM, and BLEU over matched blocks, and TEDS over eligible matched table blocks. Thus, $n$ denotes image pairs rather than the number of blocks contributing to each metric.

% 中文：VDE Bench 的主要排名结论在两种 OCR 评测下保持稳定。共同样本上的 Hunyuan 排名与正文 PaddleOCR-VL 主表的 Spearman 相关系数为 0.893--1.000，十组指标的平均值为 0.971。Qwen 在全部五项全局指标以及四项局部指标上仍然领先。唯一的榜首变化是局部 IoU，LongCat 为 0.6849，Qwen 为 0.6841，二者相差 0.0008。五组指标的完整排名相同，其余变化涉及少数模型。因此，这些结果支持主要相对表现结论对 OCR 模型选择具有稳健性，同时也表明个别指标的精细名次会变化。此处比较的是 Hunyuan 共同样本与正文全量结果，故也包含样本覆盖和局部 OCR 提取流程的差异。
The main ranking conclusions of VDE Bench remain stable across the two OCR evaluations. Comparing the common-subset Hunyuan rankings with the PaddleOCR-VL main table yields Spearman correlations of 0.893--1.000, averaging 0.971 over the ten metric-setting combinations. Qwen remains first on all five global metrics and four local metrics. The sole change at the top is local IoU, where LongCat scores 0.6849 versus Qwen's 0.6841, a gap of 0.0008. Five combinations preserve the complete ordering; the remaining changes involve a small number of models. These results support the robustness of the main relative-performance conclusions to OCR model choice, while individual metric ranks can vary. This comparison uses the Hunyuan common subset and the full main-table results, and therefore also includes differences in sample coverage and local OCR extraction.

% 中文：另外，在原始研究中，我们人工检查了 200 张生成图像上的 PaddleOCR-VL 字符识别结果，观测到的字符准确率约为 99.7%。这项检查补充了基准构建过程中对参考文本的人工核验。
Separately, manual inspection of PaddleOCR-VL outputs on 200 generated images in the original study records approximately 99.7\% character accuracy, complementing the verification of reference text during benchmark construction.

\begin{table*}[!t]
\centering
\fontsize{8}{9.5}\selectfont
\setlength{\tabcolsep}{2.2pt}
\renewcommand{\arraystretch}{1.12}
\begin{tabularx}{\textwidth}{@{}l r *{5}{>{\centering\arraybackslash}X}@{}}
\toprule
 & & \multicolumn{5}{c}{\textbf{OCR metrics} $\uparrow$} \\
\cmidrule(lr){3-7}
Model & $n$ & IoU & TextSim & CDM & BLEU & TEDS \\
\midrule
\multicolumn{7}{l}{\textbf{All available pairs}} \\
\midrule
Step1X & 3,498 & \cellcolor[HTML]{97CFEE} 0.6424 & \cellcolor[HTML]{C7E5F6} 0.4666 & \cellcolor[HTML]{BDE0F4} 0.5520 & \cellcolor[HTML]{DBEEF9} 0.1998 & \cellcolor[HTML]{BCE0F4} 0.5039 \\
LongCat & 3,487 & \cellcolor[HTML]{84C6EB} \textbf{0.6817} & \cellcolor[HTML]{91CCED} 0.7388 & \cellcolor[HTML]{88C8EC} 0.7925 & \cellcolor[HTML]{9BD0EF} 0.4571 & \cellcolor[HTML]{8ECBED} 0.7089 \\
FireRed & 3,492 & \cellcolor[HTML]{92CCED} 0.6529 & \cellcolor[HTML]{90CCED} 0.7411 & \cellcolor[HTML]{8BC9EC} 0.7812 & \cellcolor[HTML]{9BD0EF} 0.4574 & \cellcolor[HTML]{9AD0EF} 0.6573 \\
Qwen & 3,489 & \cellcolor[HTML]{86C7EB} 0.6781 & \cellcolor[HTML]{84C6EB} \textbf{0.8036} & \cellcolor[HTML]{84C6EB} \textbf{0.8132} & \cellcolor[HTML]{84C6EB} \textbf{0.5481} & \cellcolor[HTML]{84C6EB} \textbf{0.7541} \\
Instruct & 3,499 & \cellcolor[HTML]{FFFFFF} 0.4256 & \cellcolor[HTML]{FFFFFF} 0.1806 & \cellcolor[HTML]{FFFFFF} 0.2451 & \cellcolor[HTML]{FFFFFF} 0.0536 & \cellcolor[HTML]{FFFFFF} 0.2040 \\
HiDream & 3,499 & \cellcolor[HTML]{F1F8FD} 0.4552 & \cellcolor[HTML]{B0DAF2} 0.5814 & \cellcolor[HTML]{AAD8F1} 0.6384 & \cellcolor[HTML]{BBDFF4} 0.3273 & \cellcolor[HTML]{B4DCF3} 0.5395 \\
JoyAI & 3,495 & \cellcolor[HTML]{99D0EE} 0.6390 & \cellcolor[HTML]{93CDED} 0.7299 & \cellcolor[HTML]{8BC9EC} 0.7787 & \cellcolor[HTML]{98CFEE} 0.4674 & \cellcolor[HTML]{91CCED} 0.6972 \\
\midrule
\multicolumn{7}{l}{\textbf{Common subset across all seven models}} \\
\midrule
Step1X & 3,469 & \cellcolor[HTML]{95CEEE} 0.6494 & \cellcolor[HTML]{C7E5F6} 0.4674 & \cellcolor[HTML]{BDE0F4} 0.5539 & \cellcolor[HTML]{DBEEF9} 0.2005 & \cellcolor[HTML]{BCE0F4} 0.5039 \\
LongCat & 3,469 & \cellcolor[HTML]{84C6EB} \textbf{0.6849} & \cellcolor[HTML]{91CCED} 0.7409 & \cellcolor[HTML]{89C8EC} 0.7948 & \cellcolor[HTML]{9BD1EF} 0.4585 & \cellcolor[HTML]{8ECBED} 0.7091 \\
FireRed & 3,469 & \cellcolor[HTML]{91CCED} 0.6583 & \cellcolor[HTML]{90CCED} 0.7431 & \cellcolor[HTML]{8BC9EC} 0.7839 & \cellcolor[HTML]{9BD1EF} 0.4589 & \cellcolor[HTML]{9AD0EF} 0.6571 \\
Qwen & 3,469 & \cellcolor[HTML]{84C6EB} 0.6841 & \cellcolor[HTML]{84C6EB} \textbf{0.8067} & \cellcolor[HTML]{84C6EB} \textbf{0.8166} & \cellcolor[HTML]{84C6EB} \textbf{0.5510} & \cellcolor[HTML]{84C6EB} \textbf{0.7544} \\
Instruct & 3,469 & \cellcolor[HTML]{FFFFFF} 0.4310 & \cellcolor[HTML]{FFFFFF} 0.1806 & \cellcolor[HTML]{FFFFFF} 0.2457 & \cellcolor[HTML]{FFFFFF} 0.0535 & \cellcolor[HTML]{FFFFFF} 0.2042 \\
HiDream & 3,469 & \cellcolor[HTML]{F1F8FD} 0.4600 & \cellcolor[HTML]{B0DAF2} 0.5831 & \cellcolor[HTML]{AAD8F1} 0.6411 & \cellcolor[HTML]{BBE0F4} 0.3284 & \cellcolor[HTML]{B4DCF3} 0.5405 \\
JoyAI & 3,469 & \cellcolor[HTML]{97CFEE} 0.6464 & \cellcolor[HTML]{93CDED} 0.7314 & \cellcolor[HTML]{8CCAEC} 0.7808 & \cellcolor[HTML]{98CFEE} 0.4694 & \cellcolor[HTML]{91CCED} 0.6974 \\
\midrule
Spearman $\rho$ vs.\ main table & --- & 0.929 & 1.000 & 0.964 & 0.964 & 1.000 \\
\bottomrule
\end{tabularx}
% 中文：Hunyuan-OCR 的局部结果。上半部分使用各模型全部有效的参考/输出图像对，下半部分对所有模型使用共同的 3,469 条样本。n 表示图像对数；蓝色越深表示各部分同一指标的分数越高，加粗表示最佳值。秩相关比较共同样本结果与正文 3,506 条样本主表。
\caption{\textbf{Local evaluation with Hunyuan-OCR.} The upper panel uses all available reference/output pairs for each model; the lower panel uses the same 3,469 samples for all seven models. $n$ counts image pairs. Darker blue indicates higher scores within each metric and panel; bold marks the best score. All metrics are higher-is-better. Rank correlations compare the common-subset results with the PaddleOCR-VL main table on 3,506 samples.}
\label{tab:hunyuan_local}
\label{tab:ocr_engines}
\end{table*}

\begin{table*}[!t]
\centering
\fontsize{8}{9.5}\selectfont
\setlength{\tabcolsep}{2.2pt}
\renewcommand{\arraystretch}{1.12}
\begin{tabularx}{\textwidth}{@{}l r *{5}{>{\centering\arraybackslash}X}@{}}
\toprule
 & & \multicolumn{5}{c}{\textbf{OCR metrics} $\uparrow$} \\
\cmidrule(lr){3-7}
Model & $n$ & IoU & TextSim & CDM & BLEU & TEDS \\
\midrule
\multicolumn{7}{l}{\textbf{All available pairs}} \\
\midrule
Step1X & 3,498 & \cellcolor[HTML]{94CDEE} 0.6708 & \cellcolor[HTML]{A8D7F1} 0.6840 & \cellcolor[HTML]{A3D4F0} 0.7102 & \cellcolor[HTML]{C0E2F5} 0.2995 & \cellcolor[HTML]{B3DCF3} 0.5222 \\
LongCat & 3,487 & \cellcolor[HTML]{88C8EC} 0.7107 & \cellcolor[HTML]{90CCED} 0.8195 & \cellcolor[HTML]{8ECBED} 0.8282 & \cellcolor[HTML]{9FD3EF} 0.4109 & \cellcolor[HTML]{8CCAEC} 0.7067 \\
FireRed & 3,492 & \cellcolor[HTML]{92CDED} 0.6764 & \cellcolor[HTML]{88C8EC} 0.8662 & \cellcolor[HTML]{87C7EC} 0.8678 & \cellcolor[HTML]{8DCAED} 0.4733 & \cellcolor[HTML]{91CCED} 0.6840 \\
Qwen & 3,489 & \cellcolor[HTML]{84C6EB} \textbf{0.7265} & \cellcolor[HTML]{84C6EB} \textbf{0.8902} & \cellcolor[HTML]{84C6EB} \textbf{0.8862} & \cellcolor[HTML]{84C6EB} \textbf{0.5056} & \cellcolor[HTML]{84C6EB} \textbf{0.7453} \\
Instruct & 3,499 & \cellcolor[HTML]{FFFFFF} 0.2923 & \cellcolor[HTML]{FFFFFF} 0.1825 & \cellcolor[HTML]{FFFFFF} 0.1844 & \cellcolor[HTML]{FFFFFF} 0.0800 & \cellcolor[HTML]{FFFFFF} 0.1649 \\
HiDream & 3,499 & \cellcolor[HTML]{D0E9F7} 0.4584 & \cellcolor[HTML]{9FD3EF} 0.7321 & \cellcolor[HTML]{9CD1EF} 0.7481 & \cellcolor[HTML]{BCE0F4} 0.3132 & \cellcolor[HTML]{B0DBF2} 0.5360 \\
JoyAI & 3,495 & \cellcolor[HTML]{B7DEF3} 0.5450 & \cellcolor[HTML]{8FCBED} 0.8244 & \cellcolor[HTML]{8DCAED} 0.8334 & \cellcolor[HTML]{97CFEE} 0.4406 & \cellcolor[HTML]{8FCBED} 0.6916 \\
\midrule
\multicolumn{7}{l}{\textbf{Common subset across all seven models}} \\
\midrule
Step1X & 3,469 & \cellcolor[HTML]{91CCED} 0.6980 & \cellcolor[HTML]{A8D7F1} 0.6909 & \cellcolor[HTML]{A3D4F0} 0.7175 & \cellcolor[HTML]{C0E2F5} 0.3028 & \cellcolor[HTML]{B3DCF3} 0.5230 \\
LongCat & 3,469 & \cellcolor[HTML]{8AC9EC} 0.7236 & \cellcolor[HTML]{90CCED} 0.8306 & \cellcolor[HTML]{8ECBED} 0.8397 & \cellcolor[HTML]{9FD3EF} 0.4169 & \cellcolor[HTML]{8CCAEC} 0.7081 \\
FireRed & 3,469 & \cellcolor[HTML]{91CCED} 0.6970 & \cellcolor[HTML]{88C8EC} 0.8751 & \cellcolor[HTML]{87C8EC} 0.8770 & \cellcolor[HTML]{8ECAED} 0.4788 & \cellcolor[HTML]{91CCED} 0.6847 \\
Qwen & 3,469 & \cellcolor[HTML]{84C6EB} \textbf{0.7451} & \cellcolor[HTML]{84C6EB} \textbf{0.9006} & \cellcolor[HTML]{84C6EB} \textbf{0.8970} & \cellcolor[HTML]{84C6EB} \textbf{0.5124} & \cellcolor[HTML]{84C6EB} \textbf{0.7461} \\
Instruct & 3,469 & \cellcolor[HTML]{FFFFFF} 0.3063 & \cellcolor[HTML]{FFFFFF} 0.1828 & \cellcolor[HTML]{FFFFFF} 0.1849 & \cellcolor[HTML]{FFFFFF} 0.0802 & \cellcolor[HTML]{FFFFFF} 0.1654 \\
HiDream & 3,469 & \cellcolor[HTML]{CFE9F7} 0.4765 & \cellcolor[HTML]{A0D3F0} 0.7381 & \cellcolor[HTML]{9DD1EF} 0.7547 & \cellcolor[HTML]{BCE0F4} 0.3146 & \cellcolor[HTML]{B0DAF2} 0.5380 \\
JoyAI & 3,469 & \cellcolor[HTML]{B7DEF3} 0.5632 & \cellcolor[HTML]{8FCBED} 0.8352 & \cellcolor[HTML]{8DCAEC} 0.8445 & \cellcolor[HTML]{97CFEE} 0.4467 & \cellcolor[HTML]{8FCBED} 0.6934 \\
\midrule
Spearman $\rho$ vs.\ main table & --- & 0.964 & 1.000 & 1.000 & 1.000 & 0.893 \\
\bottomrule
\end{tabularx}
% 中文：Hunyuan-OCR 的全局结果。上半部分使用各模型全部有效的参考/输出图像对，下半部分对所有模型使用共同的 3,469 条样本。n 表示图像对数；蓝色越深表示各部分同一指标的分数越高，加粗表示最佳值。秩相关比较共同样本结果与正文 3,506 条样本主表。
\caption{\textbf{Global evaluation with Hunyuan-OCR.} The upper panel uses all available reference/output pairs for each model; the lower panel uses the same 3,469 samples for all seven models. $n$ counts image pairs. Darker blue indicates higher scores within each metric and panel; bold marks the best score. All metrics are higher-is-better. Rank correlations compare the common-subset results with the PaddleOCR-VL main table on 3,506 samples.}
\label{tab:hunyuan_global}
\end{table*}

% 中文标题：不同指令类型具有不同程度的难度
\subsection{Different instruction types pose varying levels of difficulty}
\label{sec:appendix instruction types analysis}
% 中文：正文展示了同时考虑全局与局部指标的平均结果；全局编辑类型表和局部编辑类型表则分别报告两个设置下的指标结果。
In the main text, we present the averaged results considering both global and local metrics, while Table ~\ref{tab:edit_type_global} and Table ~\ref{tab:edit_type_local} report the global and local metric results separately.

% 中文：按编辑类型划分的局部图像结果。
% 所有编辑模型均接收源页面和指令；局部评测从生成页面中裁剪区域进行评分。
% 在局部编辑类型表中，FireRed 的修改操作 IoU（0.899）和 CDM（0.886）最高，添加操作 IoU（0.603）也最高。
% Qwen 在删除和表格结构编辑的四项指标上均领先，其中结构编辑的 IoU 与 TEDS 分别为 0.907 和 0.750。
% 这些结果区分了不同操作下的定位能力与内容保真度。
\paragraph{Local-image results by edit type.}
All editors receive the source page and instruction; local evaluation crops the generated page for scoring. In Table~\ref{tab:edit_type_local}, FireRed leads modification IoU (0.899) and CDM (0.886), as well as addition IoU (0.603). Qwen leads all four deletion and table-structure metrics; its structural IoU and TEDS are 0.907 and 0.750. These results separate localization and content fidelity across operations.

% 中文：按编辑类型划分的全局图像结果。
% 全局编辑类型表对相同模型输出的整页进行评测。
% LongCat 的修改操作 IoU 最高，为 0.913；Qwen 在修改操作的其他指标，以及添加、删除和结构编辑的全部四项指标上均领先。
% 对于结构编辑，Qwen 的 IoU、CDM、BLEU 和 TEDS 分别达到 0.900、0.925、0.615 和 0.756。
% 局部与全局分数分别描述编辑区域本身及其完整文档上下文。
\paragraph{Global-image results by edit type.}
Table~\ref{tab:edit_type_global} evaluates the same outputs over the full page. LongCat leads modification IoU (0.913), while Qwen leads the other modification metrics and all four addition, deletion, and structure metrics. For structural edits, Qwen reaches IoU of 0.900, CDM of 0.925, BLEU of 0.615, and TEDS of 0.756. Local and global scores describe the edited region and its full-document context, respectively.

% 中文：按当前3506条数据的操作子集、官方块级分母汇总；均值设置先分别计算全局和局部，再等权平均。
\begin{table*}[tbp]
\centering\small
\setlength{\tabcolsep}{3pt}
\renewcommand{\arraystretch}{1.1}
\begin{tabularx}{\textwidth}{@{}l*{8}{>{\centering\arraybackslash}X}@{}}
\toprule
\textbf{Metric} & \textbf{Step1X} & \textbf{LongCat} & \textbf{FireRed} & \textbf{Qwen} & \textbf{Instruct} & \textbf{HiDream} & \textbf{JoyAI} & \textbf{Avg.} \\
\midrule
\multicolumn{9}{l}{\textbf{Text Modify} (1099 samples)} \\
\midrule
IoU & 0.832 & \cellcolor[HTML]{D6ECE8}\textbf{0.913} & 0.896 & 0.905 & 0.392 & 0.589 & 0.719 & \cellcolor[HTML]{F6E6C9} 0.750 \\
CDM & 0.605 & 0.736 & 0.812 & \cellcolor[HTML]{D6ECE8}\textbf{0.850} & 0.213 & 0.748 & 0.751 & \cellcolor[HTML]{F6E6C9} 0.674 \\
BLEU & 0.274 & 0.344 & 0.397 & \cellcolor[HTML]{D6ECE8}\textbf{0.439} & 0.120 & 0.322 & 0.366 & \cellcolor[HTML]{F6E6C9} 0.323 \\
TEDS & 0.558 & 0.749 & 0.793 & \cellcolor[HTML]{D6ECE8}\textbf{0.833} & 0.148 & 0.602 & 0.764 & \cellcolor[HTML]{F6E6C9} 0.635 \\
\midrule
\multicolumn{9}{l}{\textbf{Text Addition} (976 samples)} \\
\midrule
IoU & 0.803 & 0.840 & 0.842 & \cellcolor[HTML]{D6ECE8}\textbf{0.899} & 0.531 & 0.591 & 0.644 & \cellcolor[HTML]{F6E6C9} 0.736 \\
CDM & 0.622 & 0.831 & 0.903 & \cellcolor[HTML]{D6ECE8}\textbf{0.933} & 0.187 & 0.737 & 0.833 & \cellcolor[HTML]{F6E6C9} 0.721 \\
BLEU & 0.303 & 0.446 & 0.541 & \cellcolor[HTML]{D6ECE8}\textbf{0.602} & 0.091 & 0.311 & 0.459 & \cellcolor[HTML]{F6E6C9} 0.393 \\
TEDS & 0.564 & 0.826 & 0.864 & \cellcolor[HTML]{D6ECE8}\textbf{0.903} & 0.148 & 0.662 & 0.782 & \cellcolor[HTML]{F6E6C9} 0.679 \\
\midrule
\multicolumn{9}{l}{\textbf{Text Deletion} (971 samples)} \\
\midrule
IoU & 0.843 & 0.858 & 0.838 & \cellcolor[HTML]{D6ECE8}\textbf{0.897} & 0.558 & 0.488 & 0.691 & \cellcolor[HTML]{F6E6C9} 0.739 \\
CDM & 0.686 & 0.850 & 0.912 & \cellcolor[HTML]{D6ECE8}\textbf{0.930} & 0.188 & 0.794 & 0.842 & \cellcolor[HTML]{F6E6C9} 0.743 \\
BLEU & 0.350 & 0.484 & 0.577 & \cellcolor[HTML]{D6ECE8}\textbf{0.624} & 0.092 & 0.404 & 0.486 & \cellcolor[HTML]{F6E6C9} 0.431 \\
TEDS & 0.620 & 0.814 & 0.843 & \cellcolor[HTML]{D6ECE8}\textbf{0.882} & 0.155 & 0.639 & 0.802 & \cellcolor[HTML]{F6E6C9} 0.679 \\
\midrule
\multicolumn{9}{l}{\textbf{Table Structure Edit} (460 samples)} \\
\midrule
IoU & 0.855 & 0.856 & 0.769 & \cellcolor[HTML]{D6ECE8}\textbf{0.900} & 0.553 & 0.527 & 0.672 & \cellcolor[HTML]{F6E6C9} 0.733 \\
CDM & 0.655 & 0.812 & 0.850 & \cellcolor[HTML]{D6ECE8}\textbf{0.925} & 0.185 & 0.730 & 0.857 & \cellcolor[HTML]{F6E6C9} 0.716 \\
BLEU & 0.321 & 0.422 & 0.500 & \cellcolor[HTML]{D6ECE8}\textbf{0.615} & 0.091 & 0.302 & 0.489 & \cellcolor[HTML]{F6E6C9} 0.391 \\
TEDS & 0.380 & 0.668 & 0.508 & \cellcolor[HTML]{D6ECE8}\textbf{0.756} & 0.154 & 0.426 & 0.678 & \cellcolor[HTML]{F6E6C9} 0.510 \\
\bottomrule
\end{tabularx}
% 中文：3,506 条样本上按编辑类型划分的表现。
% 所有分数均采用全局图像设置。
% Text Modify 合并替换与修改；添加和删除均包含文本区域与表格区域。
% Avg. 为七种编辑模型的平均值。
% 青绿色及加粗标记每行最佳编辑模型。
\caption{\textbf{Performance by edit type on 3,506 samples.} All scores use the global-image setting. Text Modify combines replacement and modification; addition and deletion include text and table regions. Avg. is the mean over the seven editing models. Teal and bold mark the best editing model per row.}
\label{tab:edit_type_global}
\end{table*}

% 中文：按当前3506条数据的操作子集、官方块级分母汇总；均值设置先分别计算全局和局部，再等权平均。
\begin{table*}[tbp]
\centering\small
\setlength{\tabcolsep}{3pt}
\renewcommand{\arraystretch}{1.1}
\begin{tabularx}{\textwidth}{@{}l*{8}{>{\centering\arraybackslash}X}@{}}
\toprule
\textbf{Metric} & \textbf{Step1X} & \textbf{LongCat} & \textbf{FireRed} & \textbf{Qwen} & \textbf{Instruct} & \textbf{HiDream} & \textbf{JoyAI} & \textbf{Avg.} \\
\midrule
\multicolumn{9}{l}{\textbf{Text Modify} (1099 samples)} \\
\midrule
IoU & 0.797 & 0.862 & \cellcolor[HTML]{D6ECE8}\textbf{0.899} & 0.876 & 0.592 & 0.645 & 0.772 & \cellcolor[HTML]{F6E6C9} 0.778 \\
CDM & 0.567 & 0.840 & \cellcolor[HTML]{D6ECE8}\textbf{0.886} & 0.857 & 0.193 & 0.675 & 0.861 & \cellcolor[HTML]{F6E6C9} 0.697 \\
BLEU & 0.188 & 0.363 & 0.406 & \cellcolor[HTML]{D6ECE8}\textbf{0.459} & 0.049 & 0.290 & 0.406 & \cellcolor[HTML]{F6E6C9} 0.309 \\
TEDS & 0.584 & 0.812 & 0.830 & \cellcolor[HTML]{D6ECE8}\textbf{0.877} & 0.153 & 0.605 & 0.824 & \cellcolor[HTML]{F6E6C9} 0.669 \\
\midrule
\multicolumn{9}{l}{\textbf{Text Addition} (976 samples)} \\
\midrule
IoU & 0.536 & 0.586 & \cellcolor[HTML]{D6ECE8}\textbf{0.603} & 0.598 & 0.440 & 0.429 & 0.553 & \cellcolor[HTML]{F6E6C9} 0.535 \\
CDM & 0.519 & 0.812 & 0.864 & \cellcolor[HTML]{D6ECE8}\textbf{0.864} & 0.167 & 0.674 & 0.794 & \cellcolor[HTML]{F6E6C9} 0.671 \\
BLEU & 0.193 & 0.558 & 0.614 & \cellcolor[HTML]{D6ECE8}\textbf{0.647} & 0.048 & 0.386 & 0.531 & \cellcolor[HTML]{F6E6C9} 0.425 \\
TEDS & 0.549 & 0.845 & 0.853 & \cellcolor[HTML]{D6ECE8}\textbf{0.903} & 0.149 & 0.696 & 0.791 & \cellcolor[HTML]{F6E6C9} 0.684 \\
\midrule
\multicolumn{9}{l}{\textbf{Text Deletion} (971 samples)} \\
\midrule
IoU & 0.688 & 0.749 & 0.714 & \cellcolor[HTML]{D6ECE8}\textbf{0.750} & 0.651 & 0.348 & 0.720 & \cellcolor[HTML]{F6E6C9} 0.660 \\
CDM & 0.515 & 0.815 & 0.820 & \cellcolor[HTML]{D6ECE8}\textbf{0.837} & 0.162 & 0.648 & 0.766 & \cellcolor[HTML]{F6E6C9} 0.652 \\
BLEU & 0.256 & 0.562 & 0.592 & \cellcolor[HTML]{D6ECE8}\textbf{0.651} & 0.045 & 0.381 & 0.531 & \cellcolor[HTML]{F6E6C9} 0.431 \\
TEDS & 0.553 & 0.825 & 0.798 & \cellcolor[HTML]{D6ECE8}\textbf{0.845} & 0.164 & 0.617 & 0.800 & \cellcolor[HTML]{F6E6C9} 0.657 \\
\midrule
\multicolumn{9}{l}{\textbf{Table Structure Edit} (460 samples)} \\
\midrule
IoU & 0.862 & 0.890 & 0.639 & \cellcolor[HTML]{D6ECE8}\textbf{0.907} & 0.771 & 0.515 & 0.819 & \cellcolor[HTML]{F6E6C9} 0.772 \\
CDM & 0.385 & 0.669 & 0.497 & \cellcolor[HTML]{D6ECE8}\textbf{0.768} & 0.170 & 0.491 & 0.686 & \cellcolor[HTML]{F6E6C9} 0.524 \\
BLEU & 0.126 & 0.349 & 0.214 & \cellcolor[HTML]{D6ECE8}\textbf{0.612} & 0.026 & 0.239 & 0.429 & \cellcolor[HTML]{F6E6C9} 0.285 \\
TEDS & 0.366 & 0.661 & 0.467 & \cellcolor[HTML]{D6ECE8}\textbf{0.750} & 0.157 & 0.425 & 0.670 & \cellcolor[HTML]{F6E6C9} 0.499 \\
\bottomrule
\end{tabularx}
% 中文：3,506 条样本上按编辑类型划分的表现。
% 所有分数均采用局部图像设置。
% Text Modify 合并替换与修改；添加和删除均包含文本区域与表格区域。
% Avg. 为七种编辑模型的平均值。
% 青绿色及加粗标记每行最佳编辑模型。
\caption{\textbf{Performance by edit type on 3,506 samples.} All scores use the local-image setting. Text Modify combines replacement and modification; addition and deletion include text and table regions. Avg. is the mean over the seven editing models. Teal and bold mark the best editing model per row.}
\label{tab:edit_type_local}
\end{table*}

% 中文标题：相关工作评述
\section{Related Works Assessment}

% 中文标题：文档理解与评测的研究基础
\subsection{Document Understanding and Evaluation Foundations}
\label{sec:expanded foundations}

% 中文：布局感知表示。
% 文档图像中的信息同时编码在文本内容与空间组织中。
% LayoutLM 结合文本与二维布局，LayoutLMv2 融合视觉特征及跨模态预训练，LayoutLMv3 则采用统一的文本与图像掩码。
% DocFormer 同样学习用于文档理解的多模态表示。
% 这些方法说明，评估文档编辑时必须考虑其空间上下文：仅保留可识别的词语，并不能证明它们与相邻内容的关系也得到了保持。
% VDE Bench 评估最终输出图像，而非这些理解模型的内部表示。
\paragraph{Layout-aware representations.}
Document images encode information in both their textual content and spatial organization. LayoutLM combines text and two-dimensional layout, LayoutLMv2 integrates visual features and cross-modal pretraining, and LayoutLMv3 uses unified text and image masking \citep{xu2020layoutlm,xu2021layoutlmv2,huang2022layoutlmv3}. DocFormer likewise learns multimodal representations for document understanding \citep{appalaraju2021docformer}. These approaches establish why a document edit must be assessed in its spatial context: retaining recognizable words does not by itself establish that their relationships to neighboring content are preserved. VDE Bench evaluates resulting images rather than the internal representations of these understanding models.

% 中文：无需 OCR 的解析与文档生成。
% Donut 将文档理解建模为无需外部 OCR 引擎的图像到序列预测；Pix2Struct 从截图到 HTML 的解析任务中学习，Nougat 则面向学术文档转写。
% UDOP 统一视觉、文本与布局，并明确展示了文档生成和编辑能力。
% 因此，文档编辑任务早于 VDE Bench 已经存在。
% 我们的资源聚焦于利用配对参考结果、局部标注和文档上下文保持，对指令引导的修改进行评估。
% 生成编辑后页面的能力，与可靠衡量其修改正确性的能力，是两项不同要求。
\paragraph{OCR-free parsing and document generation.}
Donut frames document understanding as image-to-sequence prediction without an external OCR engine, while Pix2Struct learns from screenshot-to-HTML parsing and Nougat targets academic document transcription \citep{kim2022donut,lee2023pix2struct,blecher2023nougat}. UDOP unifies vision, text, and layout, and explicitly demonstrates document generation and editing \citep{tang2023udop}. Thus, document editing predates VDE Bench. Our resource instead focuses on evaluating instruction-guided changes using paired references, localized annotations, and document-context preservation. The ability to produce an edited page and the ability to reliably measure the correctness of its changes are distinct requirements.

% 中文：面向文档的多模态模型。
% mPLUG-DocOwl 将多模态语言模型适配到无需 OCR 的文档理解；DocOwl 1.5 强调结构学习与高分辨率文档感知。
% GOT 在统一 OCR 模型中支持格式化识别和区域级识别。
% 这些系统为结构化文档处理提供研究背景，但在此引用并不意味着我们将其作为编辑基线进行了评测。
% 实际实现的标注流程使用 PaddleOCR-VL，并对编辑区域内的预测结果进行人工检查。
\paragraph{Document-oriented multimodal models.}
mPLUG-DocOwl adapts multimodal language models to OCR-free document understanding; DocOwl 1.5 emphasizes structure learning and high-resolution document perception \citep{ye2023docowl,hu2024docowl15}. GOT supports formatted and region-level recognition in a unified OCR model \citep{wei2024got}. These systems provide context for structured document processing, but their inclusion here does not imply that they were evaluated as editing baselines. Our implemented annotation pipeline uses PaddleOCR-VL, whose predictions in edited regions are manually checked \citep{cui2025paddleocrvlboostingmultilingualdocument}.

% 中文：页面布局资源。
% PubLayNet 从科学出版物中获取大规模布局标注，DocBank 提供词元级文档布局标注，DocLayNet 则强调针对多样文档布局开展人工标注。
% OmniDocBench 在多种页面及结构元素上评测文档解析。
% 这些资源支持对异构页面结构进行评测的思路，但仅有布局识别标注，并不能指定编辑指令或正确的编辑后参考结果。
% VDE Bench 的种子文档来自 OmniDocBench、DocLayNet、PMC，以及通过 DocGenome 获取的 arXiv 论文；PubLayNet 和 DocBank 在此作为相关工作讨论。
\paragraph{Page-layout resources.}
PubLayNet obtains large-scale layout annotations from scientific publications, DocBank provides token-level document-layout annotations, and DocLayNet emphasizes human annotations across varied document layouts \citep{zhong2019publaynet,li2020docbank,pfitzmann2022doclaynet}. OmniDocBench evaluates document parsing across diverse pages and structural elements \citep{ouyang2024omnidocbenchbenchmarkingdiversepdf}. Such resources motivate evaluating heterogeneous page structures, although layout recognition annotations alone do not specify an editing instruction or a correct edited reference. VDE Bench draws seed documents from OmniDocBench, DocLayNet, PMC, and arXiv papers accessed through DocGenome. PubLayNet and DocBank are discussed here as related work.

% 中文：表单与多语言文档。
% FUNSD 研究有噪声扫描文档中的表单理解。
% LayoutXLM 提出 XFUND，提供七种语言的人工标注表单信息。
% 这些数据集表明，语言覆盖与文档结构是评测中彼此独立的维度。
% 我们的中文和英文子集为考察编辑过程中两者的交互提供了有限场景，但并不代表已覆盖其他文字系统，也不意味着不同语言的分布等价。
\paragraph{Forms and multilingual documents.}
FUNSD studies form understanding in noisy scanned documents. LayoutXLM introduces XFUND, with manually labeled form information in seven languages \citep{jaume2019funsd,xu2021layoutxlm}. These datasets illustrate that language coverage and document structure are separate dimensions of evaluation. Our Chinese and English subsets provide a limited setting for examining their interaction during editing; they do not establish coverage of other scripts or equivalence between language distributions.

% 中文：阅读与推理基准。
% TextVQA 和 ST-VQA 测试系统能否读取图像中的文本，并据此回答问题。
% DocVQA 聚焦文档图像，InfographicVQA 和 ChartQA 则进一步拓展文本内容、视觉结构与推理之间的关系。
% OCRBench 使用一组视觉文本任务评估多模态模型。
% 这些基准与文档编辑中的识别环节相关。
% 然而，回答正确并不要求重建整张页面，而编辑任务还必须生成指定修改，并保持周围内容。
\paragraph{Reading and reasoning benchmarks.}
TextVQA and ST-VQA test whether systems can read text in images and use it to answer questions \citep{singh2019textvqa,biten2019stvqa}. DocVQA focuses on document images, while InfographicVQA and ChartQA broaden the relationship between textual content, visual structure, and reasoning \citep{mathew2021docvqa,mathew2022infographicvqa,masry2022chartqa}. OCRBench assesses multimodal models on a collection of visual-text tasks \citep{liu2023ocrbench}. These benchmarks are relevant to the recognition component of document editing. However, an answer can be correct without reconstructing the entire page, whereas editing additionally requires producing the requested change and preserving surrounding content.

% 中文：表格结构与评测。
% TableBank 支持表格检测与识别，SciTSR 研究复杂单元格关系，PubTables-1M 提供完整的表格提取标注。
% TableFormer 使用 Transformer 预测表格结构，LGPMA 则处理局部与全局表格表示的对齐。
% PubTabNet 为基于图像的表格识别引入基于树编辑距离的评测。
% 这些研究说明，除了识别字符，还应测量结构错误。
% 本文的 TEDS-like 指标使用相应附录中规定的表示，不应将其数值视为可与原始 PubTabNet TEDS 实现直接互换。
\paragraph{Table structure and evaluation.}
TableBank supports table detection and recognition, SciTSR studies complicated cell relationships, and PubTables-1M provides comprehensive table-extraction annotations \citep{li2020tablebank,chi2019scitsr,smock2022pubtables}. TableFormer predicts table structure with transformers, while LGPMA addresses alignment of local and global table representations \citep{nassar2022tableformer,qiao2021lgpma}. PubTabNet introduced tree-edit-distance-based evaluation for image-based table recognition \citep{zhong2020imagebasedtablerecognitiondata}. These studies motivate measuring structural errors in addition to recognized characters. Our TEDS-like measure uses the representation specified in Appendix~\ref{subsubsec:TEDS-like Similarity}; it should not be interpreted as numerically interchangeable with the original PubTabNet TEDS implementation.

% 中文：局部场景文本编辑。
% SRNet 将文本转换、背景修复和融合分离处理；TextStyleBrush 从单个示例迁移文本外观。
% MOSTEL 使用笔画级引导，DiffSTE 结合字符编码器与指令编码器，以控制可读性和风格。
% 这些方法揭示，应将所请求的文本修改与背景、风格保持区分开来。
% VDE Bench 将这一要求放入密集页面和表格布局中考察，而非把每个文本区域都当作孤立的词语图像。
\paragraph{Localized scene-text editing.}
SRNet separates text conversion, background inpainting, and fusion; TextStyleBrush transfers text appearance from a single example \citep{wu2019srnet,krishnan2021textstylebrush}. MOSTEL uses stroke-level guidance, and DiffSTE combines character and instruction encoders for legibility and style control \citep{qu2023mostel,ji2023diffste}. These methods expose the need to separate requested textual changes from preservation of background and style. VDE Bench places that requirement within dense pages and tabular layouts, rather than treating every text region as an isolated word image.

% 中文：视觉文本渲染。
% TextDiffuser 与 TextDiffuser-2 将扩散式渲染与明确的文本放置、生成机制相结合。
% Glyph-ByT5 开发了与字形对齐的字符感知文本编码器，其后续版本进一步扩展多语言渲染。
% AnyText2 增加对字体和颜色等文本属性的控制。
% 渲染与编辑相关但不同：视觉上合理的新页面未必满足以源页面为条件的编辑指令。
% 通用潜空间扩散与空间条件控制，为这些系统提供了方法背景。
\paragraph{Visual text rendering.}
TextDiffuser and TextDiffuser-2 combine diffusion-based rendering with explicit mechanisms for text placement and generation \citep{chen2023textdiffuser,chen2024textdiffuser2}. Glyph-ByT5 develops character-aware text encoders aligned with glyphs, and its successor extends multilingual rendering \citep{liu2024glyphbyt5,liu2024glyphbyt5v2}. AnyText2 adds control over text attributes such as font and color \citep{tuo2025anytext2}. Rendering and editing are related but distinct: a visually plausible new page need not satisfy a source-conditioned editing instruction. General latent diffusion and spatial conditioning provide methodological context for these systems \citep{rombach2022highresolutionimagesynthesislatent,zhang2023addingconditionalcontroltexttoimage}.

% 中文：互补的自动测量。
% SSIM 评估结构图像相似性，LPIPS 使用深度特征测量感知差异，CLIP 学习共享的视觉—语言表示。
% CLIPScore 将跨模态兼容性用于图像描述评测；因此，必须明确说明基于 CLIP 的分数所采用的输入配对方式与归一化方法，而不能仅根据模型名称推断。
% 同样，chrF 和 BERTScore 分别提供基于字符与上下文的文本比较方法，但本文表格并未将其作为额外实测结果报告。
% 精确文本修改、区域对齐和图像保持应分别报告，因为某一项分数较好，并不足以说明三个方面均已成功。
\paragraph{Complementary automatic measurements.}
SSIM assesses structural image similarity, LPIPS measures perceptual differences using deep features, and CLIP learns shared visual-language representations \citep{wang2004image,zhang2018unreasonable,radford2021learning}. CLIPScore applies cross-modal compatibility to caption evaluation \citep{hessel2021clipscore}; the input pairing and normalization of a CLIP-based score must therefore be specified rather than assumed from the model name. Likewise, chrF and BERTScore offer character-based and contextual alternatives for comparing text \citep{popovic2015chrf,zhang2020bertscore}, but are not additional measured results in our tables. Exact textual changes, region alignment, and image preservation warrant separate reporting because one favorable score need not establish success on all three.

% 中文：可复现性与标注质量。
% BLEU 的可复现性取决于是否报告预处理和评分约定。
% 一致性度量衡量标注人员之间的一致程度，不应与任务准确率或样本代表性混淆。
% 数据集说明应明确交代收集过程、样本组成、推荐用途和局限，相关原则见数据集说明书与数据声明研究。
% 模型卡同样倡导记录系统特征与评测条件。
% 这些原则支持透明报告源语料、筛选过程、语言、参考生成器和评测范围，但不能证明筛选已消除依赖生成器的选择效应。
\paragraph{Reproducibility and annotation quality.}
The reproducibility of BLEU depends on reporting preprocessing and scoring conventions \citep{post2018sacrebleu}. Agreement measures address consistency among annotators, and should not be confused with task accuracy or representativeness \citep{artstein2008agreement}. Dataset documentation should make collection, composition, recommended uses, and limitations explicit, as discussed in datasheets and data statements \citep{gebru2021datasheets,bender2018datastatements}. Model cards similarly motivate documenting system characteristics and evaluation conditions \citep{mitchell2019modelcards}. These principles support transparent reporting of the source corpus, filtering procedure, languages, reference generator, and evaluation scope; they do not establish that filtering eliminates generator-dependent selection effects.

% 中文标题：图像生成与可控编辑
\subsection{Image Generation and Controllable Editing}

% 中文：扩散与流模型。
% DDPM 通过逐步去噪学习图像生成，DDIM 提供非马尔可夫采样过程，基于分数的随机微分方程框架则将扩散过程与连续时间生成联系起来。
% 无分类器引导通过结合有条件与无条件预测调节条件约束的强度。
% 潜空间扩散降低了高分辨率图像生成的计算开销，扩散 Transformer 与流匹配进一步拓展了生成骨干网络和训练目标。
% 这些工作构成现代编辑模型的生成基础；源图像约束决定了这些生成能力如何用于局部修改。
\paragraph{Diffusion and flow models.}
DDPM learns image generation through iterative denoising, DDIM provides a non-Markovian sampling process, and score-based stochastic differential equations connect diffusion to continuous-time generation \citep{ho2020ddpm,song2020ddim,song2021sde}. Classifier-free guidance controls conditioning strength by combining conditional and unconditional predictions \citep{ho2022cfg}. Latent diffusion reduces the cost of high-resolution synthesis, while diffusion transformers and flow matching broaden generative backbones and training objectives \citep{rombach2022highresolutionimagesynthesislatent,peebles2023dit,lipman2023flowmatching}. These methods underpin modern editors; source-image conditioning determines how their generative capacity is applied to localized changes.

% 中文：文本条件与图像生成。
% DALL-E 2 利用 CLIP 图像潜变量进行分层生成，Imagen 将大型语言编码器与级联扩散模型结合，SDXL 则扩展潜空间扩散系统以提升高分辨率生成能力。
% 这些研究为基于文字的视觉控制提供了技术基础。
% 在文档编辑中，文字条件还必须落实为页面上的准确字符与位置，并与保留区域协调一致。
\paragraph{Text conditioning and image synthesis.}
DALL-E 2 uses CLIP image latents in hierarchical generation, Imagen combines large language encoders with cascaded diffusion, and SDXL scales latent diffusion for high-resolution synthesis \citep{ramesh2022dalle2,saharia2022imagen,podell2023sdxl}. These systems establish a technical basis for textual control over visual content. Document editing further requires that such control produce the intended characters at the intended locations while remaining consistent with preserved regions.

% 中文：源图像重建与反演。
% SDEdit 先向输入图像加入噪声，再通过去噪实现引导合成。
% Null-text Inversion 通过优化无条件嵌入，使真实图像适用于基于扩散的编辑；Direct Inversion 则在反演与编辑过程中补偿重建偏差。
% 这一研究方向关注可编辑性与源图像保真度之间的关系，与文档编辑中同时检查目标区域和周围页面的要求密切相关。
\paragraph{Source reconstruction and inversion.}
SDEdit adds noise to an input image and then denoises it for guided synthesis \citep{meng2022sdedit}. Null-text Inversion optimizes unconditional embeddings to enable diffusion-based editing of real images, while Direct Inversion compensates for reconstruction deviations during inversion and editing \citep{mokady2023nulltext,ju2023directinversion}. This line of work examines the relationship between editability and source fidelity, motivating the joint inspection of edited regions and surrounding page content.

% 中文：注意力与特征控制。
% Prompt-to-Prompt 通过控制交叉注意力引导编辑；MasaCtrl 使用互相自注意力，Plug-and-Play Diffusion Features 则将源图像的特征与注意力信息注入生成过程。
% pix2pix-zero 利用文本方向和交叉注意力引导实现无需额外训练的图像转换。
% 这些方法将编辑语义与空间结构联系起来，是理解局部改动和内容保持机制的重要参照。
\paragraph{Attention and feature control.}
Prompt-to-Prompt controls cross-attention, MasaCtrl uses mutual self-attention, and Plug-and-Play Diffusion Features injects source features and attention into generation \citep{hertz2022prompttoprompt,cao2023masactrl,tumanyan2023plugandplay}. The pix2pix-zero method uses textual directions and cross-attention guidance for training-free image translation \citep{parmar2023pix2pixzero}. These approaches connect editing semantics to spatial structure and provide complementary mechanisms for localized modification and content preservation.

% 中文：掩码与参考图像引导。
% Blended Diffusion 在指定区域执行文本引导编辑；RePaint 使用预训练扩散模型进行图像修复。
% DiffEdit 根据文本条件自动推断编辑掩码，Paint by Example 则使用参考图像指定修复内容。
% 它们分别展示了显式空间约束、自动定位和视觉参考条件的作用，也说明评测协议需要区分“提供给模型的掩码”与“用于评分的区域标注”。
\paragraph{Masks and visual references.}
Blended Diffusion performs text-guided editing within specified regions, and RePaint uses a pretrained diffusion model for inpainting \citep{avrahami2022blended,lugmayr2022repaint}. DiffEdit infers editing masks from textual conditions, whereas Paint by Example uses an image reference to specify inpainting content \citep{couairon2023diffedit,yang2023paintbyexample}. These methods illustrate explicit spatial constraints, automatic localization, and visual conditioning, respectively. Evaluation protocols should therefore distinguish masks supplied to an editor from region annotations used for scoring.

% 中文：指令理解与偏好对齐。
% InstructDiffusion 以指令驱动的扩散框架统一多种视觉任务；MGIE 和 SmartEdit 使用多模态语言模型解释更复杂的编辑意图。
% HIVE 将人类反馈用于指令编辑对齐，ICEdit 研究上下文式图像编辑。
% 从扩散条件到多模态推理的演进，使同一条自然语言请求可以通过不同机制落实到图像修改。
% VDE Bench 使用统一的输出评测，比较这些机制在密集文档内容上的实际表现。
\paragraph{Instruction understanding and alignment.}
InstructDiffusion unifies visual tasks within an instruction-driven diffusion framework; MGIE and SmartEdit use multimodal language models to interpret more complex editing intentions \citep{geng2024instructdiffusion,fu2024mgie,huang2024smartedit}. HIVE incorporates human feedback into instruction-editing alignment, and ICEdit studies in-context image editing \citep{zhang2024hive,zhang2025icedit}. The progression from diffusion conditioning to multimodal reasoning offers different mechanisms for translating a natural-language request into an image modification. VDE Bench assesses resulting outputs with a common protocol on dense document content.

% 中文：统一生成与文本密集场景。
% HiDream-O1-Image 在统一 Transformer 中处理像素、文本与任务条件；JoyAI-Image 将空间增强的多模态语言模型与扩散 Transformer 相连接。
% 针对图像中的文字，GlyphDraw 研究字形条件下的文本绘制，TextAtlas5M 提供大规模密集文本图像资源及相应评测。
% 这些工作将统一生成能力与文字渲染要求联系起来，为文档页面上的精确编辑提供了直接研究背景。
\paragraph{Unified generation and text-rich images.}
HiDream-O1-Image processes pixels, text, and task conditions within a unified transformer, while JoyAI-Image connects a spatially enhanced multimodal language model with a diffusion transformer \citep{cai2026hidream,song2026joy}. For text within images, GlyphDraw studies glyph-conditioned rendering, and TextAtlas5M provides large-scale dense-text image resources and evaluation \citep{ma2023glyphdraw,wang2025textatlas5mlargescaledatasetdense}. These efforts connect unified generation to text rendering and provide direct context for precise editing of document pages.

% 中文标题：结构化阅读与细粒度评测
\subsection{Structured Reading and Fine-grained Evaluation}

% 中文：文本识别与空间关系。
% TrOCR 以预训练图像 Transformer 和文本 Transformer 完成文字识别，PaddleOCR 3.0 将文本识别与文档解析扩展为更完整的工具体系。
% StrucTexT 和 BROS 建模文本与布局关系，LiLT 研究与语言无关的布局表示，LayoutReader 通过文本和布局学习阅读顺序。
% DocLLM 进一步将边界框信息融入文档语言建模。
% 这些研究说明，页面编辑的错误既可能出现在字符内容，也可能出现在区域之间的组织关系。
\paragraph{Recognition and spatial relationships.}
TrOCR combines pretrained image and text transformers for recognition, and PaddleOCR 3.0 expands text recognition and document parsing into a broader toolkit \citep{li2021trocr,cui2025paddleocr30technicalreport}. StrucTexT and BROS model textual and spatial relationships, LiLT learns language-independent layout representations, and LayoutReader uses text and layout to learn reading order \citep{li2021structext,hong2021bros,wang2022lilt,wang2021layoutreader}. DocLLM further incorporates bounding-box information into document language modeling \citep{wang2024docllm}. These studies show why editing errors can concern both character content and the organization of regions.

% 中文：端到端文档理解与 OCR 评测。
% DUE 将视觉问答、关键信息抽取和机器阅读理解纳入统一的端到端文档评测。
% OCRBench v2 扩展了对多模态模型文本识别、定位、解析和推理能力的评测。
% 两者强调了从局部识别到整页理解的多层次能力；文档编辑还需在这些能力的基础上生成修改后的页面。
\paragraph{End-to-end understanding and OCR evaluation.}
DUE unifies visual question answering, key-information extraction, and machine reading comprehension in an end-to-end document benchmark \citep{borchmann2021due}. OCRBench v2 broadens the assessment of multimodal models across recognition, localization, parsing, and reasoning \citep{fu2025ocrbenchv2improvedbenchmark}. Together, they motivate evaluation across local recognition and page-level understanding; document editing additionally requires generating the modified page.

% 中文：表格与组合关系的评分。
% GriTS 在网格结构上评估表格相似性，补充了基于树结构的表格评价方式。
% 在文本到图像生成中，T2I-CompBench 考察属性绑定、物体关系与复杂组合，GenEval 使用对象层面的分析，TIFA 则通过视觉问答检查生成结果是否满足文本要求。
% 这些研究共同支持将复杂任务拆分为可核查的要求；文档编辑中的对应要求包括字符正确性、目标定位和结构保持。
\paragraph{Tables and compositional relationships.}
GriTS evaluates table similarity over grid structure, complementing tree-based table measures \citep{smock2022grits}. In text-to-image generation, T2I-CompBench examines attribute binding, object relationships, and complex composition; GenEval uses object-focused analysis; and TIFA checks textual requirements through visual question answering \citep{huang2023t2icompbench,ghosh2023geneval,hu2023tifa}. These approaches support decomposing a complex task into verifiable requirements. For document editing, the corresponding requirements include character correctness, target localization, and structural preservation.

% 中文：感知质量与人类偏好。
% FID 比较生成图像与真实图像的特征分布，DISTS 在成对图像比较中结合结构相似性与纹理相似性。
% ImageReward 从人类偏好学习生成图像奖励，Pick-a-Pic 提供用户对文本到图像生成结果的成对偏好。
% 分布质量、感知相似性和用户偏好各自衡量不同的输出属性；文档编辑评测还需直接核查所请求的文字和结构修改。
\paragraph{Perceptual quality and human preferences.}
FID compares feature distributions of generated and real images, whereas DISTS combines structure and texture similarity in paired image comparisons \citep{heusel2017fid,ding2020dists}. ImageReward learns a reward from human preferences, and Pick-a-Pic collects pairwise user preferences for text-to-image outputs \citep{xu2023imagereward,kirstain2023pickascore}. Distributional quality, perceptual similarity, and user preference measure different output properties; document editing evaluation also requires direct checks of the requested textual and structural modifications.

% 中文：编辑区域的定位与分割。
% Grounding DINO 根据类别名称或指代表达定位目标，Segment Anything 则支持提示驱动的区域分割。
% 文本区域具有更细致的字符结构：CRAFT 通过字符区域及字符间关联进行检测，DBNet 将可微分二值化融入文本检测流程。
% 这些方法提供从语义目标到精细区域的不同定位机制，有助于区分目标选错与目标内部内容生成错误。
\paragraph{Grounding and segmentation.}
Grounding DINO localizes objects from category names or referring expressions, while Segment Anything provides promptable region segmentation \citep{liu2023groundingdino,kirillov2023sam}. Text regions require finer character structure: CRAFT detects character regions and their affinities, and DBNet incorporates differentiable binarization into text detection \citep{baek2019craft,liao2020dbnet}. These methods provide complementary localization mechanisms, motivating a distinction between selecting the wrong target and generating incorrect content within the correct target.

% 中文：视觉特征与字符转写。
% DINO 和 DINOv2 通过自监督学习获取可迁移的视觉特征，为外观与区域结构比较提供表征基础。
% 字符转写具有不同的目标：PARSeq 使用排列式语言建模学习自回归识别模型，SVTR 则以单一视觉模型识别场景文字。
% 因而，表征空间的视觉接近与文字序列的精确匹配应分别测量。
\paragraph{Visual features and character transcription.}
DINO and DINOv2 learn transferable visual features through self-supervision, providing representations for comparing appearance and regional structure \citep{caron2021dino,oquab2023dinov2}. Character transcription has a different objective: PARSeq learns autoregressive recognition through permutation language modeling, whereas SVTR recognizes scene text with a single visual model \citep{bautista2022parseq,du2022svtr}. Visual proximity in feature space and exact agreement of text sequences therefore warrant separate measurements.

% 中文：文档类型、业务字段与多页上下文。
% RVL-CDIP 支持文档图像分类与检索研究，DocILE 聚焦业务文档中的关键信息定位、抽取及行项目识别。
% MP-DocVQA 将文档视觉问答扩展到多页文档，DUDE 进一步涵盖多个行业、领域与页面的文档理解。
% 这些资源从文档类型、字段结构和跨页上下文等方面拓展了文档研究的范围，与本文聚焦单页局部修改的评测形成互补。
\paragraph{Document types, business fields, and multi-page context.}
RVL-CDIP supports document-image classification and retrieval, while DocILE targets key-information localization, extraction, and line-item recognition in business documents \citep{harley2015rvlcdip,simsa2023docile}. MP-DocVQA extends visual question answering to multi-page documents, and DUDE broadens evaluation across industries, domains, and pages \citep{tito2023multipage,vanlandeghem2023dude}. These resources extend document research along document type, field structure, and cross-page context, complementing our focus on localized changes within a page.

% 中文：图像参考与视觉条件。
% IP-Adapter 通过图像提示适配器引入与文本兼容的视觉条件。
% DiffusionCLIP 将 CLIP 引导与扩散模型相结合，实现基于文字的图像操作。
% 这些方法展示了参考外观、文本条件与源图像信息的不同结合方式，为考察编辑中的内容保持提供方法层面的背景。
\paragraph{Image references and visual conditioning.}
IP-Adapter introduces text-compatible visual conditioning through an image-prompt adapter \citep{ye2023ipadapter}. DiffusionCLIP combines CLIP guidance with diffusion for text-driven image manipulation \citep{kim2022diffusionclip}. These methods illustrate different ways of combining reference appearance, textual conditions, and source-image information, providing methodological context for content preservation during editing.

% 中文：编辑数据的规模与任务覆盖。
% UltraEdit 通过自动化流程生成大规模细粒度编辑数据，SEED-Data-Edit 将自动生成、人工标注和视频来源的样本结合为混合数据集。
% EditWorld 根据不同现实场景定义世界指令驱动的图像编辑，并构建相应数据与编辑方法。
% 这些工作说明，数据构建策略与任务定义共同决定编辑模型能够学习的修改范围；VDE Bench 将这一问题聚焦到密集文档中的文字和表格操作。
\paragraph{Editing-data scale and task coverage.}
UltraEdit generates large-scale fine-grained editing data through an automated pipeline, and SEED-Data-Edit combines automatically generated, manually annotated, and video-derived samples in a hybrid dataset \citep{zhao2024ultraedit,ge2024seeddata}. EditWorld defines world-instructed editing across real-world scenarios and develops corresponding data and editing methods \citep{yang2024editworld}. These studies show how construction strategy and task definition jointly determine the modifications available for learning; VDE Bench focuses this question on textual and tabular operations in dense documents.

% 中文标题：各基准的逐项评述
\subsection{Benchmark-by-Benchmark Assessment}

% 中文：基准对比表从三个方面概括既有基准数据集：（1）采用单轮还是多轮交互；（2）是否包含文本编辑；（3）是否涉及人工核验。
% 下面简要说明各项分类的依据。
Table ~\ref{tab:compare} summarizes prior benchmark datasets from three aspects: (1) whether they involve single-turn or multi-turn dialogue; (2) whether they include text editing; and (3) whether human verification is involved. We briefly justify our assignments here.

\paragraph{I2EBench.}

% 中文：I2EBench 是一个主要采用单轮设置的指令驱动图像编辑基准。
% 每项任务通常由一条编辑指令与一张图像组成，不涉及多轮对话。
% 其核心任务是视觉内容编辑，而非文本修改。
% 此外，数据集在构建和／或评测过程中引入一定程度的人工标注或人工核验，以确保指令与编辑结果的一致性，以及评测的可靠性。
I2EBench is an instruction-driven image editing benchmark that is primarily single-turn. Each task typically consists of a single editing instruction paired with a single image, without involving multi-turn dialogue. The core task focuses on visual content editing rather than text modification. In addition, the dataset incorporates a certain degree of human annotation or manual verification during its construction and/or evaluation process to ensure consistency between instructions and edited results, as well as the reliability of the evaluation.

\paragraph{EditBench.}

% 中文：EditBench 随 Imagen Editor 提出，用于评测文本引导的图像修复。每个样本包含掩码图像、文本提示和参考图像，评测覆盖物体、属性与场景，其中也包含文字渲染，并使用人工评测检查生成结果。
EditBench~\cite{wang2023imagen}, introduced with Imagen Editor, evaluates text-guided inpainting using masked images, prompts, and reference images. Its evaluation covers objects, attributes, and scenes, including text rendering, and uses human judgments to assess outputs.

\paragraph{EditVal.}

% 中文：EditVal 用于评估基于扩散模型的文本引导图像编辑方法。
% 它定义了多种编辑操作，例如添加、移除或替换物体，以及修改属性等，每项操作均为独立的单轮任务。
% 基准经过人工核验以确保评测可靠性。
% 它不涉及图像内文本编辑，也不提供编辑区域掩码。
EditVal~\cite{basu2023editval} is a benchmark designed to evaluate diffusion-based text-guided image editing methods. It defines a diverse set of editing operation types (e.g., object addition, removal, replacement, attribute modification, etc.), where each editing operation is an independent single-turn task. The benchmark is human-verified to ensure evaluation reliability. It does not involve text editing within images and does not provide editing region masks.

\paragraph{EmuEdit.}

% 中文：EmuEdit 是 Meta 提出的基于指令的精确图像编辑模型及其配套基准。
% 该基准覆盖多种编辑任务，包括区域编辑、自由形式编辑和风格迁移等，各项任务均采用单轮设置。
% 评测数据经过人工核验，以保证指令与编辑结果一致。
% 它不涉及图像内文本编辑，也不提供或要求编辑区域掩码。
EmuEdit~\cite{sheynin2024emu} is an instruction-based precise image editing model and its accompanying benchmark proposed by Meta. The benchmark covers multiple editing task types (including region-based editing, free-form editing, style transfer, etc.), each operating in a single-turn setting. The evaluation data is human-verified to ensure consistency between instructions and editing results. It does not involve text editing within images and does not provide or require editing region masks.

\paragraph{AnyEdit.}

% 中文：AnyEdit 通过自动化流程构建大规模、多类型图像编辑数据。其评测集 AnyEdit-Test 则经过人工筛选，包含覆盖 25 类任务的 1,250 个编辑样本对。训练数据的自动生成与评测集的人工筛选是两个不同环节。
AnyEdit~\cite{Yu_2025_CVPR} constructs large-scale, multi-type editing data through an automated pipeline. Its evaluation set, AnyEdit-Test, is manually curated and contains 1,250 editing pairs across 25 categories. Automated training-data generation and manual test-set curation are distinct stages.

\paragraph{CompBench.}

% 中文：CompBench 使用多模态大模型与人工协作的流程，构建复杂指令驱动的图像编辑基准。其任务包括多物体编辑、空间与隐式推理，以及多轮编辑，可考察模型对细粒度复合要求的执行能力。
CompBench~\cite{jia2025compbench} uses multimodal-model--human collaboration to construct a benchmark for complex instruction-guided editing. Its tasks include multi-object editing, spatial and implicit reasoning, and multi-turn editing, testing execution of fine-grained compound requirements.

\paragraph{Omni IIE Bench.}

% 中文：Omni IIE Bench 分别设置单轮一致性与多轮协调性两条评测轨道。数据流程生成编辑图像与空间掩码，再进行人工筛选。它考察模型在不同语义尺度的编辑任务之间能否保持稳定表现。
Omni IIE Bench~\cite{yang2026omniiiebench} separates single-turn consistency from multi-turn coordination. Its construction pipeline generates edited images and spatial masks, followed by manual curation. It examines whether models perform consistently across edits at different semantic scales.

\paragraph{MagicBrush.}

% 中文：MagicBrush 提供人工标注的源图像、编辑指令和目标图像三元组，同时支持单轮、多轮、有掩码和无掩码编辑设置。其构建流程由标注人员利用图像编辑工具创建并核查编辑结果。
MagicBrush~\cite{zhang2023magicbrush} provides manually annotated source-image, instruction, and target-image triplets. It supports single-turn, multi-turn, mask-provided, and mask-free settings, with annotators creating and checking edits through an image-editing tool.

\paragraph{ImgEdit-Bench.}

% 中文：ImgEdit 将大规模图像编辑数据与分层评测相结合。ImgEdit-Bench 包含基础测试、具有挑战性的单轮测试和专门的多轮测试，考察复杂指令执行及连续编辑中的内容保持。
ImgEdit~\cite{ye2025imgedit} combines large-scale editing data with layered evaluation. ImgEdit-Bench includes a basic test suite, a challenging single-turn suite, and a dedicated multi-turn suite, examining complex instruction following and content retention across successive edits.

\paragraph{MuCIE.}

% 中文：MuCIE 研究多轮连续编辑中的一致性。该方法结合基于流匹配的反演、稳定采样与自适应注意力控制，减少迭代编辑中的误差累积；相应实验评估连续修改后的编辑效果与视觉保真度。
MuCIE~\cite{zhou2025multi} studies consistency in successive editing rounds. Its framework combines flow-based inversion, stable sampling, and adaptive attention control to reduce error accumulation, with experiments assessing editing success and visual fidelity over iterative modifications.

\paragraph{AnyText.}

% 中文：AnyText 是阿里巴巴达摩院提出的多语言视觉文本生成与编辑模型及其配套基准。
% 该基准采用单轮设置，聚焦场景图像中的文本生成与编辑任务。
% 模型以文本位置掩码为输入，指定文本区域。
% 该基准的核心任务是图像内文本编辑，其数据构建过程包含一定程度的人工核验。
AnyText is a multilingual visual text generation and editing model along with its accompanying benchmark proposed by Alibaba DAMO Academy. The benchmark focuses on text generation and editing tasks within scene images, operating in a single-turn setting. The model uses text position masks as input to specify text regions. The core task of this benchmark is text editing within images. The dataset undergoes a certain degree of human verification during its construction process.

\paragraph{TextEditBench.}

% 中文：TextEditBench 专门面向场景文本编辑，评估模型修改、替换或生成图像中文字的能力。
% 基准采用单轮设置，并经过人工核验，以确保编辑质量与评测可靠性。
% 它提供编辑区域掩码来指定文本编辑位置，核心任务是图像内文本编辑。
TextEditBench~\cite{gui2025texteditbenchevaluatingreasoningawaretext} is a benchmark dedicated to scene text editing, evaluating models' ability to modify, replace, or generate text within images. The benchmark operates in a single-turn setting and is human-verified to ensure editing quality and evaluation reliability. It provides editing region masks to specify text editing locations, and its core task is text editing within images.

\paragraph{Kontext-Bench.}

% 中文：Kontext-Bench 从众包真实使用场景构建图像—指令对，覆盖局部编辑、全局编辑、文字编辑、风格参考和角色参考五类任务。FLUX.1 Kontext 报告使用人工比较评价这些输出。
Kontext-Bench~\cite{labs2025flux1kontext} compiles image--prompt pairs from crowd-sourced real-world use cases across local editing, global editing, text editing, style reference, and character reference. The FLUX.1 Kontext report evaluates these outputs through human comparisons.

\paragraph{GIE-Bench.}

% 中文：GIE-Bench 是通用图像编辑基准，用于评估模型在多种编辑场景下的综合能力。
% 它采用单轮设置，并经过人工核验。
% 基准提供编辑区域掩码以辅助评估编辑精度，但不涉及图像内文本编辑。
GIE-Bench~\cite{qian2025giebench} is a general image editing benchmark that evaluates models' comprehensive capabilities across multiple editing scenarios. The benchmark operates in a single-turn setting and is human-verified. It provides editing region masks to assist in evaluating editing precision, but does not involve text editing within images.

\paragraph{Complex-Edit.}

% 中文：Complex-Edit 聚焦复杂编辑指令，评估模型处理包含多种编辑操作或复杂语义的自然语言指令的能力。
% 该基准采用单轮设置，并经过人工核验。
% 它通过复杂自然语言指令指定编辑内容，不提供编辑区域掩码，也不涉及图像内文本编辑。
Complex-Edit~\cite{yang2025complexedit} is an image editing benchmark focusing on complex editing instructions, evaluating models' ability to handle natural language instructions containing multiple editing operations or complex semantics. The benchmark operates in a single-turn setting and is human-verified. It specifies editing content through complex natural language instructions without providing editing region masks, and does not involve text editing within images.

\paragraph{EBench-18K.}

% 中文：EBench-18K 从 1,080 个源图像与编辑提示出发，收集 17 个编辑模型生成的约 18,000 张图像，并提供细粒度人工偏好标注。LMM4Edit 在这些数据上学习评价编辑质量、指令对齐与内容一致性。
EBench-18K~\cite{xu2025lmm4edit} collects approximately 18,000 outputs from 17 editors applied to 1,080 source-image and prompt pairs, with fine-grained human preference annotations. LMM4Edit learns to assess editing quality, alignment, and consistency from these data.

\paragraph{HQ-Edit.}

% 中文：HQ-Edit 是高质量、覆盖广泛的通用图像编辑数据集，编辑指令由 GPT-4V 生成，编辑结果由 DALL-E 3 生成。
% 它采用单轮设置，数据经自动化流程产生，没有对每条样本进行系统人工核验。
% 它不涉及图像内文本编辑，也不提供编辑区域掩码。
HQ-Edit~\cite{hui2024hqedit} is a high-quality and high-coverage general image editing dataset, where editing instructions are generated by GPT-4V and editing results are produced by DALL-E 3. The dataset operates in a single-turn setting. The data is generated through an automated pipeline without systematic per-sample human verification. It does not involve text editing within images and does not provide editing region masks.

\paragraph{AURORA-Bench.}

% 中文：AURORA-Bench 是包含人工标注的八类编辑任务基准，重点涵盖动作与推理相关的修改。该工作通过人工判断和判别式评测，考察编辑模型对物体交互、动作和场景变化的理解。
AURORA-Bench~\cite{krojer2024aurora} is a manually annotated benchmark covering eight editing tasks, with particular attention to action- and reasoning-centric changes. The work uses human judgments and discriminative evaluation to examine understanding of interactions, actions, and scene changes.

\paragraph{PIE-Bench++.}

% 中文：PIE-Bench++ 将 PIE-Bench 拓展到多属性和多物体编辑。其标注包含源提示、目标提示、编辑动作与属性映射，并由第二位标注人员复核，用于检验多个修改要求能否同时实现。
PIE-Bench++~\cite{huang2024paralleledits} extends PIE-Bench to multi-aspect and multi-object editing. Its annotations specify source and target prompts, edit actions, and aspect mappings, with review by a second annotator, enabling evaluation of whether several requested changes are achieved together.

\paragraph{TEdBench++.}

% 中文：TEdBench++ 是随 LEDITS++ 一同提出的 TEdBench 扩展版本，用于评估基于文本的图像编辑方法。
% 基准采用单轮设置，并经过人工核验。
% 它通过文本指令指定编辑内容，不提供编辑区域掩码，也不涉及图像内文本编辑。
TEdBench++~\cite{brack2024leditslimitlessimageediting} is an extended version of TEdBench, proposed alongside LEDITS++, for evaluating text-based image editing methods. The benchmark operates in a single-turn setting and is human-verified. It specifies editing content through text instructions without providing editing region masks, and does not involve text editing within images.

\paragraph{ImagenWorld.}

% 中文：ImagenWorld 在六项任务和六个领域上评估生成与编辑能力，涵盖文本密集图形和截图。
% 它使用细粒度人工标注和局部错误描述补充自动分数，为文档编辑提供更广泛的诊断背景，而非仅面向文档的专用基准。
ImagenWorld~\cite{sani2026imagenworld} evaluates generation and editing across six tasks and six domains, including text-heavy graphics and screenshots. It uses fine-grained human annotations and localized error descriptions to complement automated scores. It provides broader diagnostic context for document editing, rather than being a document-only benchmark.

\subsection{Multimodal Representations, Agents, and Evaluation}
\label{sec:broader multimodal research}

\paragraph{Representation geometry and incomplete observations.}
Document evaluation combines textual and visual signals whose representation geometry affects similarity. TNCSE constrains the norms of positive-pair sentence embeddings, while JTCSE combines magnitude alignment with cross-attention to improve sentence representations \citep{zong2025tncse,zong2026jtcse}. L2Dir extends joint norm and directional alignment to multimodal retrieval \citep{zong2026l2dir}. For incomplete multi-view data, dynamic deep graph learning reconstructs view-specific relations with a masked graph reconstruction objective \citep{zhang2025dynamic}. These approaches provide complementary background for understanding what learned similarity preserves; the region-level OCR measurements in VDE Bench additionally expose explicit textual and structural changes.

\paragraph{Data distribution and bias.}
The composition of evaluation data determines which capabilities aggregate scores emphasize. LCGC studies consistency-gradient conflicts for debiasing class-imbalanced semi-supervised learning, and ABM calibrates predictions using batch-wise feature information \citep{xing2025lcgc,yi2025abm}. Class-Aware Spectral Distribution Matching studies distribution alignment and long-tailed classes in dataset distillation \citep{wu2026spectral}. These works motivate attention to data composition alongside average performance. VDE Bench reports document-source, language, and editing-operation distributions and separates operation- and language-level results.

\paragraph{Interactive workflows and factual reasoning.}
Document manipulation is one component of broader workflows that retrieve information, reason over evidence, and execute changes. BrowserAgent interacts with web pages through human-inspired browsing actions \citep{yu2025browseragent}, while EntCollabBench evaluates role-specialized collaboration through execution traces and changes in enterprise system state \citep{yu2026entcollabbench}. SCOPE-Router examines cost-aware model selection for execution-oriented tasks \citep{yu2026scoperouter}. The survey of scientific agents organizes autonomous research around workflow capabilities and evaluation \citep{wang2026scientificagents}. At the reasoning level, RPO trains on reasoning suffixes using experience caching, and MR-ALIGN targets factuality through reasoning-process alignment \citep{yi2026rpo,wang2026mralign}. Together, these settings distinguish a system's intermediate reasoning and tool use from the correctness of its final artifact.

\paragraph{Retrieval and external evidence.}
Open-world retrieval evaluates whether a model can locate evidence satisfying detailed requirements. ShotFinder studies open-domain video-shot retrieval under controllable constraints, whereas RVMS-Bench studies video search and moment localization from incomplete, memory-like descriptions \citep{yu2026shotfinder,yu2026rvms}. Omni-DeepSearch starts from audio and requires cross-modal evidence search; EVID-Bench uses cross-video evidence to assess misinformation \citep{yu2026omnideepsearch,yu2026evid}. These benchmarks emphasize evidence localization and verification, which are also central to checking whether a document edit changes the requested region and preserves its surrounding content.

\paragraph{Temporal understanding and state preservation.}
Video understanding adds temporal relationships to the spatial structure present in documents. HIMMEL allocates semantic and motion information to separate encoding paths \citep{jin2026himmel}. Four Eyes Are Better Than Two and Team of One combine differentiated reasoning or complementary model responses for video understanding \citep{xie2025foureyes,xie2025team}. Delta-JEPA uses latent differences to retain action-sensitive dynamics, while OPIS evaluates the presence, identity, and structure of input-grounded object instances in generated video \citep{zhang2026deltajepa,wang2026opis}. The latter makes preservation of specific input content an explicit evaluation target, complementing document editing's requirement to preserve unedited text and layout.

\paragraph{Task-specific reliability and human judgments.}
Reliability also depends on the task and the judgments used to assess it. Multimodal Video Emotion Recognition with Reliable Reasoning Priors incorporates modality-specific reasoning into emotion recognition \citep{wang2025emotionpriors}. More Is Better combines modality experts and preference-oriented prediction refinement, while ZeroES studies zero-shot ensembles for open-vocabulary video emotion recognition \citep{xie2025more,xie2025zeroes}. FAIRGAMER evaluates social biases in interactive game-agent decisions \citep{shi2026fairgamer}. These studies address different output properties, reinforcing the value of explicit evaluation dimensions and human assessment rather than a single undifferentiated quality score.

\end{document}